\documentclass[runningheads]{llncs}

\usepackage{eccv}

\usepackage{eccvabbrv}

\usepackage{graphicx}
\usepackage{booktabs}
\usepackage{xcolor}
\usepackage{xspace}
\usepackage{enumitem}
\usepackage{wrapfig}

\usepackage{multirow}
\usepackage[numbers]{natbib}
\usepackage[ruled,vlined]{algorithm2e}
\def\xnet{ReCap\xspace}
\def\drift{SemDrift\xspace}
\def\core{CORE\xspace}
\def\corefull{COnditional frame REferencing\xspace}
\def\driftfull{Guided \underline{Sem}antic \underline{Drift} Correction\xspace}
\def\dino{DINOv3\xspace}

\newcommand{\colhead}[1]{%
  \parbox[t]{.23\textwidth}{\centering\scriptsize #1}%
}
\usepackage{tabularx}
\usepackage{dsfont} 

\usepackage[accsupp]{axessibility}  

\usepackage[pagebackref=true,breaklinks,colorlinks,citecolor=blue,linkcolor=blue,urlcolor=blue]{hyperref}
\usepackage{orcidlink}

\usepackage{pifont}
\newcommand{\cmark}{\ding{51}}%
\newcommand{\xmark}{\ding{55}}%

\begin{document}

\title{\vspace{-0.5em}ReCap: Lightweight Referential Grounding for Coherent Story Visualization\vspace{-0.5em}}

\titlerunning{ReCap: Lightweight Referential Grounding for Coherent Story Visualization}

\author{Aditya Arora\inst{1} \and
Akshita Gupta\inst{1} \and Pau Rodriguez \and Marcus Rohrbach \inst{1}}

\authorrunning{Arora et al.}

\institute{${}^{1}$Technical University of Darmstadt \& hessian.AI, Germany\\
\email{aditya.arora@tu-darmstadt.de}
\vspace{-2em}
}

\maketitle
\begin{abstract}
Story Visualization aims to generate a sequence of images that faithfully depicts a textual narrative that preserve character identity, spatial configuration, and stylistic coherence as the narratives unfold. Maintaining such cross-frame consistency has traditionally relied on explicit memory banks, architectural expansion, or auxiliary language models, resulting in substantial parameter growth and inference overhead.
We introduce \xnet, a lightweight consistency framework that improves character stability and visual fidelity without modifying the base diffusion backbone. \xnet's \core (\corefull) module treats anaphors, in our case pronouns, as visual anchors, activating only when characters are referred to by a pronoun and conditioning on the preceding frame to
propagate visual identity. This selective design avoids unconditional cross-frame conditioning and introduces only 149K additional parameters, a fraction of the cost of memory-bank and LLM-augmented approaches. To further stabilize identity, we incorporate \drift (\driftfull) applied only during
training. When text is vague or referential, the denoiser
lacks a visual anchor for identity-defining attributes,
causing character appearance to drift across frames, \drift corrects this by aligning denoiser representations
with pretrained DINOv3 visual embeddings, enforcing
semantic identity stability at zero inference cost. \xnet outperforms previous state-of-the-art, StoryGPT-V~\citep{shen2025storygptv}, on the two main benchmarks for story visualization by 2.63\% Character-Accuracy on FlintstonesSV
 and by 5.65\% on PororoSV, establishing a new state-of-the-art character consistency
on both benchmarks.
Furthermore, we extend story visualization to human-centric narratives derived from real films, demonstrating the
capability of \xnet beyond stylized cartoon domains.
\end{abstract}
    
\section{Introduction}
\label{sec:intro}

\begin{figure}[t]
    \centering
    \scalebox{0.9}{
    \includegraphics[width=\linewidth]{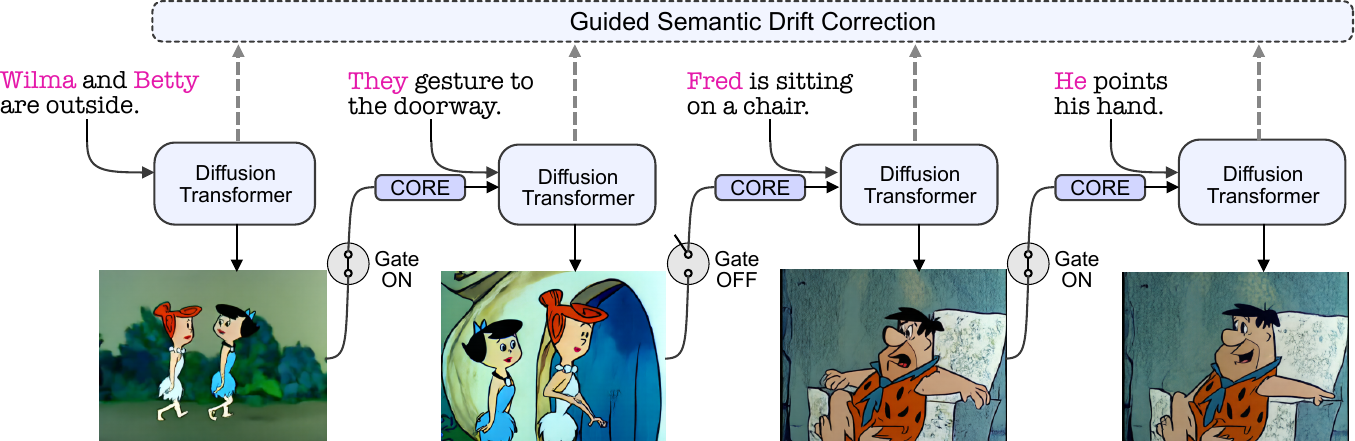}
    }
    \caption{Diffusion models condition each frame on text alone, failing when narrative references like ``\textcolor{magenta}{They}'' or ``\textcolor{magenta}{He}'' carry no appearance information. \xnet addresses this through a lightweight pipeline, its \core (\corefull) module activates selectively via Text Conditioned Gating, conditioning generation on the preceding frame only when the narrative references a character anaphorically, as in frames 2 and 4, while bypassing cross-frame conditioning for frames with explicit character names (frame 1). During training, \drift (\driftfull) aligns intermediate denoiser representations with pretrained visual embeddings, instilling temporally stable identity features that persist at inference without additional cost.}
    \label{fig:teaser_recap}
    \vspace{-5mm}
\end{figure}

\begin{figure*}[!t]
    \centering
    \scalebox{0.9}{
    \includegraphics[width=\linewidth]{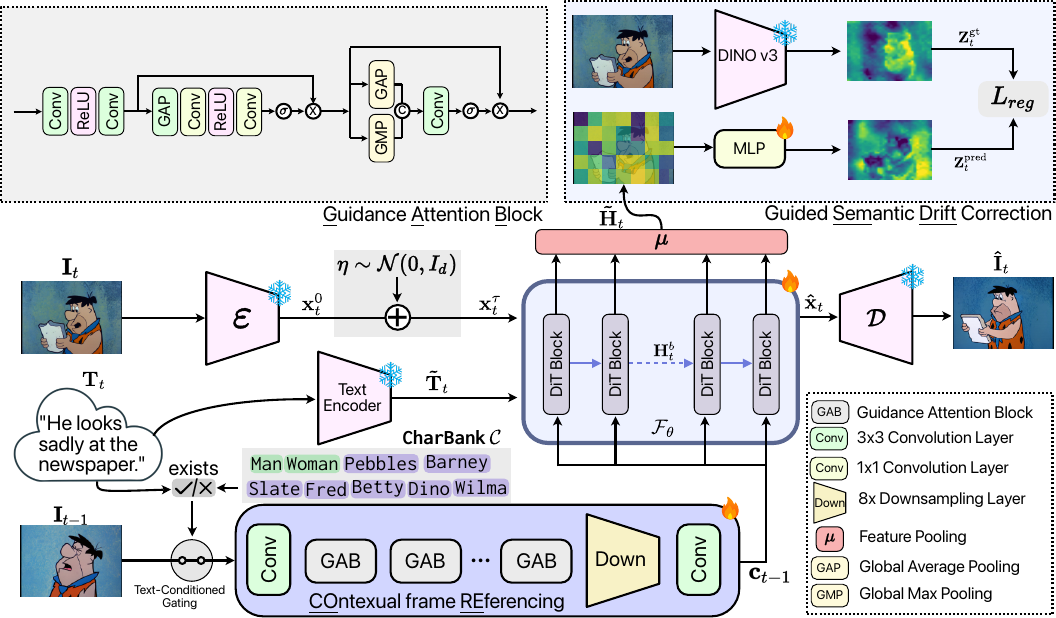}}
    \caption{\textbf{Overview of ReCap Architecture.} Our method extends
    Stable Diffusion 3 with two components: (1) \textbf{\core} (bottom):
    encodes the previous frame $\mathbf{I}_{t-1}$ into context embedding
    $\mathbf{c}_{t-1}$ (Eq. \ref{eq:core}) via a lightweight convolutional module with
    Guidance Attention Blocks, and injects it as a residual into each
    transformer block (Eq.~\ref{eq:recap_attention}), activated only when
    no character name appears in the current text prompt via Text-Conditioned
    Gating (Eq.~\ref{eq:guidance_mask}). (2) \textbf{\drift} (top right):
    aligns aggregated denoiser representations $\mathbf{Z}_{t}^{\mathrm{pred}}$,
    obtained via a three-layer MLP projecting from SD3 to DINOv3 feature dimensions, over all transformer
    block activations, with frozen DINOv3 embeddings $\mathbf{Z}_{t}^{\mathrm{gt}}$
    via spatial cosine similarity loss $\mathcal{L}_{\mathrm{reg}}$
    (Eq.~\ref{eq:dinov3_loss}), applied during training only and discarded
    at inference. The full training objective is given in
    Eq.~\ref{eq:total_loss}. The input text $\mathbf{T}_t$ (e.g., ``He looks sadly at the newspaper.'') is denoted as \textit{story-text}. Frozen with snowflake icons; trainable with fire icons.}
\label{fig:overview}
\end{figure*}

Generating coherent visual narratives from text requires more than photorealistic image synthesis, it demands that characters remain visually consistent across frames as the story unfolds. Modern text-to-image models~\citep{ramesh2022dalle2, saharia2022imagen, esser2024scaling, labs2025flux, qin2025lumina} have brought automation closer, synthesizing high-fidelity images from text with remarkable realism — but they process each frame independently, with no memory of who appeared before or what they looked like~\citep{liu2024intelligent, he2025dreamstory}. The same character looks different in every panel: a failure that is not merely aesthetic but narrative, because a story whose characters cannot be recognized across frames is not a story at all. The challenge runs deeper than visual quality. Story narratives introduce characters by name and then refer to them through pronouns and implicit relations across subsequent sentences~\citep{li2019storygan, Maharana2021VLCStoryGAN, Maharana2022StoryDALLE, rahman2023makeastory, shen2025storygptv}. Consider the narrative shown in Fig.~\ref{fig:overview}: ``Wilma and Betty are outside. They gesture to the doorway. Fred is sitting on a chair. He points his hand.'' A model must (1) establish Wilma and Betty's visual identities in frame 1, (2) correctly resolve ``They'' to those same two characters in frame 2, (3) introduce a distinct male character Fred in frame 3, and (4) correctly resolve ``He'' to Fred in frame 4 without conflating identities with Wilma or Betty. Pronouns~\citep{yu2019vispro, goel2023cin} must ground to stable visual entities across frames, even as actions and contexts evolve — a problem of referential grounding~\citep{lee2017coref, joshi2019bertcoref} that standard diffusion models, which condition each frame on text alone, are not designed to handle.

Recent diffusion-based story visualization systems have made substantial progress in image quality and stylistic richness~\citep{ho2020ddpm, rombach2022ldm, peebles2022dit}, but maintaining cross-frame coherence has come at a high cost. The dominant approach encodes prior context into explicit memory banks, recurrent modules, or auxiliary language models, mechanisms queried at every generation step~\citep{rahman2023makeastory, shen2025storygptv, shen2025boosting}. While effective, these designs increase parameter count, training complexity, and inference cost, treating coherence as an architectural problem requiring external storage rather than a capacity a well-trained model can develop internally. We argue this overhead is unnecessary.

The text stream already carries a reliable signal for when referential grounding is needed: pronouns mark exactly the moments requiring backward reference. And when it does, the single most relevant piece of visual context is the immediately preceding frame — which already contains the identity information needed to resolve the reference. This observation motivates \xnet (Fig.~\ref{fig:teaser_recap}): rather than maintaining a growing memory of all previous frames, we condition each generation step on only the previous image, and only when the text signals that grounding is required. The core of our approach is \core (\corefull), a lightweight attention module that injects this context directly into the cross-attention computation via a single additive projection — introducing no recurrent states, no external memory, and no auxiliary models.

To further stabilize identity and scene consistency across frames, we introduce a \drift (\driftfull) that aligns intermediate denoiser representations with pretrained \dino~\citep{Simeoni2025DINOv3} embeddings of ground-truth frames via cosine similarity. Rather than enforcing pixel-level matching, this representation-level supervision encourages the model to develop semantically stable internal descriptors of characters and scenes, so that coherence at inference time emerges from learned semantic representations rather than
pixel-level constraints. The regularizer is applied only during training through a compact MLP projection and is not needed at inference, adding no runtime cost.

\xnet achieves state-of-the-art results on both FlintstonesSV~\citep{gupta2018imagine} and PororoSV~\citep{li2019storygan}, outperforming StoryGPT-V which relies on an auxiliary LLM and multi-frame context encoding by 2.63\% Char-Acc and +1.69 FID on FlintstonesSV and by 5.65\% Char-Acc and 4.51\% Char-F1 on PororoSV, while adding only 149K additional parameters to the base model. These gains generalize to narratives derived from real films~\citep{hong2023visual}, showing that our lightweight referential conditioning approach, \xnet,
is not domain-specific. 

Our contributions are as follows: (i) \core (\corefull), a lightweight module that conditions generation on the preceding frame only when the text signals anaphoric reference, adding only 149K parameters to the base diffusion model. (ii) A \drift (\driftfull) that enforces semantic identity stability during training only, at zero inference cost.

\section{Related Work}
\label{sec:relatedwork}

\textbf{Story Visualization.} The task of generating coherent image sequences from narrative text has evolved significantly since StoryGAN~\citep{li2019storygan}, which introduced a sequential conditional GAN with dual discriminators to maintain frame-level quality and story-level consistency. Building on this foundation, CP-CSV~\citep{song2020cpcsv}, DuCoStoryGAN~\citep{maharana2021improving}, and VLCStoryGAN~\citep{Maharana2021VLCStoryGAN} improved coherence through contextual planning, semantic-consistency training, and multimodal structure, while VP-CSV~\citep{chen2022vpcsv} and multimodal semantic alignment approaches~\citep{li2022multimodalsemanticstory} further emphasized character consistency. StoryDALL-E~\citep{Maharana2022StoryDALLE} demonstrated that adapting pretrained text-to-image transformers~\citep{ramesh2021zero, ramesh2022hierarchical} for story continuation improves visual continuity, establishing the now-standard practice of building story visualization systems on top of powerful single-image generators~\citep{rombach2022ldm, saharia2022imagen, peebles2022dit, esser2024scaling}. More recent frameworks including AR-LDM~\citep{pan2024arldm}, StoryImager~\citep{tao2024storyimager}, and TaleCrafter~\citep{gong2023talecrafter} continue to improve quality and multi-character coherence by leveraging increasingly capable diffusion backbones. However, despite these advances, evaluation has remained concentrated on cartoon-specific benchmarks such as FlintstonesSV~\citep{gupta2018imagine} and PororoSV~\citep{li2019storygan}. While finetuning strong diffusion models on these datasets yields competitive character accuracy on non-referential frames, performance degrades on sentences containing pronouns rather than character names, a failure mode that aggregate metrics obscure but referential benchmarks expose~\citep{rahman2023makeastory, shen2025storygptv}. We extend evaluation to Visual Writing Prompts~\citep{hong2023visual}, a benchmark of human-written visual narratives with naturalistic, pronoun-heavy language, demonstrating that lightweight referential grounding transfers beyond the cartoon domain.\smallskip\\

\noindent \textbf{Reference Resolution in Story Visualization.} In narrative text, pronouns reliably signal referential dependency on prior discourse context~\citep{lee2017coref, joshi2019bertcoref}, and visual reference resolution has been studied in visual dialog~\citep{das2017visual,Kottur_2018_ECCV}, movie descriptions \cite{rohrbach17cvpr}, and grounding~\citep{yu2019vispro, goel2023cin}. Translating this into story visualization, Make-A-Story~\citep{rahman2023makeastory} introduced an autoregressive diffusion framework with a visual memory module and sentence-conditioned soft attention, while CMOTA~\citep{ahn2023cmota} employs context memory with online text augmentation. StoryGPT-V~\citep{shen2025storygptv} addresses pronoun resolution by aligning a 6.7B-parameter large language model with character-aware latent diffusion using cross-attention map supervision, and StoryImager~\citep{tao2024storyimager} introduces a contextual feature extractor within a unified visualization and completion framework. These methods share a common assumption: that coherence requires dedicated external storage, auxiliary language models, or growing context encoders that scale in cost with story length. We challenge this assumption on two fronts. First, \core conditions generation on only the immediately preceding frame and only when the story-text signals referential grounding is required, adding 149K parameters with no auxiliary models and no inference-time memory overhead. Second, rather than enforcing consistency through architectural mechanisms at inference time, our \drift module aligns intermediate denoiser representations with pretrained self-supervised embeddings~\citep{caron2021emerging, oquab2023dinov2} via cosine similarity during training only — a form of semantic supervision not previously applied to story visualization, discarded entirely at inference, that encourages identity stability to emerge from the model's learned weights rather than from external retrieval.

\section{\xnet: \underline{Re}ferential \underline{C}ontext \underline{A}ttention \underline{P}rojection}
\label{sec:method}

\subsection{Overview}

\textbf{Problem Formulation.} We address the problem of generating visually coherent story sequences from story-text, \ie narratives where entities are introduced by name and later referenced through pronouns (\eg, ``A woman enters. She smiles.''). Although pretrained diffusion models such as \emph{SD3}~\citep{esser2024scaling} achieve impressive image fidelity, they often fail to preserve character identity once explicit name tokens disappear. This occurs because the text encoder and cross-attention layers in standard diffusion transformers lack a mechanism to maintain referential grounding across frames~\citep{vaswani2017attention,peebles2022dit,esser2024scaling}. Unlike single-image generation, this requires maintaining \emph{referential grounding}: the model must correctly resolve pronouns (\eg, ``she'', ``he'') to the appropriate visual entities established in previous frames.

During training, we assume access to triplets $(\mathbf{I}_t, \mathbf{I}_{t-1}, \mathbf{T}_t)$,
where $\mathbf{I}_t$ is the ground-truth image for the associated story-text $\mathbf{T}_t$, and $\mathbf{I}_{t-1}$ provides visual context from the preceding frame. We denote the SD3 denoiser by $\mathcal{F}_{\theta}$. At diffusion step $\tau$: The input image $\mathbf{I}_t \in \mathbb{R}^{3 \times H \times W}$ is first encoded into a latent representation $\mathbf{x}_t^0 \in \mathbb{R}^{C \times H' \times W'}$ using a VAE encoder $\mathcal{E}$, where $H' = H/8$ and $W' = W/8$. Gaussian noise is then added to obtain the noisy latent $\mathbf{x}_t^\tau$ at diffusion timestep $\tau$. The SD3 denoiser $\mathcal{F}_{\theta}$ then predicts the noise:
\begin{equation}
\left(\widehat{\boldsymbol{\epsilon}}_t,\mathbf{H}_t\right)=
\mathcal{F}_{\theta}\!\left(
\mathbf{x}_{t}^\tau,\tilde{\mathbf{T}}_t;\mathbf{c}_{t-1},m_t
\right),
\end{equation}
where $\tilde{\mathbf{T}}_t \in \mathbb{R}^{N \times d}$ is the text embedding obtained by encoding $\mathbf{T}_t$ via the frozen text encoder, $\widehat{\boldsymbol{\epsilon}}_t$ is the predicted noise and $\mathbf{H}_t=\{\mathbf{H}_t^b\}_{b=1}^{B}$ denotes stacked hidden features from SD3 blocks. The predicted noise is used to iteratively denoise $\mathbf{x}_t^\tau$ to obtain the denoised estimate $\hat{\mathbf{x}}_t$, which is decoded into the final image $\hat{\mathbf{I}}_t$ via a VAE decoder $\mathcal{D}$. Here, $\mathbf{c}_{t-1}$ is the context embedding and $m_t$ is the guidance mask (both explained in~\cref{sec:core}). This autoregressive conditioning provides implicit visual grounding for pronouns without requiring future frames or annotated entity links~\citep{li2019storygan,Maharana2022StoryDALLE,rahman2023makeastory,pan2024arldm}. To address referential grounding in story generation, we introduce \xnet, which consists of two components. The first is \core (COnditional frame REferencing), a lightweight attention module that conditions each denoising step on the previously generated frame through a single cross-attention projection, activated only when the text signals that referential grounding is required. The second is \drift (Guided Semantic Drift Correction) that strengthens semantic consistency by aligning aggregated denoiser features with ground-truth self-supervised backbone representations~\citep{Simeoni2025DINOv3} via cosine similarity, applied during training only and discarded at inference, thus adding only 149K parameters to the base SD3 model.

\begin{algorithm}[t]
\resizebox{.93\columnwidth}{!}{
\begin{minipage}{1.0\columnwidth}
\small
\caption{\xnet Training and Inference}
\label{alg:recap}
\DontPrintSemicolon
\KwIn{Story-text $\{\mathbf{T}_t\}_{t=1}^n$, guidance masks $\{\mathbf{m}_t\}_{t=1}^n$ where $m_t=\mathds{1}[\mathbf{T}_t \cap \mathcal{C} = \emptyset]$ (Eq.~\ref{eq:guidance_mask}), GT frames $\{\mathbf{I}^{\mathrm{gt}}_t\}_{t=1}^n$ (training only)}
\KwOut{Generated frames $\{\mathbf{I}_1,\dots,\mathbf{I}_n\}$}
\BlankLine
\textbf{Training:}\;
\For{$t=1$ \KwTo $n$}{
$\mathbf{c}_{t-1} \gets \mathds{1}[t>1]\cdot \texttt{CORE}(\mathbf{I}^{\mathrm{gt}}_{t-1})$
\tcp*{Eq.~\ref{eq:core}}
  Sample $(\tau,\boldsymbol{\epsilon})$, form noisy latent $\mathbf{x}_t^\tau$ from $\mathbf{I}^{\mathrm{gt}}_t$\;
  $(\widehat{\boldsymbol{\epsilon}}_t,\mathbf{H}_t)\gets \mathcal{F}_{\theta}\!\left(\mathbf{x}_{t}^\tau,\tilde{\mathbf{T}}_t;\,\texttt{context\_img}=\mathbf{c}_{t-1},\,\texttt{guidance\_mask}=\mathbf{m}_t\right)$
  \tcp*{Eq.~\ref{eq:recap_attention}}
  $\tilde{\mathbf{H}}_t \gets \frac{1}{B}\sum_{b=1}^{B}\mathbf{H}_t^b$
  \tcp*{Eq.~\ref{eq:dinov3_loss}}
  $\mathbf{Z}_t^{\mathrm{pred}} \gets \operatorname{\texttt{MLP}}(\tilde{\mathbf{H}}_t)$
  \;
  $\mathbf{Z}_t^{\mathrm{gt}} \gets \mathrm{DINOv3}(\mathbf{I}^{\mathrm{gt}}_t)$
  \;
  $\mathcal{L}_{\text{denoise}} \gets \ell_{\text{FM}}(\widehat{\boldsymbol{\epsilon}}_t, \boldsymbol{\epsilon};\tau)$
  \tcp*{SD3 flow-matching loss~\cite{lipman2023flowmatching}}
  $\mathcal{L}_{\text{reg}} \gets \ell_{\cos}\!\left(\mathbf{Z}_t^{\mathrm{pred}},\mathbf{Z}_t^{\mathrm{gt}}\right)$ \tcp*{spatial cosine similarity, Eq.~\ref{eq:dinov3_loss}}
  $\mathcal{L}_{\text{total}} \gets \mathcal{L}_{\text{denoise}} + \lambda_{\text{reg}}\mathcal{L}_{\text{reg}}$
  \tcp*{$\lambda_{\text{reg}}{=}0.5$; Eq.~\ref{eq:total_loss}}
  Update $(\mathcal{F}_\theta, \texttt{CORE}, \texttt{MLP})$ using $\nabla \mathcal{L}_{\text{total}}$\;
}
\BlankLine
\textbf{Inference:}\;
\For{$t=1$ \KwTo $n$}{
  $\mathbf{c}_{t-1} \gets \mathds{1}[t>1]\cdot \texttt{CORE}(\mathbf{I}_{t-1})$
  \;
  $\mathbf{I}_t \gets \mathrm{SD3Sample}_{\theta}\!\left(\mathbf{T}_t;\,\texttt{context\_img}=\mathbf{c}_{t-1},\,\texttt{guidance\_mask}=\mathbf{m}_t\right)$
}
\end{minipage}
}
\end{algorithm}

\subsection{\core: \underline{CO}nditional frame \underline{RE}ferencing}
\label{sec:core}

At each denoising timestep $\tau$, SD3 operates on a noisy latent
$\mathbf{x}_t^\tau$, with cross-attention conditioned on the story-text embedding $\tilde{\mathbf{T}}_t$~\citep{vaswani2017attention, rombach2022ldm,
peebles2022dit}. To enable referential grounding, we condition the
denoising process on the previous frame whenever character names are
missing. Given the previous frame $\mathbf{I}_{t-1}$, we extract a
context embedding
\begin{equation}
\mathbf{c}_{t-1} = \text{\core}(\mathbf{I}_{t-1}),
\label{eq:core}
\end{equation}
where \core encodes $\mathbf{I}_{t-1}$ through a lightweight learned
module and projects the result to the text-token dimension $d$.
During
training, $\mathbf{I}_{t-1} = \mathbf{I}^{\mathrm{gt}}_{t-1}$ is the
ground-truth frame; at inference, $\mathbf{I}_{t-1}$ is the previously
generated frame.

\core operates by injecting a residual context tensor into the attention
logits of each transformer block. The standard attention logits are
computed as:
\begin{equation}
\mathbf{A}_i = \frac{\mathbf{Q}_i \mathbf{K}_i^{\top}}{\sqrt{d}},
\end{equation}
where $\mathbf{Q}_i, \mathbf{K}_i \in \mathbb{R}^{L \times d}$ are the
query and key matrices of block $i$. \core modifies these logits by
injecting the previous-frame context:
\begin{equation}
\mathbf{A}_i' = \mathbf{A}_i + \mathbf{c}_{t-1} \cdot m_t
\label{eq:recap_attention}
\end{equation}
where $m_t \in \{0,1\}$ is the binary guidance mask. When $m_t = 0$, the residual is zeroed and the block operates identically to standard SD3.

\noindent\textbf{Text-conditioned gating.} Unlike prior sequential models that
condition uniformly on all previous frames, \core is text-aware,
activating only when referential grounding is needed. We define the
binary guidance mask $m_t \in \{0,1\}$ as:
\begin{equation}
m_t = \mathds{1}\!\left[\mathbf{T}_t \cap \mathcal{C} = \emptyset\right],
\label{eq:guidance_mask}
\end{equation}
where $\mathcal{C}$ is the set of all character names in the narrative,
constructed via string matching on tokenized
text~\citep{rahman2023makeastory, shen2025storygptv}. $m_t$ is
precomputed before the denoising loop: when $m_t = 1$, no character
name appears in $\mathbf{T}_t$ and \core injects the context residual;
when $m_t = 0$, the residual is zeroed and the block operates
identically to standard SD3, retaining full flexibility to generate
novel content.

\subsection{SemDrift: Guided \underline{Sem}antic \underline{Drift} Correction}

While \core provides referential grounding at the attention level, the diffusion model may still drift semantically across frames, as the denoiser lacks a reliable visual anchor for identity-defining attributes such as facial structure, clothing, and pose when text descriptions are vague or referential~\citep{avrahami2024chosen, zhou2024storydiffusion}. To address this, we introduce a feature-level regularizer that aligns intermediate denoiser activations with pretrained self-supervised backbone descriptors of the corresponding ground-truth frame~\citep{Simeoni2025DINOv3}. Unlike guidance mechanisms applied at inference, this regularizer operates during training only and is discarded
at inference. Among available self-supervised encoders, we chose DINOv3, as it provides strong object- and region-level correspondence through ViT-based self-distillation pretraining, capturing dense semantic structure beyond the global classification features of encoders such as CLIP~\citep{radford2021learning} or MAE~\citep{he2022mae}~\citep{oquab2023dinov2, Zhang2023SDComplementsDINO}, providing a language agnostic visual anchor for the identity defining attributes that referential text omits. Let the SD3 forward pass output stacked hidden features
$\mathbf{H}_t=\{\mathbf{H}_t^b\}_{b=1}^{B}$.
We then aggregate across the block dimension before SemDrift:
\begin{equation}
\tilde{\mathbf{H}}_t=\frac{1}{B}\sum_{b=1}^{B}\mathbf{H}_{t}^{b}.
\end{equation}
The predicted semantic feature is then obtained by passing $\tilde{\mathbf{H}}_{t}$ through a three layer MLP block $\mathbf{Z}_{t}^{\mathrm{pred}} = \operatorname{MLP}\!\left(\tilde{\mathbf{H}}_{t}\right)$. The resulting representation is supervised against the DINOv3 embedding of the corresponding ground-truth frame
$\mathbf{Z}_{t}^{\mathrm{gt}} = \operatorname{DINOv3}\!\left(\mathbf{I}^{\mathrm{gt}}_t\right)$
via a spatial cosine similarity loss:
\begin{equation}
\mathcal{L}_{\mathrm{reg}} = 1-\frac{1}{|\Omega|}\sum_{u\in\Omega}
\frac{\mathbf{Z}_{t}^{\mathrm{pred}}(u)\cdot \mathbf{Z}_{t}^{\mathrm{gt}}(u)}
{\|\mathbf{Z}_{t}^{\mathrm{pred}}(u)\|_2\|\mathbf{Z}_{t}^{\mathrm{gt}}(u)\|_2},
\label{eq:dinov3_loss}
\end{equation}
where $u$ indexes spatial locations over the set $\Omega$. We adopt cosine similarity because it enforces directional
alignment in the embedding space while remaining invariant
to scale differences across layers, stabilizing training
by penalizing semantic drift without collapsing feature
magnitudes.

\subsection{Training Objective}

The overall training objective combines the standard SD3 flow-matching denoising loss with the \drift module~\citep{lipman2023flowmatching}:
\begin{equation}
\mathcal{L}_{\text{total}} =
\mathcal{L}_{\text{denoise}} +
\lambda_{\text{reg}} \, \mathcal{L}_{\text{reg}},
\label{eq:total_loss}
\end{equation}
where $\lambda_{\text{reg}}$ balances image fidelity and semantic consistency, set to $0.5$ in all experiments. All \core and \drift parameters are optimized jointly. The DINOv3 regularizer is discarded entirely at inference, contributing zero runtime cost.

\section{Experiments}
\label{sec:experiments}

\noindent\textbf{Datasets.} We evaluate our approach on two standard story-visualization benchmarks: FlintstonesSV~\citep{gupta2018imagine} and PororoSV~\citep{li2019storygan}. FlintstonesSV contains 20,132 training, 2,071 validation, and 2,309 test stories, with 7 main characters and 323 background categories. PororoSV contains 10,191 training, 2,334 validation, and 2,208 test samples, with 9 main characters. We additionally evaluate on Visual Writing Prompts (VWP)~\citep{hong2023visual}, a benchmark of human-written narratives derived from real films, to assess effectiveness
beyond stylized cartoon domains. Both datasets consist of short story segments in which characters and backgrounds reoccur across frames, forming a scene, making them suitable for evaluating referential consistency under autoregressive generation. We follow the referential-text setting introduced by Make-A-Story~\citep{rahman2023makeastory}. Given a text sequence $\{\mathbf{T}_1,\dots,\mathbf{T}_n\}$, models generate one image per sentence autoregressively. For $t>1$, repeated entity mentions in $\{\mathbf{T}_t\}$ are replaced with the appropriate
pronoun or possessive from $\mathcal{P} := \{\text{he, she, they, her, his, their, \ldots}\}$, selected according to the gender and number of the referenced entity, to enforce implicit reference resolution.\smallskip\\

\noindent\textbf{Implementation Details.} Our backbone is Stable Diffusion 3 (SD3)~\citep{esser2024scaling}. We fine-tune SD3 with the proposed \core module and \drift using triplets $(\mathbf{I}_t, \mathbf{I}_{t-1}, \mathbf{T}_t)$. Images are resized to $512 \times 512$ pixels during training. The total loss follows Eq.~\eqref{eq:total_loss} with $\lambda_{\text{feat}}=0.5$. Training runs for 100k iterations on $8\times$H100 GPUs with a global batch size of 8, mixed precision, and a learning rate of $5\times10^{-5}$ for the CORE module and a smaller learning rate of $1\times10^{-5}$ for the diffusion transformer. The additional parameters introduced by \core and the projection MLP in \drift account for less than 0.5\% of the base model size. At inference time, only the text sequence is available. We generate the first frame from $\mathbf{T}_1$ and proceed autoregressively for $t>1$, conditioning each frame on the previously generated image via \core when referential grounding is required.\smallskip\\

\noindent\textbf{Evaluation Metrics.} We follow standard story-generation evaluation protocols~\citep{li2019storygan,Maharana2021VLCStoryGAN,Maharana2022StoryDALLE} and the super-resolution setup from StoryGPT-V~\citep{shen2025storygptv}. Generated frames are evaluated using the fine-tuned Inception-v3 classifiers released with StoryGPT-V~\cite{shen2025storygptv}. For FlintstonesSV, we report character accuracy and F1 (Char-Acc/F1), as well as background accuracy and F1 (BG-Acc/F1); for PororoSV, only character metrics are available. We additionally report Fréchet Inception Distance (FID) to measure the distributional alignment between real and generated images. Higher classification scores and lower FID indicate better consistency and visual quality. For VWP, we report VBench~\citep{vbench} metrics, including subject consistency, background consistency, and FID.

\section{Results}
\label{sec:results}

\begin{table*}[t]
\begin{center}
\caption{\textbf{Quantitative comparison on story generation for 
FlintstonesSV~\citep{gupta2018imagine} and 
PororoSV~\citep{li2019storygan}.} Methods are shared across both 
datasets and reported in a unified table. Background labels are 
available only for FlintstonesSV, so BG Acc and BG F1 are reported 
only for that dataset. Params reflects the number of parameters 
added on top of the base diffusion model. Lower FID and fewer Params are better; higher scores are better for all other metrics.}
\label{table:main_results}
\vspace{-2mm}
{\setlength{\tabcolsep}{4pt}
\resizebox{\textwidth}{!}{
\begin{tabular}{l c c c c c @{\hspace{3pt}\vrule width 0.6pt\hspace{3pt}} c c c @{\hspace{3pt}\vrule width 0.6pt\hspace{3pt}} c}
\toprule[0.15em]
\textbf{} & \multicolumn{5}{c}{\textbf{FlintstonesSV}} & \multicolumn{3}{c}{\textbf{PororoSV}} & \textbf{Coherence} \\
\textbf{Method} & FID~$\downarrow$ & Char Acc~$\uparrow$ & Char F1~$\uparrow$ & BG Acc~$\uparrow$ & BG F1~$\uparrow$ & FID~$\downarrow$ & Char Acc~$\uparrow$ & Char F1~$\uparrow$ & \textbf{Params}~$\downarrow$ \\
\midrule[0.15em]
StoryLDM~\cite{rahman2023makeastory} & 36.34 & 77.23 & 88.26 & 54.97 & 60.99 & 26.64 & 29.14 & 57.56 & 16.75M \\
StoryDALL-E~\cite{Maharana2022StoryDALLE} & 44.66 & 61.83 & 78.36 & 48.10 & 54.92 & 40.39 & 21.03 & 50.56 & 29.1M \\
StoryGPT-V~\cite{shen2025storygptv} & 26.41 & 89.23 & 95.10 & 59.71 & 64.72 & 19.56 & 36.06 & 62.70 & 35.0M \\
\midrule[0.15em]
\textbf{\xnet (Ours)} & \textbf{24.72} & \textbf{91.86} & \textbf{96.07} & \textbf{59.74} & \textbf{64.95} & \textbf{19.26} & \textbf{41.71} & \textbf{67.21} & \textbf{149K}\\
\bottomrule[0.15em]
\end{tabular}}}
\end{center}\vspace{-1.5em}
\end{table*}

\subsection{State-of-the-Art Comparison}
Table~\ref{table:main_results} summarizes quantitative results on \textit{FlintstonesSV}~\citep{gupta2018imagine} and \textit{PororoSV}~\citep{li2019storygan} compared to prior story generation methods. ReCap achieves the state-of-the-art performance across character consistency and image quality metrics on both datasets. On FlintstonesSV, ReCap achieves the lowest FID (24.72) and improves character consistency over StoryGPT-V by +2.63\% Char-Acc and +0.97\% Char-F1.
It also improves BG-Acc and BG-F1, indicating better scene stability and reduced background drift across frames. On PororoSV, ReCap again achieves the lowest FID (19.26) and delivers larger character-consistency gains over StoryGPT-V by +5.65\% Char-Acc (36.06 $\rightarrow$ 41.71) and +4.51\% Char-F1 (62.70 $\rightarrow$ 67.21). PororoSV does not provide background labels, so BG metrics are not reported. These results indicate that \xnet improves character consistency across both datasets while maintaining competitive image quality. Compared to StoryGPT-V~\citep{shen2025storygptv}, which relies on a 6.7B-parameter LLM and multi-frame context encoding, adding only 149K parameters to the base model.

\subsection{Efficiency Analysis}
ReCap achieves its accuracy gains while remaining highly efficient compared
to prior coherence mechanisms, as reflected in the \textit{Params} column of
Table~\ref{table:main_results}. StoryDALL-E~\citep{Maharana2022StoryDALLE}
relies on heavier continuation blocks to copy visual elements across frames
(29.1M), Make-A-Story~\citep{rahman2023makeastory} adds
explicit visual-memory attention over all prior frames (16.75M), and
StoryGPT-V~\citep{shen2025storygptv} introduces a full LLM-driven
reference-resolution path with mapper modules (35.0M). ReCap, in contrast,
keeps coherence inside the diffusion pipeline through a single additive
cross-frame attention projection activated only for referential sentences,
adding only 149K parameters to the base diffusion model, a fraction of the cost
of all competing methods. Crucially, \core activates only when referential
grounding is required: computational cost scales with the number of
pronoun-containing sentences rather than story length, making the method
particularly efficient for narratives with varied referential density.
The \drift module is discarded entirely after training, contributing
zero additional parameters at inference time.

\subsection{Ablation Studies}
\begin{table*}[t]
\begin{center}
\caption{Ablation study of ReCap components on FlintstonesSV~\citep{gupta2018imagine}. 
\core provides the primary performance gain over the baseline, 
while \drift further improves character consistency.}
\label{table:ablation}
\vspace{-2mm}
\setlength{\tabcolsep}{3pt}
\scalebox{0.85}{
\begin{tabular}{l c c}
\toprule[0.15em]
\textbf{Method} & Char Acc~$\textcolor{black}{\uparrow}$ & Char F1~$\textcolor{black}{\uparrow}$\\
\midrule[0.15em]
ReCap & \textbf{91.86} & \textbf{96.07} \\
\midrule
ReCap w/o \core & 82.20 & 91.20 \\
ReCap w/o \drift & 91.04 & 95.91 \\
ReCap w/o \core~\&~\drift & 78.48 & 89.44 \\
\bottomrule[0.1em]
\end{tabular}}
\end{center}\vspace{-1.5em}
\end{table*}

Table~\ref{table:ablation} validates the contribution of each ReCap component
on FlintstonesSV. The ReCap model achieves 91.86\% Char-Acc and 96.07\% Char-F1.
Removing \core drops Char-Acc to 82.20\% ($-$9.66), demonstrating that
referential attention projection is crucial for identity propagation, without
it, the model has no mechanism to connect pronouns to previously established
visual entities and is forced to generate characters from story-text alone. Removing
\drift drops Char-Acc to 91.04\% ($-$0.82), confirming that semantic
alignment stabilizes character embeddings; without \drift, the diffusion process
drifts into semantically inconsistent regions of the latent space, causing
character appearance to degrade across frames. Using \drift alone without \core
achieves only 82.20\% Char-Acc, showing that semantic regularization helps but
cannot substitute for explicit referential grounding — \drift provides global
consistency but lacks frame-to-frame entity linking. Removing both \core and
\drift reduces Char-Acc to 78.48\%, confirming the complementary nature of the
two components: \core provides targeted cross-frame attention when pronouns
appear, while \drift stabilizes the entire generation process, together
accounting for a total gain of $+$13.38 Char-Acc over the baseline.

\begin{table*}[!t]
\begin{center}
\begin{minipage}{0.48\textwidth}
\centering
\caption{Ablation on semantic regularization weight $\lambda_{\text{reg}}$ 
on FlintstonesSV~\citep{gupta2018imagine}. Setting $\lambda_{\text{reg}}=0.5$ 
in Eq.~\ref{eq:total_loss} achieves optimal background consistency.}
\label{table:lambda_ablation}
\vspace{-2mm}
\setlength{\tabcolsep}{6pt}
\scalebox{.85}{
\begin{tabular}{l c c}
\toprule[0.1em]
\textbf{Method} & BG Acc~$\uparrow$ & BG F1~$\uparrow$\\
\midrule[0.1em]
$\lambda_{\text{reg}}=0.01$ & 57.44 & 62.83 \\
$\lambda_{\text{reg}}=0.1$  & 58.64 & 63.65 \\
$\lambda_{\text{reg}}=1.0$  & 59.34 & 64.44 \\
\midrule
$\lambda_{\text{reg}}=0.5$  & \textbf{59.74} & \textbf{64.95} \\
\bottomrule[0.1em]
\end{tabular}}
\end{minipage}
\hfill
\begin{minipage}{0.48\textwidth}
\centering
\caption{Ablation on classifier-free guidance scale on 
FlintstonesSV~\citep{gupta2018imagine}. CFG=5.0 achieves 
the best balance between image quality and character consistency across 
all metrics.}
\label{table:cfg_ablation}
\vspace{-2mm}
\setlength{\tabcolsep}{6pt}
\scalebox{.85}{
\begin{tabular}{l c c c}
\toprule[0.1em]
\textbf{Method} & FID~$\downarrow$ & Char Acc~$\uparrow$ & Char F1~$\uparrow$\\
\midrule[0.1em]
cfg=1.0 & 35.56 & 74.47 & 85.36 \\
cfg=7.5 & 26.22 & 92.09 & 96.16 \\
\midrule
cfg=5.0 & \textbf{24.72} & \textbf{91.86} & \textbf{96.07} \\
\bottomrule[0.1em]
\end{tabular}}
\end{minipage}
\end{center}
\vspace{-1.5em}
\end{table*}
\begin{table}[!t]
\centering
\caption{Comparison of visual backbones for \drift on 
FlintstonesSV~\citep{gupta2018imagine}. Both CLIP and DINOv3 improve 
over the baseline, with DINOv3 achieving the lowest FID while CLIP 
achieves marginally better character accuracy and F1 score.}
\label{tab:backbone_comparison}
\vspace{-2mm}
\setlength{\tabcolsep}{4pt}
\scalebox{.85}{
\begin{tabular}{l c c c}
\toprule[0.1em]
\textbf{Method} & FID~$\downarrow$ & Char Acc~$\uparrow$ & Char F1~$\uparrow$ \\
\midrule[0.1em]
ReCap w/o \drift & 28.15 & 91.04 & 95.91 \\
ReCap w/ CLIP~\citep{radford2021learning} & 26.24 & \textbf{92.10} & \textbf{96.30} \\
ReCap w/ DINOv3~\citep{Simeoni2025DINOv3} (Ours) & \textbf{24.72} & 91.86 & 96.07 \\
\bottomrule[0.1em]
\end{tabular}}
\vspace{-1.5em}
\end{table}

Table~\ref{table:lambda_ablation} examines the effect of regularization  $\lambda_{\text{reg}}$ on background consistency. Background metrics peak at $\lambda_{\text{reg}}=0.5$ (BG Acc: 59.74\%, BG F1: 64.95\%),
validating the balance between denoising fidelity and semantic anchoring
introduced in Eq.~\ref{eq:total_loss}. Values that are too low (0.01, 0.1)
provide insufficient semantic guidance from the feature space,
as the cosine similarity loss in Eq.~\ref{eq:dinov3_loss} has too little
influence on the denoiser representations. Very high values (1.0)
over-constrain the total loss toward semantic anchoring at the expense
of the flow-matching denoising objective $\mathcal{L}_{\text{denoise}}$,
slightly degrading background consistency.

Table~\ref{table:cfg_ablation} analyzes the effect of classifier-free
guidance scale on image quality and character consistency. CFG$=5.0$ yields
the best tradeoff, achieving the lowest FID (24.72) alongside strong character
accuracy (91.86\% Char-Acc, 96.07\% Char-F1). Lower CFG (1.0) reduces the
influence of the text conditioning in the SD3 cross-attention layers,
harming identity fidelity (FID: 35.56, Char-Acc: 74.47\%). Higher CFG (7.5)
marginally improves character accuracy (92.09\%) at the cost of worse FID
(26.22), suggesting that overly strong text conditioning conflicts with
the semantic stability encouraged by \drift during training.

Table~\ref{tab:backbone_comparison} compares CLIP~\citep{radford2021learning}
and DINOv3~\citep{Simeoni2025DINOv3} as visual backbones for \drift. Both improve over the baseline, confirming that feature-level
regularization is effective regardless of the specific visual encoder.
DINOv3 achieves better FID (24.72 vs 26.24), suggesting that its dense,
patch-level self-supervised features are better suited for maintaining
fine-grained character identity across frames than CLIP's globally aligned
image-text features. We adopt DINOv3 as our default backbone; a more
detailed analysis is provided in the supplementary material.

\vspace{-1.0em}
\begin{figure*}
    \centering
    \includegraphics[width=0.7\linewidth]{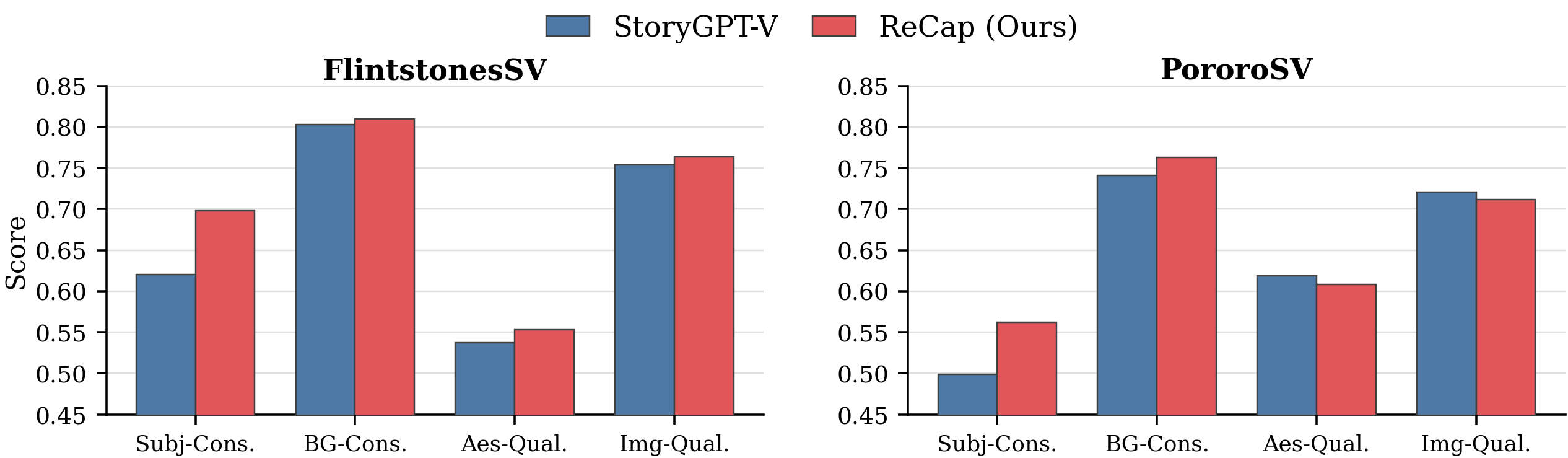}
    \caption{Visual quality and consistency FlintstonesSV~\citep{gupta2018imagine} and PororoSV~\citep{li2019storygan} datasets, evaluated using 
VBench~\citep{vbench}. Subj-Cons. = Subject Consistency, 
BG-Cons. = Background Consistency, Aes-Qual. = Aesthetic Quality, 
Img-Qual. = Imaging Quality. Higher scores are better for all metrics.}
    \label{fig:vbench_story} \vspace{-2.5em}
\end{figure*}

\subsection{Visual Quality Evaluation: VBench}
Figure~\ref{fig:vbench_story} reports VBench metrics on FlintstonesSV
and PororoSV. VBench evaluates temporal coherence through Subject
Consistency, which measures how stable character appearance remains
across frames, and Background Consistency, which captures scene-level
stability. Aesthetic Quality and Imaging Quality measure perceptual
realism of individual frames. On FlintstonesSV, ReCap outperforms
StoryGPT-V across all four dimensions, with the largest gain in
Subject Consistency by 7.8\% (0.620 $\to$ 0.698), confirming that
\core's selective conditioning improves character-level identity
stability across frames without sacrificing perceptual image quality.
On PororoSV, ReCap improves Subject Consistency by 6.30\% (0.499 $\to$ 0.562) and Background Consistency by 2.20\% (0.741 $\to$ 0.763),
while StoryGPT-V achieves marginally better Aesthetic Quality and
Imaging Quality (0.619 vs 0.608 and 0.721 vs 0.712).

\newcommand{\qualfigsize}{0.4}
\newcommand{\qualtxtindent}{3.8em}
\newcommand{\qualtxtwidth}{0.92\linewidth}

\begin{figure*}
\centering
    \hspace{-12mm}
    \begin{minipage}[t]{\qualfigsize\linewidth}
        \centering
        \setlength{\tabcolsep}{1pt}
        \renewcommand{\arraystretch}{0}
        \begin{tabular}{ccccc}

        {\tiny \raisebox{5ex}{\shortstack{SD3~\citep{esser2024scaling}}}} &
        \includegraphics[width=.23\textwidth,clip,trim=0    0 1152 1152]{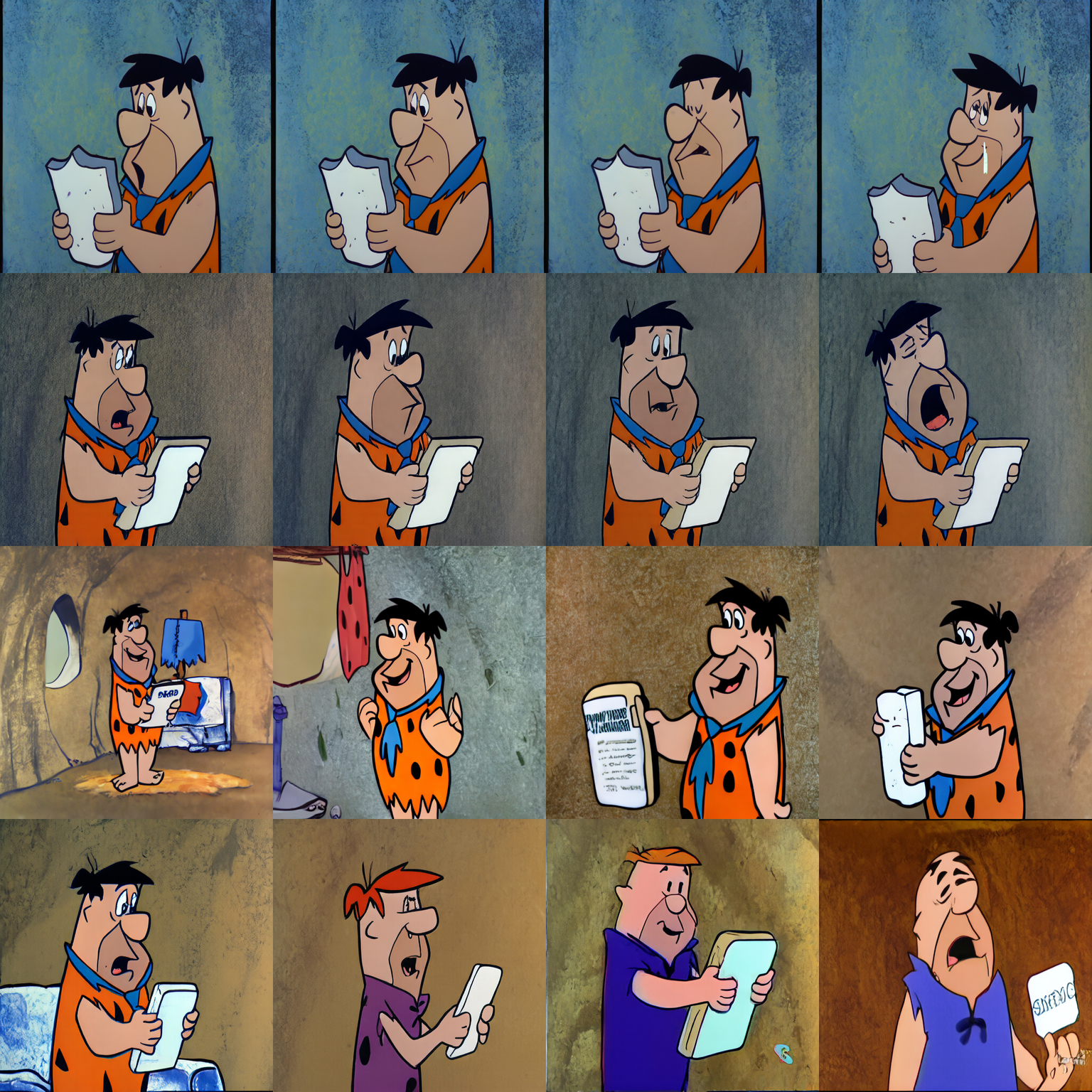} &
        \includegraphics[width=.23\textwidth,clip,trim=384  0 768  1152]{Figures/visual1c.jpg} &
        \includegraphics[width=.23\textwidth,clip,trim=768  0 384  1152]{Figures/visual1c.jpg} &
        \includegraphics[width=.23\textwidth,clip,trim=1152 0 0    1152]{Figures/visual1c.jpg} \\[1mm]

        {\tiny \raisebox{5ex}{\shortstack{Story~\citep{shen2025storygptv}\\GPT-V}}} &
        \includegraphics[width=.23\textwidth,clip,trim=0    384 1152 768 ]{Figures/visual1c.jpg} &
        \includegraphics[width=.23\textwidth,clip,trim=384  384 768  768 ]{Figures/visual1c.jpg} &
        \includegraphics[width=.23\textwidth,clip,trim=768  384 384  768 ]{Figures/visual1c.jpg} &
        \includegraphics[width=.23\textwidth,clip,trim=1152 384 0    768 ]{Figures/visual1c.jpg} \\[1mm]

        {\tiny \raisebox{5ex}{\shortstack{ReCap\\(Ours)}}} &
        \includegraphics[width=.23\textwidth,clip,trim=0    768 1152 384 ]{Figures/visual1c.jpg} &
        \includegraphics[width=.23\textwidth,clip,trim=384  768 768  384 ]{Figures/visual1c.jpg} &
        \includegraphics[width=.23\textwidth,clip,trim=768  768 384  384 ]{Figures/visual1c.jpg} &
        \includegraphics[width=.23\textwidth,clip,trim=1152 768 0    384 ]{Figures/visual1c.jpg} \\[1mm]

        {\tiny \raisebox{5ex}{\shortstack{Ground\\Truth}}} &
        \includegraphics[width=.23\textwidth,clip,trim=0    1152 1152 0]{Figures/visual1c.jpg} &
        \includegraphics[width=.23\textwidth,clip,trim=384  1152 768  0]{Figures/visual1c.jpg} &
        \includegraphics[width=.23\textwidth,clip,trim=768  1152 384  0]{Figures/visual1c.jpg} &
        \includegraphics[width=.23\textwidth,clip,trim=1152 1152 0    0]{Figures/visual1c.jpg} \\[1mm]

        & \colhead{\tiny (1)}
        & \colhead{\tiny (2)}
        & \colhead{\tiny (3)}
        & \colhead{\tiny (4)} \\[1mm]

    \end{tabular}
        \vspace{0.4em}
        \hspace*{\qualtxtindent}
        \parbox[t]{\qualtxtwidth}{\tiny\raggedright
        \begin{enumerate}[label=(\arabic*),leftmargin=1.6em,itemsep=0pt,topsep=0pt,parsep=0pt,partopsep=0pt]
            \item Fred is standing in the living room reading a plaque. He is shocked.
            \item \textcolor{magenta}{He} reads an urgent note in the room.
            \item \textcolor{magenta}{He} is reading from a stone tablet in his room.
            \item \textcolor{magenta}{He} is in a room holding a piece of slate. He is crying.
        \end{enumerate}}

	    \end{minipage}
    \quad \quad \quad \quad
    \begin{minipage}[t]{\qualfigsize\linewidth}
        \centering
        \setlength{\tabcolsep}{1pt}
        \renewcommand{\arraystretch}{0}
        \begin{tabular}{ccccc}

        {\tiny \raisebox{5ex}{\shortstack{SD3~\citep{esser2024scaling}}}} &
        \includegraphics[width=.23\textwidth,clip,trim=0    0 1152 1152]{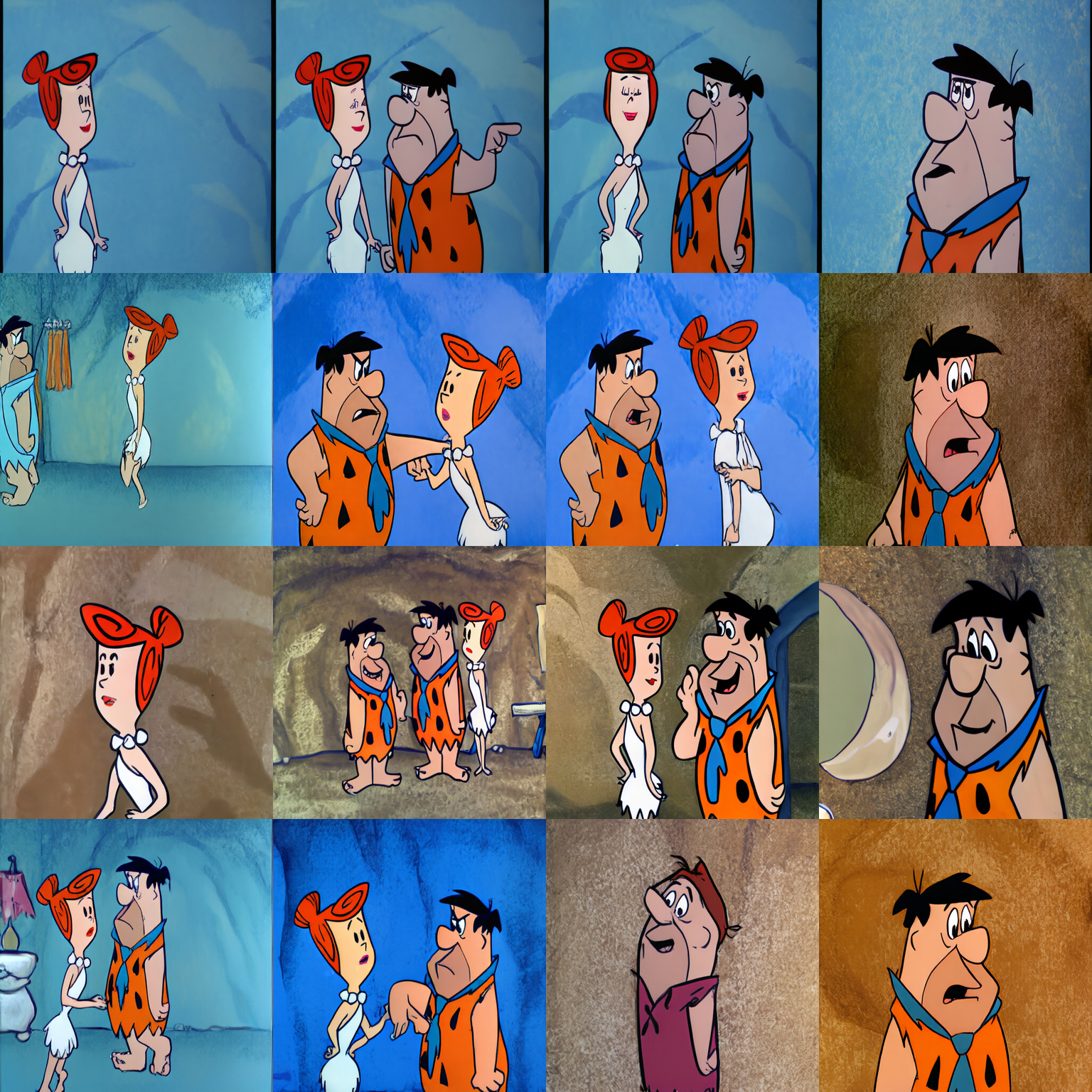} &
        \includegraphics[width=.23\textwidth,clip,trim=384  0 768  1152]{Figures/visual4c.jpg} &
        \includegraphics[width=.23\textwidth,clip,trim=768  0 384  1152]{Figures/visual4c.jpg} &
        \includegraphics[width=.23\textwidth,clip,trim=1152 0 0    1152]{Figures/visual4c.jpg} \\[1mm]

        {\tiny \raisebox{5ex}{\shortstack{Story~\citep{shen2025storygptv}\\GPT-V}}} &
        \includegraphics[width=.23\textwidth,clip,trim=0    384 1152 768 ]{Figures/visual4c.jpg} &
        \includegraphics[width=.23\textwidth,clip,trim=384  384 768  768 ]{Figures/visual4c.jpg} &
        \includegraphics[width=.23\textwidth,clip,trim=768  384 384  768 ]{Figures/visual4c.jpg} &
        \includegraphics[width=.23\textwidth,clip,trim=1152 384 0    768 ]{Figures/visual4c.jpg} \\[1mm]

        {\tiny \raisebox{5ex}{\shortstack{ReCap\\(Ours)}}} &
        \includegraphics[width=.23\textwidth,clip,trim=0    768 1152 384 ]{Figures/visual4c.jpg} &
        \includegraphics[width=.23\textwidth,clip,trim=384  768 768  384 ]{Figures/visual4c.jpg} &
        \includegraphics[width=.23\textwidth,clip,trim=768  768 384  384 ]{Figures/visual4c.jpg} &
        \includegraphics[width=.23\textwidth,clip,trim=1152 768 0    384 ]{Figures/visual4c.jpg} \\[1mm]

        {\tiny \raisebox{5ex}{\shortstack{Ground\\Truth}}} &
        \includegraphics[width=.23\textwidth,clip,trim=0    1152 1152 0]{Figures/visual4c.jpg} &
        \includegraphics[width=.23\textwidth,clip,trim=384  1152 768  0]{Figures/visual4c.jpg} &
        \includegraphics[width=.23\textwidth,clip,trim=768  1152 384  0]{Figures/visual4c.jpg} &
        \includegraphics[width=.23\textwidth,clip,trim=1152 1152 0    0]{Figures/visual4c.jpg} \\[1mm]

        & \colhead{\tiny (1)}
        & \colhead{\tiny (2)}
        & \colhead{\tiny (3)}
        & \colhead{\tiny (4)} \\[1mm]

    \end{tabular}
        \vspace{0.4em}
        \hspace*{\qualtxtindent}
        \parbox[t]{\qualtxtwidth}{\tiny\raggedright
        \begin{enumerate}[label=(\arabic*),leftmargin=1.6em,itemsep=0pt,topsep=0pt,parsep=0pt,partopsep=0pt]
            \item Wilma is in a white dress in a blue room and speaks towards Fred.\\
            \item \textcolor{magenta}{They} are standing in a blue room arguing.\\
            \item \textcolor{magenta}{They} are standing in a blue room talking. \\
            \item Fred is talking to himself in a room.
        \end{enumerate}}

	    \end{minipage}

    \vspace{0.4em}
    \hspace{-12mm}
	     \begin{minipage}[t]{\qualfigsize\linewidth}
        \centering
        \setlength{\tabcolsep}{1pt}
        \renewcommand{\arraystretch}{0}
        \begin{tabular}{ccccc}

        {\tiny \raisebox{5ex}{\shortstack{Story~\citep{Maharana2022StoryDALLE}\\DALL-E}}} &
        \includegraphics[width=.23\textwidth,clip,trim=0    0 1152 1152]{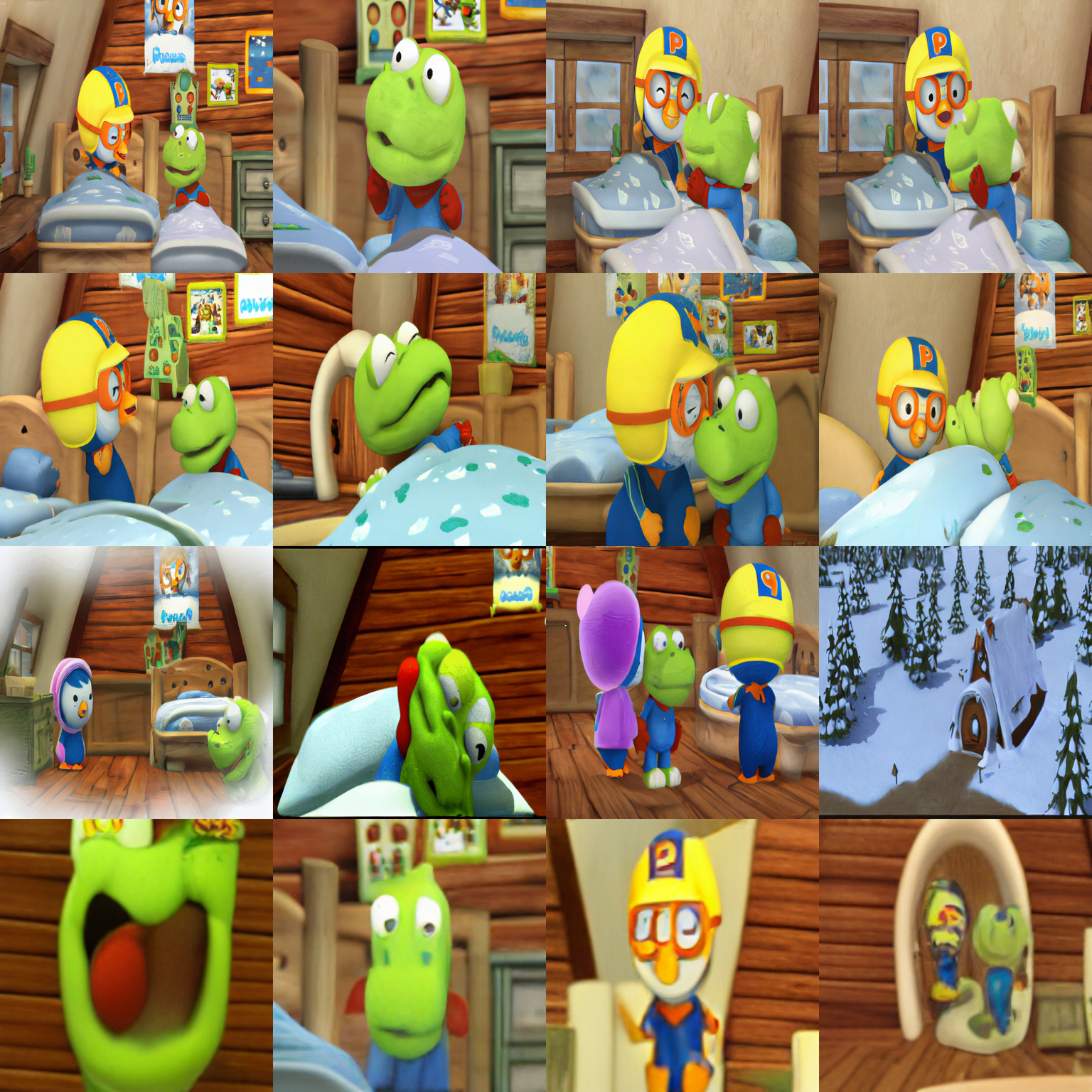} &
        \includegraphics[width=.23\textwidth,clip,trim=384  0 768  1152]{Figures/visual14b.jpg} &
        \includegraphics[width=.23\textwidth,clip,trim=768  0 384  1152]{Figures/visual14b.jpg} &
        \includegraphics[width=.23\textwidth,clip,trim=1152 0 0    1152]{Figures/visual14b.jpg} \\[1mm]

        {\tiny \raisebox{5ex}{\shortstack{Story~\citep{shen2025storygptv}\\GPT-V}}} &
        \includegraphics[width=.23\textwidth,clip,trim=0    384 1152 768 ]{Figures/visual14b.jpg} &
        \includegraphics[width=.23\textwidth,clip,trim=384  384 768  768 ]{Figures/visual14b.jpg} &
        \includegraphics[width=.23\textwidth,clip,trim=768  384 384  768 ]{Figures/visual14b.jpg} &
        \includegraphics[width=.23\textwidth,clip,trim=1152 384 0    768 ]{Figures/visual14b.jpg} \\[1mm]

        {\tiny \raisebox{5ex}{\shortstack{ReCap\\(Ours)}}} &
        \includegraphics[width=.23\textwidth,clip,trim=0    768 1152 384 ]{Figures/visual14b.jpg} &
        \includegraphics[width=.23\textwidth,clip,trim=384  768 768  384 ]{Figures/visual14b.jpg} &
        \includegraphics[width=.23\textwidth,clip,trim=768  768 384  384 ]{Figures/visual14b.jpg} &
        \includegraphics[width=.23\textwidth,clip,trim=1152 768 0    384 ]{Figures/visual14b.jpg} \\[1mm]

        {\tiny \raisebox{5ex}{\shortstack{Ground\\Truth}}} &
        \includegraphics[width=.23\textwidth,clip,trim=0    1152 1152 0]{Figures/visual14b.jpg} &
        \includegraphics[width=.23\textwidth,clip,trim=384  1152 768  0]{Figures/visual14b.jpg} &
        \includegraphics[width=.23\textwidth,clip,trim=768  1152 384  0]{Figures/visual14b.jpg} &
        \includegraphics[width=.23\textwidth,clip,trim=1152 1152 0    0]{Figures/visual14b.jpg} \\[1mm]

        & \colhead{\tiny (1)}
        & \colhead{\tiny (2)}
        & \colhead{\tiny (3)}
        & \colhead{\tiny (4)} \\[1mm]

    \end{tabular}
        \vspace{0.4em}
        \hspace*{\qualtxtindent}
        \parbox[t]{\qualtxtwidth}{\tiny\raggedright
        \begin{enumerate}[label=(\arabic*),leftmargin=1.6em,itemsep=0pt,topsep=0pt,parsep=0pt,partopsep=0pt]
            \item Pororo and Crong are in the beds.
            \item Crong looks very happy and is smiling in the bed.
            \item Pororo and Crong are having fun.
            \item \textcolor{magenta}{They} are getting ready to go to bed.
        \end{enumerate}}
	    \end{minipage}
	    \quad \quad \quad \quad
	    \begin{minipage}[t]{\qualfigsize\linewidth}
	        \centering
	        \setlength{\tabcolsep}{1pt}
	        \renewcommand{\arraystretch}{0}
	        \begin{tabular}{ccccc}

	        {\tiny \raisebox{5ex}{\shortstack{Story~\citep{Maharana2022StoryDALLE}\\DALL-E}}} &
	        \includegraphics[width=.23\textwidth,clip,trim=0    0 1152 1152]{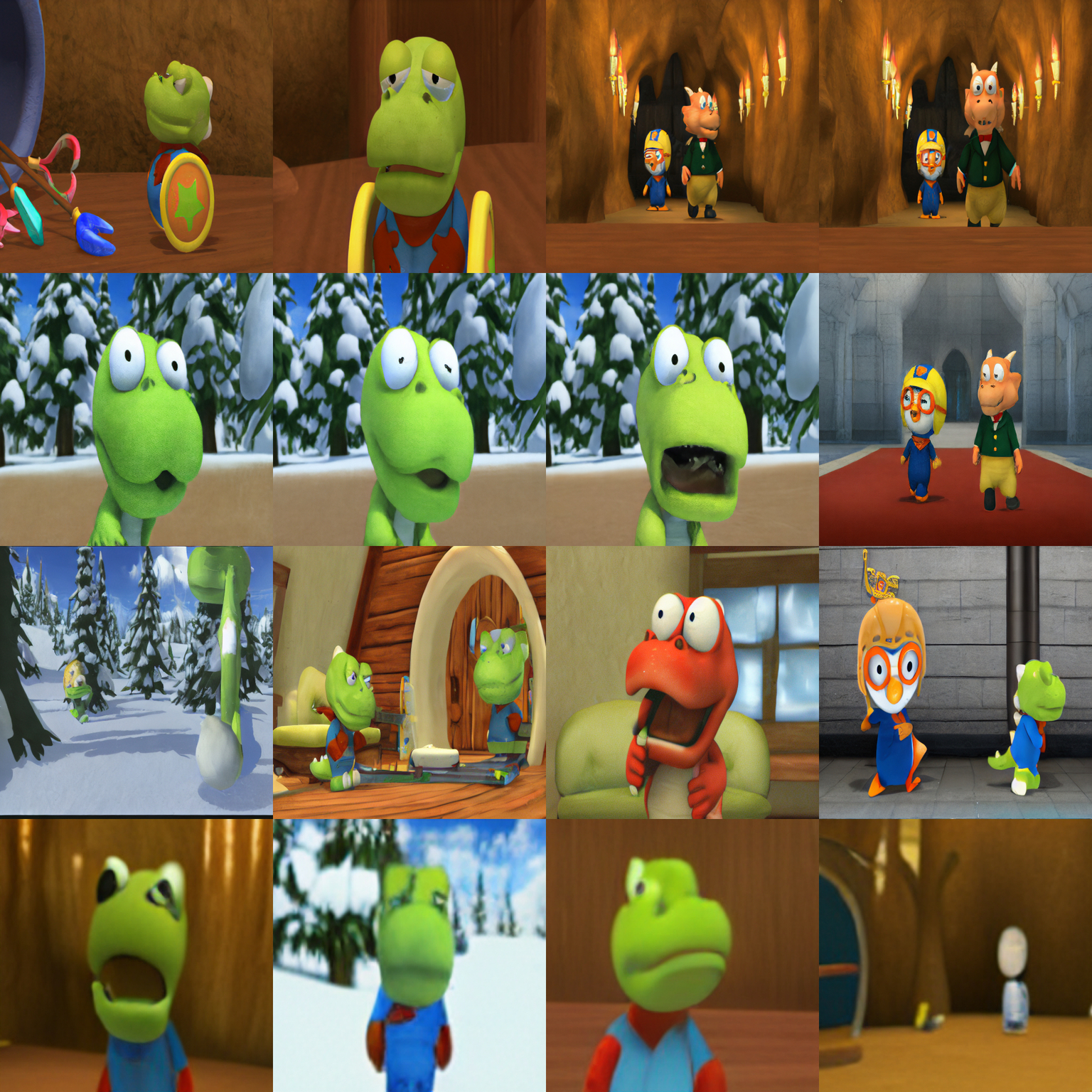} &
	        \includegraphics[width=.23\textwidth,clip,trim=384  0 768  1152]{Figures/visual15b.jpg} &
	        \includegraphics[width=.23\textwidth,clip,trim=768  0 384  1152]{Figures/visual15b.jpg} &
	        \includegraphics[width=.23\textwidth,clip,trim=1152 0 0    1152]{Figures/visual15b.jpg} \\[1mm]

	        {\tiny \raisebox{5ex}{\shortstack{Story~\citep{shen2025storygptv}\\GPT-V}}} &
	        \includegraphics[width=.23\textwidth,clip,trim=0    384 1152 768 ]{Figures/visual15b.jpg} &
	        \includegraphics[width=.23\textwidth,clip,trim=384  384 768  768 ]{Figures/visual15b.jpg} &
	        \includegraphics[width=.23\textwidth,clip,trim=768  384 384  768 ]{Figures/visual15b.jpg} &
	        \includegraphics[width=.23\textwidth,clip,trim=1152 384 0    768 ]{Figures/visual15b.jpg} \\[1mm]

	        {\tiny \raisebox{5ex}{\shortstack{ReCap\\(Ours)}}} &
	        \includegraphics[width=.23\textwidth,clip,trim=0    768 1152 384 ]{Figures/visual15b.jpg} &
	        \includegraphics[width=.23\textwidth,clip,trim=384  768 768  384 ]{Figures/visual15b.jpg} &
	        \includegraphics[width=.23\textwidth,clip,trim=768  768 384  384 ]{Figures/visual15b.jpg} &
	        \includegraphics[width=.23\textwidth,clip,trim=1152 768 0    384 ]{Figures/visual15b.jpg} \\[1mm]

	        {\tiny \raisebox{5ex}{\shortstack{Ground\\Truth}}} &
	        \includegraphics[width=.23\textwidth,clip,trim=0    1152 1152 0]{Figures/visual15b.jpg} &
	        \includegraphics[width=.23\textwidth,clip,trim=384  1152 768  0]{Figures/visual15b.jpg} &
	        \includegraphics[width=.23\textwidth,clip,trim=768  1152 384  0]{Figures/visual15b.jpg} &
	        \includegraphics[width=.23\textwidth,clip,trim=1152 1152 0    0]{Figures/visual15b.jpg} \\[1mm]

	        & \colhead{\tiny (1)}
	        & \colhead{\tiny (2)}
	        & \colhead{\tiny (3)}
	        & \colhead{\tiny (4)} \\[1mm]

	    \end{tabular}
	        \vspace{0.4em}
	        \hspace*{\qualtxtindent}
	        \parbox[t]{\qualtxtwidth}{\tiny\raggedright
	        \begin{enumerate}[label=(\arabic*),leftmargin=1.6em,itemsep=0pt,topsep=0pt,parsep=0pt,partopsep=0pt]
	            \item Crong swings his head up and down.
	            \item \textcolor{magenta}{He} makes a sad face.
	            \item \textcolor{magenta}{He} is crying with mouth opened.
	            \item Pororo and Tongtong are walking in the corridor.
	        \end{enumerate}}
	    \end{minipage}

    \caption{Qualitative comparison on FlintstonesSV~\citep{gupta2018imagine}
    (top) and PororoSV~\citep{li2019storygan} (bottom). Pink tokens indicate
    anaphoric references (pronouns) in the narrative. On FlintstonesSV (top
    left), SD3 fails to preserve character appearance across pronoun-containing
    frames (2--4), generating inconsistent clothing and visual features, while
    ReCap closely matches the ground truth throughout. On FlintstonesSV (top
    right), SD3 fails to maintain characters, background, and scene throughout
    the sequence. StoryGPT-V partially recovers appearance in some frames but
    introduces incorrect characters and background changes in others. ReCap
    correctly resolves the pronoun ``They'' in frames 2--3, preserving both
    characters and background. On PororoSV (bottom), StoryGPT-V struggles
    under pronoun reference in frame 4, and
    StoryDALL-E~\citep{Maharana2022StoryDALLE} fails to maintain Crong's
    appearance; ReCap preserves both Pororo and Crong's visual identity and
    scene throughout.}
    \label{fig:qualitative3}
\end{figure*}

\subsection{Qualitative Results}

Figure~\ref{fig:qualitative3} shows representative generations on
\textit{FlintstonesSV}~\citep{gupta2018imagine} and
\textit{PororoSV}~\citep{li2019storygan} against SD3~\citep{esser2024scaling}, Story DALL-E~\citep{Maharana2022StoryDALLE}
and StoryGPT-V~\citep{shen2025storygptv}. The failure modes visible in
the baselines reflect the absence of referential grounding: without a
mechanism to resolve pronouns to previously established visual entities,
models default to generating characters independently per frame, causing
drift in hair color, clothing, and accessories. ReCap maintains visual
coherence across pronoun-containing frames, with stable color, texture,
and character appearance throughout the sequence, and most accurately
follows the story text provided beneath each example compared to all
other methods.

\begin{figure*}[!t]
\centering
\setlength{\tabcolsep}{1.2pt}
\renewcommand{\arraystretch}{0}
\begin{tabular}{c c c c c c}
{\tiny \raisebox{4.5ex}{\shortstack{SD3\\(baseline)}}} &
\includegraphics[width=.2\textwidth]{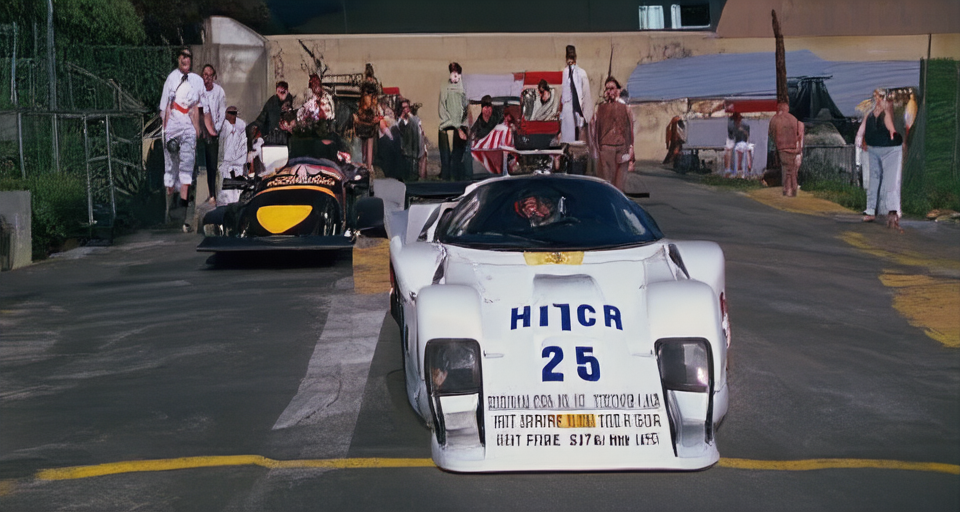} &
\includegraphics[width=.2\textwidth]{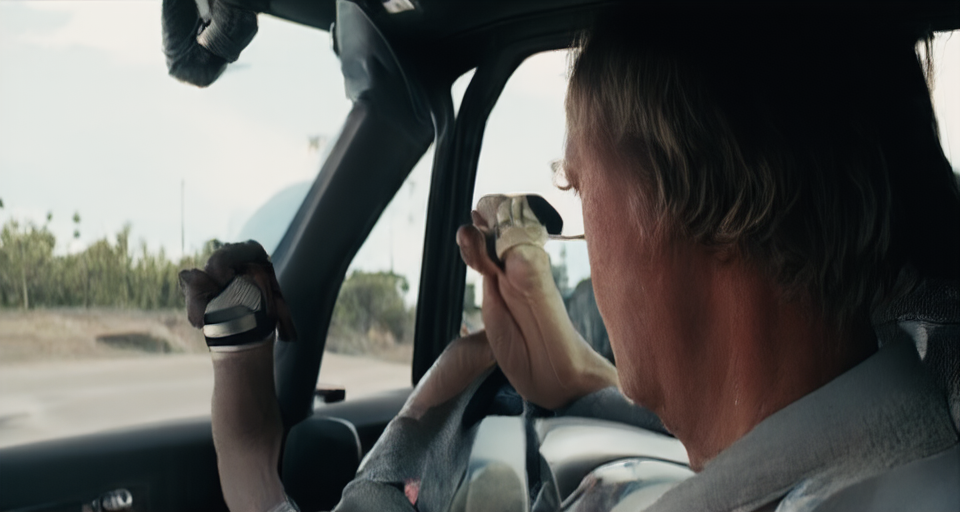} &
\includegraphics[width=.2\textwidth]{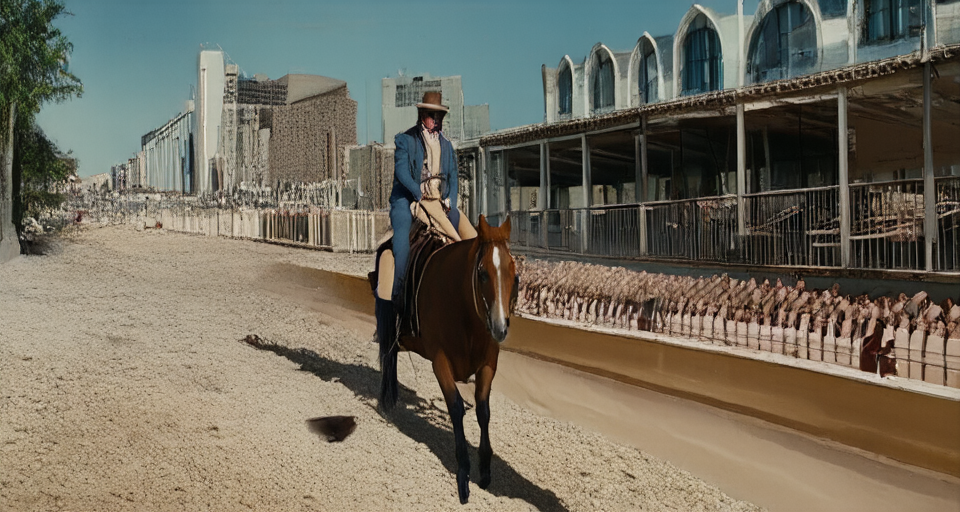} &
\includegraphics[width=.2\textwidth]{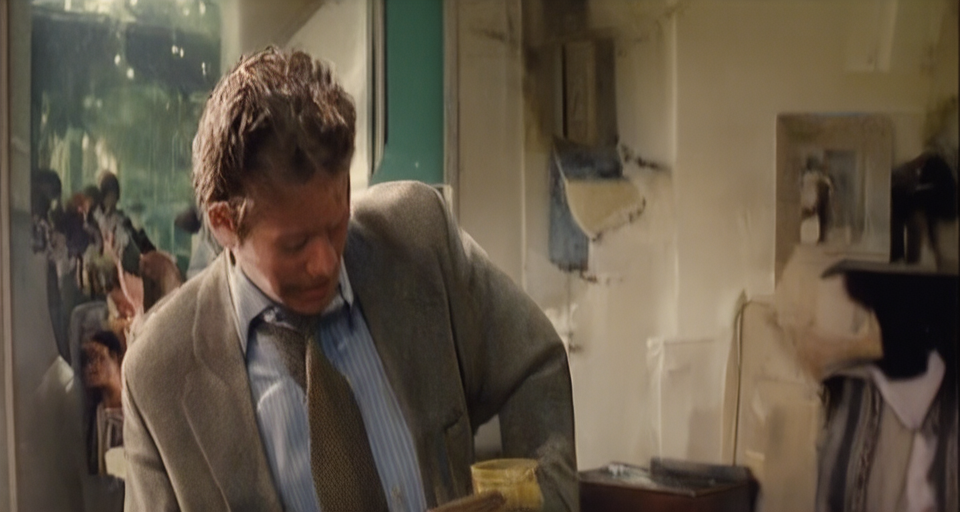} \\[1.5mm]

{\tiny \raisebox{4.5ex}{\shortstack{ReCap\\(Ours)}}} &
\includegraphics[width=.2\textwidth]{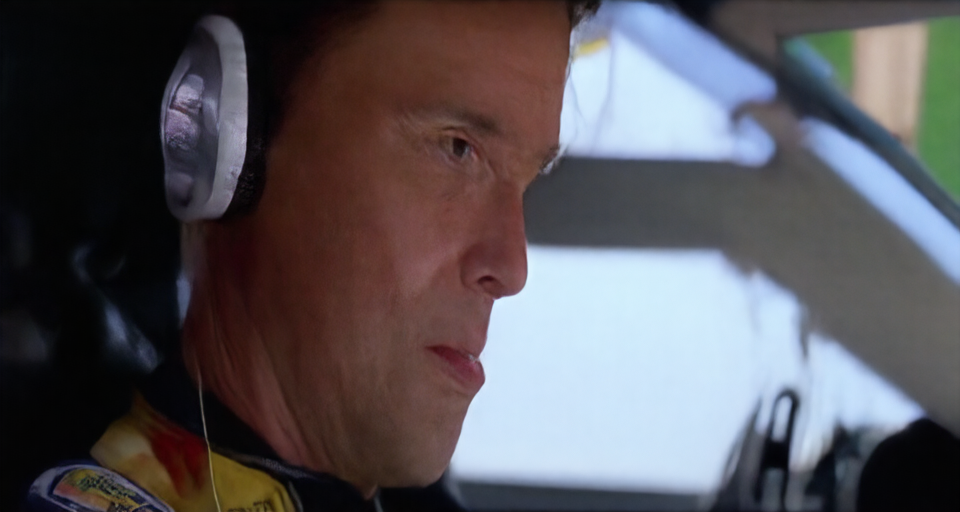} &
\includegraphics[width=.2\textwidth]{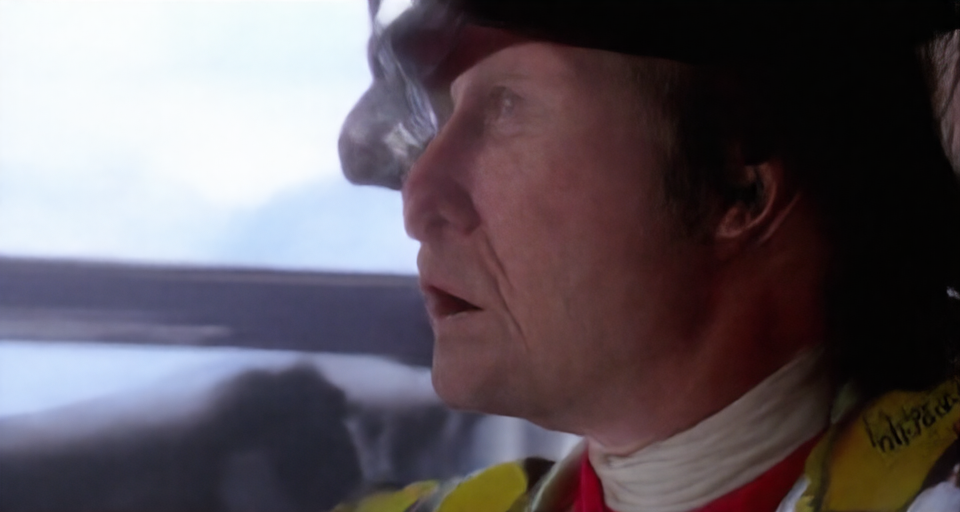} &
\includegraphics[width=.2\textwidth]{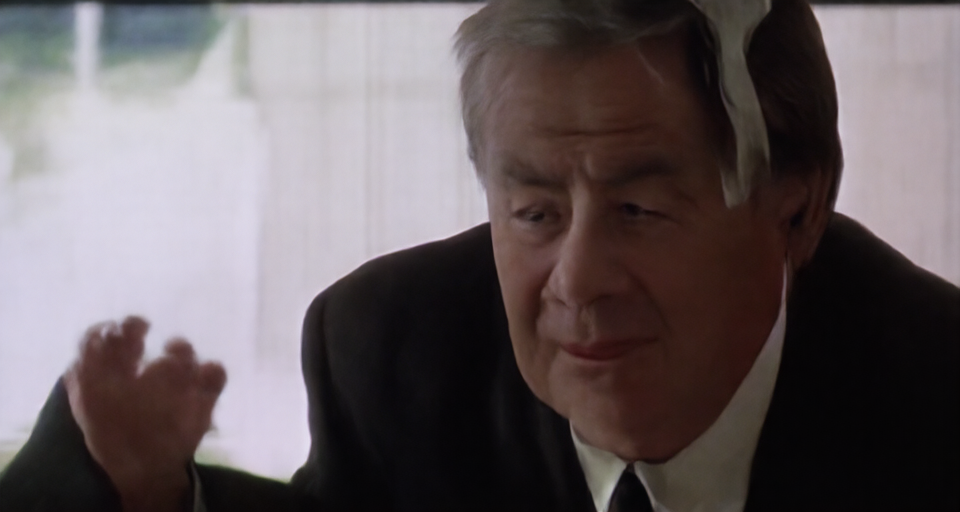} &
\includegraphics[width=.2\textwidth]{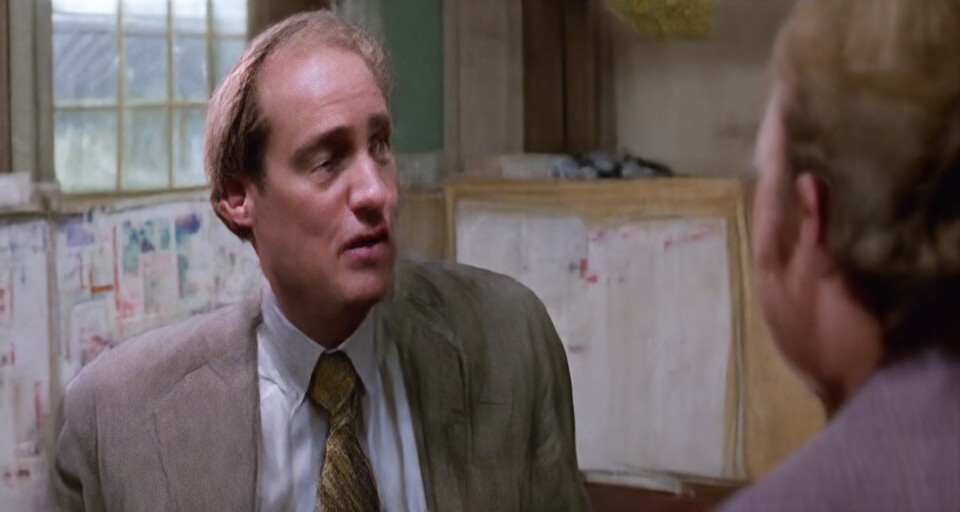} \\[0.8mm]

\end{tabular}
\vspace{0.25em}
\begin{tabular}{p{.08\textwidth}p{.2\textwidth}p{.2\textwidth}p{.2\textwidth}p{.2\textwidth}}
&
\raggedright\tiny [male0] was a professional race car driver who was very demanding &
\raggedright\tiny He asked the race owner if he would give him a raise &
\raggedright\tiny the race owner did n't think [male0] needed a raise at the moment &
\raggedright\tiny \textcolor{magenta}{He} was bitter that he would n't be able to get the raise\\
\end{tabular}

\caption{Qualitative comparison on VWP~\citep{hong2023visual}, featuring a naturalistic narrative about a race car driver referred to by pronouns. In frame 2, the story uses ``he'' but we deliberately keep CORE off to isolate whether SemDrift regularization alone preserves similar scene cues (for example, the collar in the foreground and the car door in the background). In frame 4, the story again uses ``he'' and we keep CORE on (the default behavior across scenes), showing that scene contents such as the suit and tie are generated consistently across scenes even when they are not explicitly stated in the story text.}
\label{fig:vwp_qualitative}
\end{figure*}

\begin{table*}[t]
\centering
\caption{Evaluation on VWP~\citep{hong2023visual} using 
VBench~\citep{vbench} metrics. Ground Truth represents the upper bound. 
Higher Subject Consistency and Background Consistency are better.}
\label{tab:vbench_vwp}
\setlength{\tabcolsep}{6pt}
\scalebox{0.75}{
\begin{tabular}{ccccc}
\toprule[0.15em]
\textbf{SemDrift} & \textbf{CORE} & \textbf{Method} & \textbf{Subject Consistency}~$\uparrow$ & \textbf{Background Consistency}~$\uparrow$\\
\midrule[0.15em]
\xmark & \xmark &\multirow{3}{*}{\shortstack{Baseline\\(SD3~\citep{esser2024scaling})}} & 0.317 & 0.623\\
\cmark & \xmark &  & 0.307 & 0.624\\
\xmark & \cmark &  & 0.327 & 0.610\\
\cmark & \cmark & \textbf{ReCap (Ours)} &\textbf{0.346} & \textbf{0.646} \\
\midrule[0.05em]
- & - & Ground Truth & 0.460 & 0.732 \\
\bottomrule[0.15em]
\end{tabular}}
\end{table*}

\subsection{Human Story Visualization}

Table~\ref{tab:vbench_vwp} reports two VBench consistency metrics:
\textit{Subject Consistency} measures how well the main subject's
appearance is preserved across frames, while \textit{Background
Consistency} measures temporal stability of scene layout and
environmental context. Higher values indicate better temporal
coherence for both metrics. The Ground Truth row provides an upper
bound computed from real frame sequences, and the SemDrift/CORE
combinations isolate the contribution of each component. To adapt
VWP for referential story generation, we anonymize character names and
apply pronoun replacement following the same protocol as the cartoon
benchmarks. As the original VWP images are low resolution, we apply
super-resolution and resize to $960\times512$, which preserves the
aspect ratio of the majority of film frames. We train ReCap on VWP
with no dataset-specific architectural modifications, demonstrating
that the method generalizes beyond stylized cartoon domains.
Quantitatively (Table~\ref{tab:vbench_vwp}), ReCap improves over SD3 on
both reported VBench consistency metrics (Subj-Cons.\ $0.317{\to}0.346$
and BG-Cons.\ $0.623{\to}0.646$), corresponding to gains of 9.15\% and
3.69\%, confirming that referential grounding transfers to
naturalistic, pronoun-heavy narratives. Qualitatively, as shown in Figure~\ref{fig:vwp_qualitative}, the narrative follows a professional race car driver making a demand for a raise. In frame 2, the story uses ``he'' but we deliberately keep CORE off to test whether SemDrift regularization alone preserves similar content cues (for example, the collar in the foreground and the car door in the background). In frame 4, the story again uses ``he'' and CORE is turned on (the default behavior across scenes), which shows that scene contents such as the suit and tie are generated consistently across scenes even when they are not explicitly stated in the story text.

\section{Conclusion}
\label{sec:conclusion}
\vspace{-2pt}
We presented ReCap, a lightweight diffusion framework for
story visualization that preserves character identity and semantic
coherence under referential text. \core provides efficient cross-frame
grounding by conditioning generation on the preceding frame only when
the narrative references a character anaphorically, adding only a
fraction of parameters to the base SD3 model. Combined with \drift,
which aligns intermediate denoiser representations with pretrained
DINOv3 embeddings during training at zero inference cost, ReCap achieves
state-of-the-art character consistency on both FlintstonesSV and
PororoSV. Experiments further demonstrate that ReCap generalizes to
naturalistic, pronoun-heavy narratives derived from real films,
confirming that referential grounding is not domain-specific. This work
highlights that coherence and identity persistence in story visualization
do not require heavy recurrent or LLM-based modules; instead, they can
emerge from well-designed lightweight attention mechanisms. Future work
will explore extending ReCap to multi-character interaction modeling,
longer narrative generation, and integrating automatic pronoun detection
into the generation loop for real-time narrative synthesis.

\section{Acknowledgment}
\label{sec:ack}

The research at TU Darmstadt was partially funded by a LOEWE-Spitzen-Professur (LOEWE/4a//519/05.00.002(0010)/93), and an Alexander von Humboldt Professorship in Multimodal Reliable AI, sponsored by Germany’s Federal Ministry of Research, Technology and Space.

\noindent For compute, we gratefully acknowledge support from the hessian.AI Service Center (funded by the Federal Ministry of Research, Technology and Space, BMFTR, grant no. 16IS22091) and the hessian.AI Innovation Lab (funded by the Hessian Ministry for Digital Strategy and Innovation, grant no. S-DIW04/0013/003).


%
%
\bibliographystyle{splncs04}
\bibliography{main}

\clearpage
\appendix
\newcommand{\supplementtitle}[1]{
    \clearpage
    \begingroup
    \centering
    \vspace*{0.5cm}
    {\LARGE \bfseries #1 \par}
    \vspace{1.5cm}
    \endgroup
}
\supplementtitle{ReCap: Lightweight Referential Grounding for Coherent Story Visualization\\(Supplementary Material)}

In this supplementary material, we provide additional details and results to support the main paper. We present additional details about the re-purposed VWP dataset~\citep{hong2023visual} (\cref{sec:vwp_info}), additional architectural details of the CORE (\cref{sec:recap_details}) and SemDrift (\cref{sec:why_dinov3}) modules, and experiment details (\cref{sec:exp_details}) along with quantitative and qualitative results demonstrating the effectiveness of our proposed method (\cref{sec:qualitative}).

\section{Visual Writing Prompts (VWP) dataset}
\label{sec:vwp_info}

We introduce a new benchmark setting for human story visualization using the VWP dataset~\citep{hong2023visual}, which is substantially more challenging than cartoon-based story visualization due to more intricate scenes and complex character interactions. In addition, multiple characters often appear in the same scene, while the story-text does not explicitly localize which individual is being referred to.
Originally, the VWP dataset~\citep{hong2023visual} was introduced for image-grounded story generation rather than story visualization, and therefore presents a different evaluation setting. Stories are longer and often contain abstract narrative descriptions with limited visual grounding. Characters are typically referred to using placeholders such as \texttt{[male0]} or \texttt{[female0]}, and the text does not always clearly indicate which character is being referenced.
To reduce reliance on explicit character identifiers, we apply a pronoun replacement strategy to the story-text, replacing adjacent placeholder mentions with corresponding pronouns (``he'' for male characters, ``she'' for female characters, and ``they'' when multiple characters appear together across consecutive frames). This further increases the difficulty of the task, as the model cannot rely on a fixed character bank and must instead learn more general visual representations for different characters and interactions.

In addition, frames frequently contain multiple characters and complex interactions. These factors make maintaining identity, grounding the correct character, and preserving narrative coherence across frames more challenging. As shown in Figs.~\ref{fig:supp3a}--\ref{fig:supp3d}, our method produces visually more coherent sequences that better follow the story progression.

\section{CORE: COntextual frame REferencing}
\label{sec:recap_details}

As mentioned in Section 3.2 in the main text, the CORE module encodes the previous frame into a compact contextual representation that acts as guidance for the cross-attention projections in the diffusion transformer. The architecture consists of four main components:
\vspace{2mm}

\noindent \textbf{Feature Extraction.} A $\mathrm{Conv}\,3{\times}3$ layer projects the previous frame from image space into an intermediate feature representation.

\vspace{2mm}

\noindent \textbf{Guidance Attention Blocks.} We employ six \emph{Guidance Attention Blocks} (GABs) arranged in a residual-in-residual configuration~\cite{Wang2018ESRGAN}. This design enables deeper feature refinement while maintaining stable gradients. Each GAB contains:
\begin{itemize}
    \item A \emph{Channel Attention Module} that emphasizes feature channels relevant to identity-defining cues such as facial structure and clothing patterns~\cite{Woo2018CBAM}.
    \item A \emph{Spatial Attention Module} that highlights spatially localized features related to pose, silhouette, and region-level consistency~\cite{Woo2018CBAM}.
\end{itemize}
\vspace{2mm}

\noindent \textbf{Resolution Alignment.} To match the spatial resolution of the transformer tokens, the features are downsampled by a factor of $8{\times}$ using repeated bilinear downsampling followed by $\mathrm{Conv}\,1{\times}1$ layers. This produces a $32 \times 32$ feature map corresponding to $1024$ spatial locations.

\vspace{2mm}

\noindent \textbf{Projection Conv.} A final $\mathrm{Conv}\,1{\times}1$ layer projects the features to the transformer embedding dimension (1536-d). The resulting feature map is flattened into $1024$ spatial tokens, each with $1536$ features, yielding $\mathbf{F}_{\text{CORE}} \in \mathbb{R}^{1024 \times 1536}$, which matches the transformer's feature space.

\vspace{2mm}

\noindent \textbf{Injection into Transformer Blocks.}
The CORE encoder processes the previous frame once, and the resulting representation is shared across all attention layers.
The CORE representation is injected into the cross-attention projections of every diffusion transformer block. In our implementation, a ``block'' refers to a transformer block (e.g., blocks 1--24 in SD3~\citep{esser2024scaling}).

\section{\drift: Guided Semantic Drift Correction}
\label{sec:why_dinov3}
This section provides additional details on the

choice of feature backbone used for our regularization. Narrative text descriptions in story visualization frequently underspecify visual appearance, particularly when character names are replaced by pronouns such as ``he'' ``she'', or ``they''. These descriptions often lack explicit details about identity-defining attributes such as hair shape, facial geometry, and clothing patterns~\cite{Santurkar2023CaptionWorthThousandImages}. Consequently, diffusion denoisers operating on noisy latent variables have no reliable visual reference for anchoring these attributes consistently across denoising timesteps~\cite{Huang2023NLIP}. Dense semantic guidance from self-supervised models help addresses this limitation~\cite{Zhang2023SDComplementsDINO,Wysoczanska2024CLIPDINOiser,Jose2025DINOv2MeetsText}. Unlike text encoders that rely on linguistic descriptions, we use a self-supervised backbone to extract language-agnostic, patch-level features directly from images. These features capture fine-grained visual structure, such as facial contours, clothing textures, and pose information, that remains consistent even when text descriptions are vague or referential. By aligning diffusion features with DINOv3~\citep{Simeoni2025DINOv3} embeddings during training, we constrain the denoiser toward patch level consistency, reduce appearance drift across iterations, and anchor generation to features that are invariant under small viewpoint or pose changes~\cite{Barsellotti2025TalkingToDINO}. This regularization is particularly crucial in story visualization, where maintaining character identity across frames is essential despite sparse or pronoun heavy linguistic cues~\cite{li2019storygan,rahman2023makeastory,shen2025storygptv}. Figure~\cref{fig:dino_vis} visualizes intermediate DINOv3 feature activations. The heatmaps show that the backbone responds most strongly to identity defining regions hair contours, facial structure, and clothing segments, while suppressing irrelevant background areas. This selective attention provides precisely the fine grained visual signal that narrative text typically omits. As a result, our regularization method offers language agnostic semantic guidance that complements the high-level narrative structure encoded in text and stabilizes the diffusion model's internal representation of character identity, appearance, and spatial layout across the story sequence.

\section{Experimental Details}
\label{sec:exp_details}
We train our proposed ReCap method on 8 NVIDIA H100 GPUs, each with 80 GB of memory, for 100,000 iterations with a batch size of 1 per GPU. We initialize the model from the publicly released \texttt{stabilityai/stable-diffusion-3-medium} checkpoint~\cite{esser2024scaling}.
The CORE module introduces 149K additional parameters ($\approx$ 0.12\% of the base SD3 model).

We employ a two-stage learning rate strategy: $1 \times 10^{-5}$ for the pretrained SD-3 backbone to prevent catastrophic forgetting~\cite{rombach2022ldm}, and $5 \times 10^{-5}$ for the newly introduced ReCap parameters. This differential learning rate allows the model to adapt to story visualization while preserving the pretrained generative capabilities.

For \drift, we extract features from a frozen DINOv3 ViT-B/16 encoder~\cite{Simeoni2025DINOv3}. The regularization weight is set to $\lambda_{\text{reg}} = 0.5$ throughout training. We use the AdamW optimizer with $\beta_1 = 0.9$, $\beta_2 = 0.999$, and a weight decay of $1 \times 10^{-2}$. Training takes approximately 19 hours on our hardware setup.

\section{Results}
\label{sec:qualitative}
\begin{figure*}
    \centering
\includegraphics[width=0.5\linewidth]{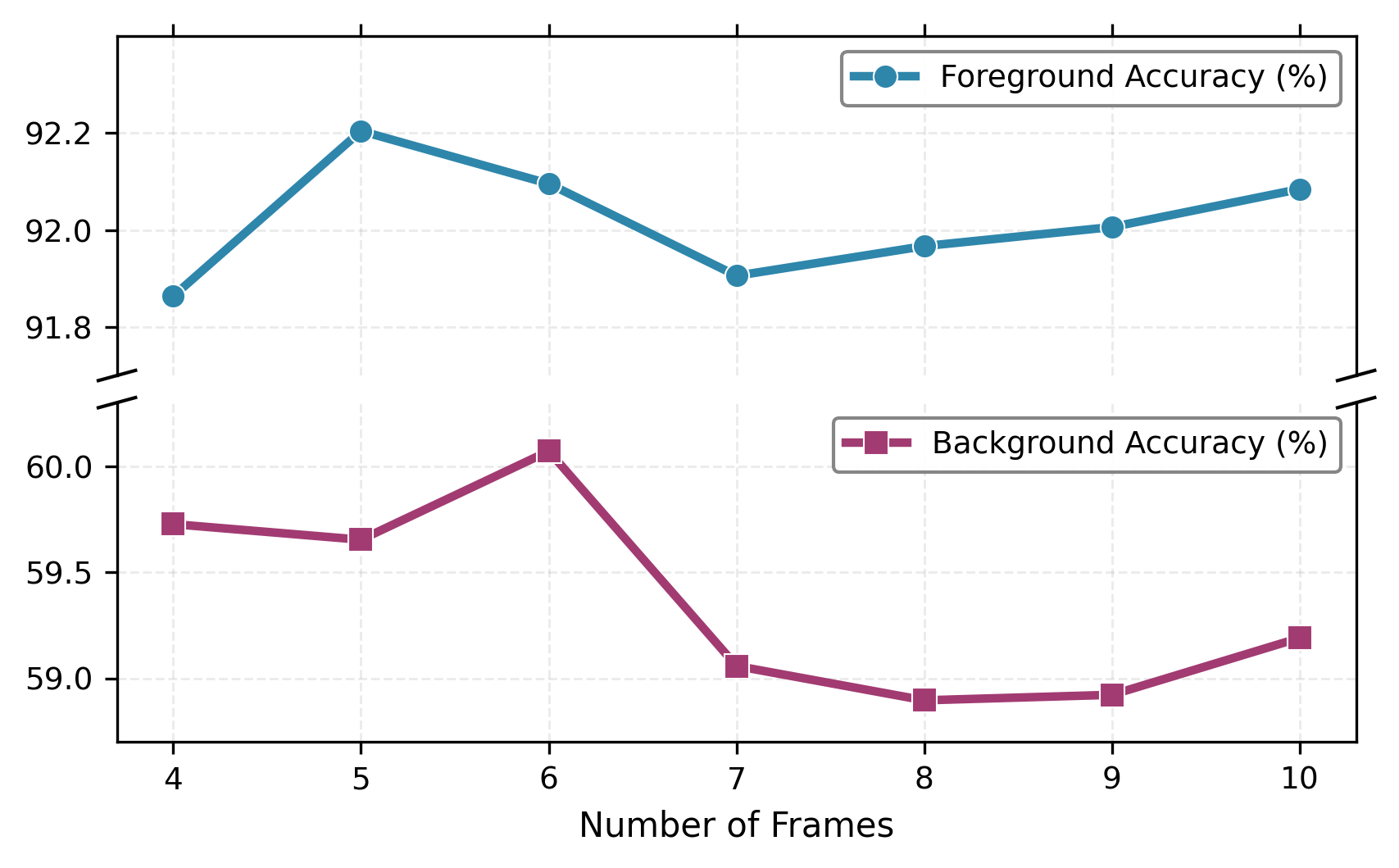}
    \caption{\textbf{Effect of temporal context length.} Foreground and background accuracy as the number of conditioning frames increases. Performance remains stable across longer temporal windows, indicating that increasing the number of frames does not degrade generation quality.}
    \label{fig:fgbg_plot}
\end{figure*}

\begin{figure*}[t]
\centering
\setlength{\tabcolsep}{1pt}
\renewcommand{\arraystretch}{0}

\begin{tabular}{ccccc}

    {\scriptsize \raisebox{5ex}{\shortstack{SD3~\citep{esser2024scaling}}}} &
    \includegraphics[width=.23\textwidth,clip,trim=0    0 1152 1152]{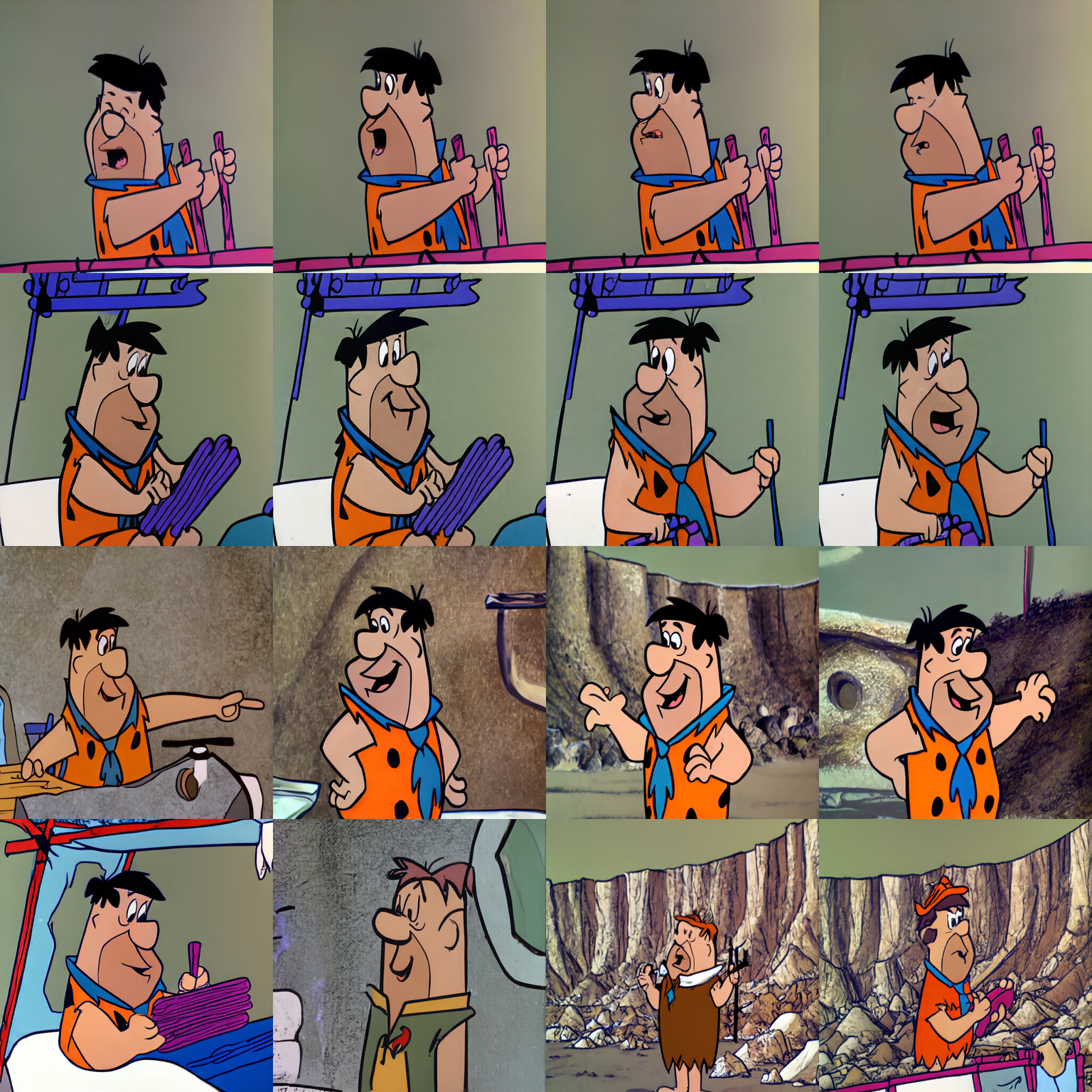} &
    \includegraphics[width=.23\textwidth,clip,trim=384  0 768  1152]{supp/visual19c.jpg} &
    \includegraphics[width=.23\textwidth,clip,trim=768  0 384  1152]{supp/visual19c.jpg} &
    \includegraphics[width=.23\textwidth,clip,trim=1152 0 0    1152]{supp/visual19c.jpg} \\[1mm]

    {\scriptsize \raisebox{5ex}{\shortstack{Story~\cite{shen2025storygptv}\\GPT-V}}} &
    \includegraphics[width=.23\textwidth,clip,trim=0    384 1152 768 ]{supp/visual19c.jpg} &
    \includegraphics[width=.23\textwidth,clip,trim=384  384 768  768 ]{supp/visual19c.jpg} &
    \includegraphics[width=.23\textwidth,clip,trim=768  384 384  768 ]{supp/visual19c.jpg} &
    \includegraphics[width=.23\textwidth,clip,trim=1152 384 0    768 ]{supp/visual19c.jpg} \\[1mm]

    {\scriptsize \raisebox{5ex}{\shortstack{ReCap\\(Ours)}}} &
    \includegraphics[width=.23\textwidth,clip,trim=0    768 1152 384 ]{supp/visual19c.jpg} &
    \includegraphics[width=.23\textwidth,clip,trim=384  768 768  384 ]{supp/visual19c.jpg} &
    \includegraphics[width=.23\textwidth,clip,trim=768  768 384  384 ]{supp/visual19c.jpg} &
    \includegraphics[width=.23\textwidth,clip,trim=1152 768 0    384 ]{supp/visual19c.jpg} \\[1mm]

    {\scriptsize \raisebox{5ex}{\shortstack{Ground\\Truth}}} &
    \includegraphics[width=.23\textwidth,clip,trim=0    1152 1152 0]{supp/visual19c.jpg} &
    \includegraphics[width=.23\textwidth,clip,trim=384  1152 768  0]{supp/visual19c.jpg} &
    \includegraphics[width=.23\textwidth,clip,trim=768  1152 384  0]{supp/visual19c.jpg} &
    \includegraphics[width=.23\textwidth,clip,trim=1152 1152 0    0]{supp/visual19c.jpg} \\[1mm]
& \colhead{Fred is seated in a room in a vehicle with both hands on purple, vertically oriented control bars.}
& \colhead{\textcolor{magenta}{He} is at work, talking to someone off screen.}
& \colhead{\textcolor{magenta}{He} is at the quarry. He is holding two sticks. He speaks to someone on his right.}
& \colhead{\textcolor{magenta}{He} is holding the steering poles of his equipment while in the quarry and is talking.} \\[1mm]

\end{tabular}

\caption{\textbf{Qualitative comparison on Flintstones SV \citep{gupta2018imagine} dataset} with method order from top to bottom: SD3, StoryGPT-V~\citep{shen2025storygptv}, ReCap (ours), and Ground Truth.}
\label{fig:supp1}
\end{figure*}

\begin{figure*}[t]
\centering
\setlength{\tabcolsep}{1pt}
\renewcommand{\arraystretch}{0}

\begin{tabular}{ccccc}

    {\scriptsize \raisebox{5ex}{\shortstack{SD3~\citep{esser2024scaling}}}} &
    \includegraphics[width=.23\textwidth,clip,trim=0    0 1152 1152]{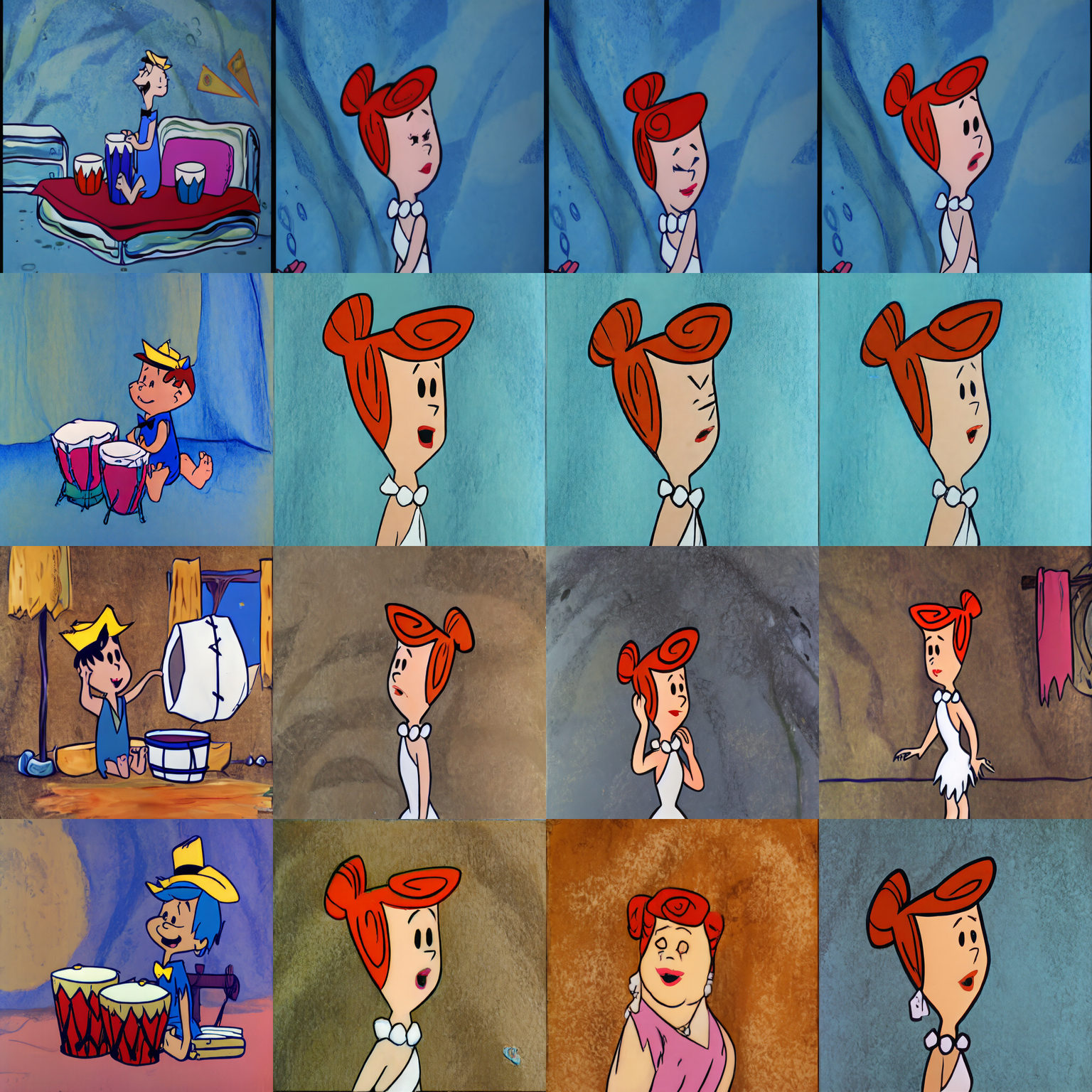} &
    \includegraphics[width=.23\textwidth,clip,trim=384  0 768  1152]{supp/visual20c.jpg} &
    \includegraphics[width=.23\textwidth,clip,trim=768  0 384  1152]{supp/visual20c.jpg} &
    \includegraphics[width=.23\textwidth,clip,trim=1152 0 0    1152]{supp/visual20c.jpg} \\[1mm]

    {\scriptsize \raisebox{5ex}{\shortstack{Story~\cite{shen2025storygptv}\\GPT-V}}} &
    \includegraphics[width=.23\textwidth,clip,trim=0    384 1152 768 ]{supp/visual20c.jpg} &
    \includegraphics[width=.23\textwidth,clip,trim=384  384 768  768 ]{supp/visual20c.jpg} &
    \includegraphics[width=.23\textwidth,clip,trim=768  384 384  768 ]{supp/visual20c.jpg} &
    \includegraphics[width=.23\textwidth,clip,trim=1152 384 0    768 ]{supp/visual20c.jpg} \\[1mm]

    {\scriptsize \raisebox{5ex}{\shortstack{ReCap\\(Ours)}}} &
    \includegraphics[width=.23\textwidth,clip,trim=0    768 1152 384 ]{supp/visual20c.jpg} &
    \includegraphics[width=.23\textwidth,clip,trim=384  768 768  384 ]{supp/visual20c.jpg} &
    \includegraphics[width=.23\textwidth,clip,trim=768  768 384  384 ]{supp/visual20c.jpg} &
    \includegraphics[width=.23\textwidth,clip,trim=1152 768 0    384 ]{supp/visual20c.jpg} \\[1mm]

    {\scriptsize \raisebox{5ex}{\shortstack{Ground\\Truth}}} &
    \includegraphics[width=.23\textwidth,clip,trim=0    1152 1152 0]{supp/visual20c.jpg} &
    \includegraphics[width=.23\textwidth,clip,trim=384  1152 768  0]{supp/visual20c.jpg} &
    \includegraphics[width=.23\textwidth,clip,trim=768  1152 384  0]{supp/visual20c.jpg} &
    \includegraphics[width=.23\textwidth,clip,trim=1152 1152 0    0]{supp/visual20c.jpg} \\[1mm]
& \colhead{A boy with blue shirt and yellow hat and bow tie is sitting in his bedroom with three drums next to him.}
& \colhead{Wilma in the room talking to someone.}
& \colhead{\textcolor{magenta}{She} is in a room. She says something then scrunches her face together.}
& \colhead{\textcolor{magenta}{She} is in the room. She has her head turned as she speaks.} \\[1mm]

\end{tabular}

\caption{\textbf{Qualitative comparison on Flintstones SV \citep{gupta2018imagine} dataset} with method order from top to bottom: SD3, StoryGPT-V~\citep{shen2025storygptv}, ReCap (ours), and Ground Truth.}
\label{fig:supp1b}
\end{figure*}

\begin{figure*}[t]
\centering
\setlength{\tabcolsep}{1pt}
\renewcommand{\arraystretch}{0}

\begin{tabular}{ccccc}

    {\scriptsize \raisebox{5ex}{\shortstack{Story~\citep{Maharana2022StoryDALLE}\\DALL-E}}} &
    \includegraphics[width=.23\textwidth,clip,trim=0    0 1152 1152]{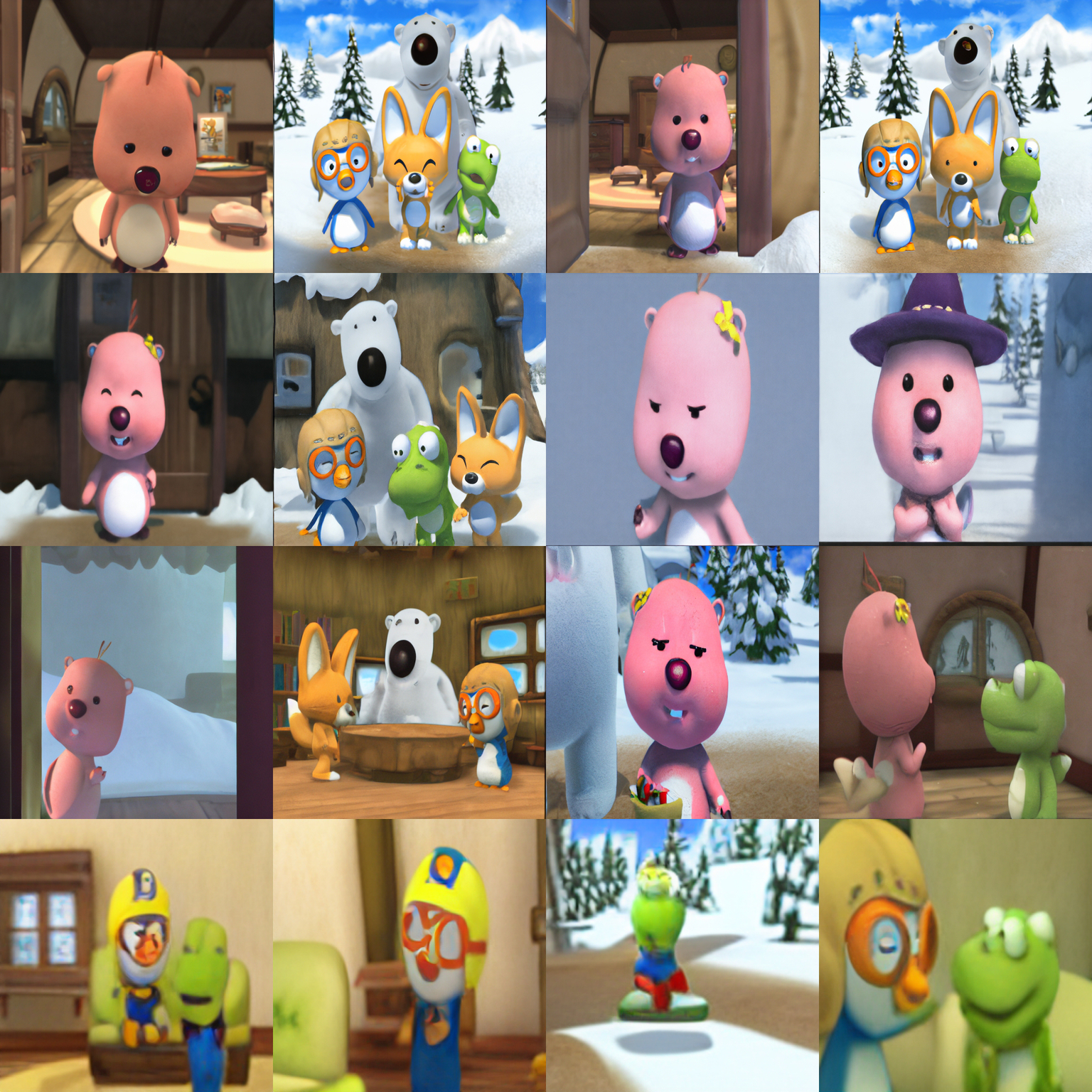} &
    \includegraphics[width=.23\textwidth,clip,trim=384  0 768  1152]{supp/Pororo_ENGLISH4_2-Pororo_ENGLISH4_2_ep1-42c.jpg} &
    \includegraphics[width=.23\textwidth,clip,trim=768  0 384  1152]{supp/Pororo_ENGLISH4_2-Pororo_ENGLISH4_2_ep1-42c.jpg} &
    \includegraphics[width=.23\textwidth,clip,trim=1152 0 0    1152]{supp/Pororo_ENGLISH4_2-Pororo_ENGLISH4_2_ep1-42c.jpg} \\[1mm]

    {\scriptsize \raisebox{5ex}{\shortstack{Story~\cite{shen2025storygptv}\\GPT-V}}} &
    \includegraphics[width=.23\textwidth,clip,trim=0    384 1152 768 ]{supp/Pororo_ENGLISH4_2-Pororo_ENGLISH4_2_ep1-42c.jpg} &
    \includegraphics[width=.23\textwidth,clip,trim=384  384 768  768 ]{supp/Pororo_ENGLISH4_2-Pororo_ENGLISH4_2_ep1-42c.jpg} &
    \includegraphics[width=.23\textwidth,clip,trim=768  384 384  768 ]{supp/Pororo_ENGLISH4_2-Pororo_ENGLISH4_2_ep1-42c.jpg} &
    \includegraphics[width=.23\textwidth,clip,trim=1152 384 0    768 ]{supp/Pororo_ENGLISH4_2-Pororo_ENGLISH4_2_ep1-42c.jpg} \\[1mm]

    {\scriptsize \raisebox{5ex}{\shortstack{ReCap\\(Ours)}}} &
    \includegraphics[width=.23\textwidth,clip,trim=0    768 1152 384 ]{supp/Pororo_ENGLISH4_2-Pororo_ENGLISH4_2_ep1-42c.jpg} &
    \includegraphics[width=.23\textwidth,clip,trim=384  768 768  384 ]{supp/Pororo_ENGLISH4_2-Pororo_ENGLISH4_2_ep1-42c.jpg} &
    \includegraphics[width=.23\textwidth,clip,trim=768  768 384  384 ]{supp/Pororo_ENGLISH4_2-Pororo_ENGLISH4_2_ep1-42c.jpg} &
    \includegraphics[width=.23\textwidth,clip,trim=1152 768 0    384 ]{supp/Pororo_ENGLISH4_2-Pororo_ENGLISH4_2_ep1-42c.jpg} \\[1mm]

    {\scriptsize \raisebox{5ex}{\shortstack{Ground\\Truth}}} &
    \includegraphics[width=.23\textwidth,clip,trim=0    1152 1152 0]{supp/Pororo_ENGLISH4_2-Pororo_ENGLISH4_2_ep1-42c.jpg} &
    \includegraphics[width=.23\textwidth,clip,trim=384  1152 768  0]{supp/Pororo_ENGLISH4_2-Pororo_ENGLISH4_2_ep1-42c.jpg} &
    \includegraphics[width=.23\textwidth,clip,trim=768  1152 384  0]{supp/Pororo_ENGLISH4_2-Pororo_ENGLISH4_2_ep1-42c.jpg} &
    \includegraphics[width=.23\textwidth,clip,trim=1152 1152 0    0]{supp/Pororo_ENGLISH4_2-Pororo_ENGLISH4_2_ep1-42c.jpg} \\[1mm]
& \colhead{Someone calls Loopy. Loopy looks at the door.}
& \colhead{Crong Pororo Poby and Eddy have come over to Loopy's house.}
& \colhead{Loopy does not look happy even thought her friends have visited her.}
& \colhead{\textcolor{magenta}{She} invites her friends in.} \\[1mm]

\end{tabular}

\caption{\textbf{Qualitative comparison on PororoSV~\citep{li2019storygan} dataset} with method order from top to bottom: SD3, StoryGPT-V~\citep{shen2025storygptv}, ReCap (ours), and Ground Truth.}
\label{fig:supp2a}
\end{figure*}

\begin{figure*}[t]
\centering
\setlength{\tabcolsep}{1pt}
\renewcommand{\arraystretch}{0}

\begin{tabular}{ccccc}

    {\scriptsize \raisebox{5ex}{\shortstack{Story~\citep{Maharana2022StoryDALLE}\\DALL-E}}} &
    \includegraphics[width=.23\textwidth,clip,trim=0    0 1152 1152]{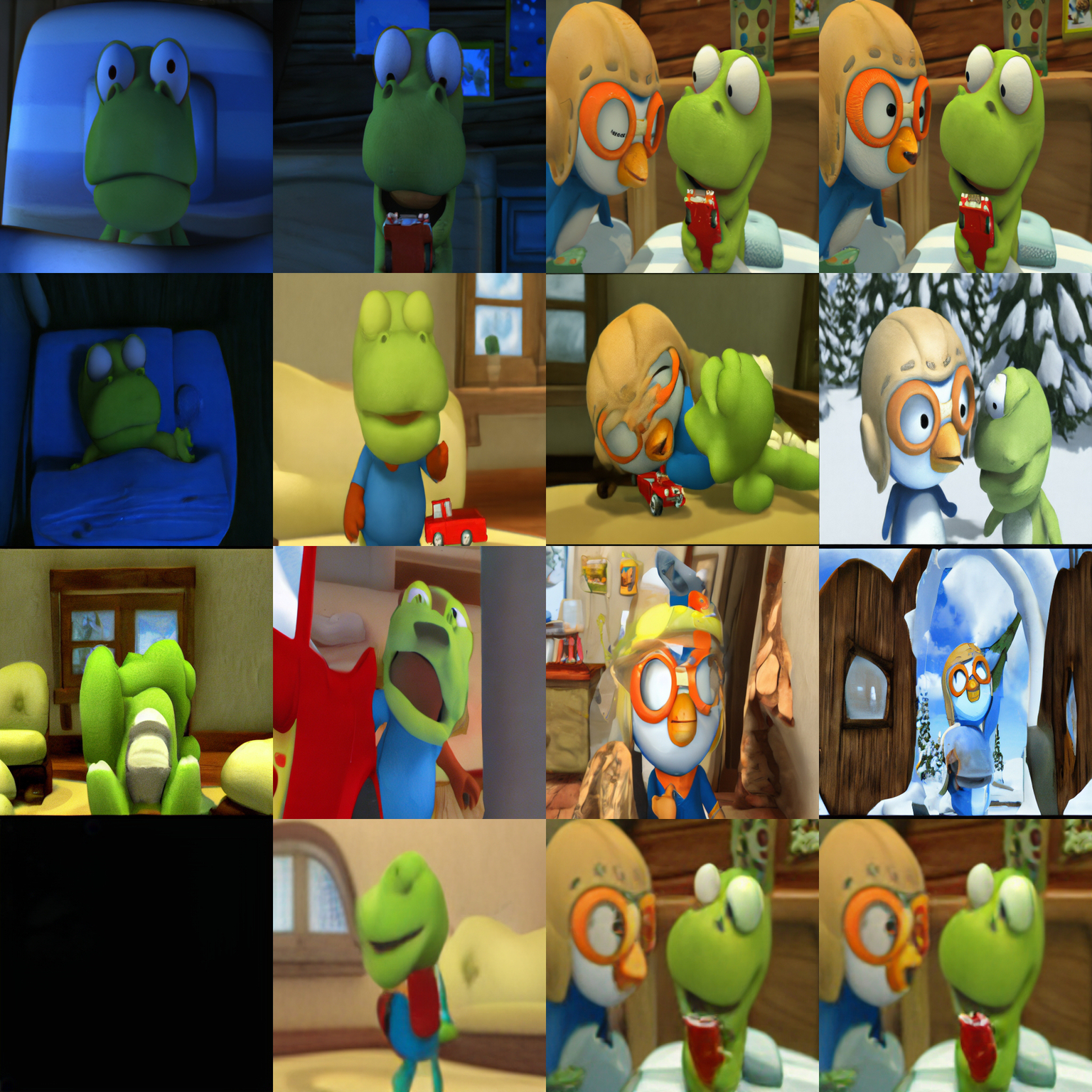} &
    \includegraphics[width=.23\textwidth,clip,trim=384  0 768  1152]{supp/Pororo_ENGLISH2_4-Pororo_ENGLISH2_4_ep12-55c.jpg} &
    \includegraphics[width=.23\textwidth,clip,trim=768  0 384  1152]{supp/Pororo_ENGLISH2_4-Pororo_ENGLISH2_4_ep12-55c.jpg} &
    \includegraphics[width=.23\textwidth,clip,trim=1152 0 0    1152]{supp/Pororo_ENGLISH2_4-Pororo_ENGLISH2_4_ep12-55c.jpg} \\[1mm]

    {\scriptsize \raisebox{5ex}{\shortstack{Story~\cite{shen2025storygptv}\\GPT-V}}} &
    \includegraphics[width=.23\textwidth,clip,trim=0    384 1152 768 ]{supp/Pororo_ENGLISH2_4-Pororo_ENGLISH2_4_ep12-55c.jpg} &
    \includegraphics[width=.23\textwidth,clip,trim=384  384 768  768 ]{supp/Pororo_ENGLISH2_4-Pororo_ENGLISH2_4_ep12-55c.jpg} &
    \includegraphics[width=.23\textwidth,clip,trim=768  384 384  768 ]{supp/Pororo_ENGLISH2_4-Pororo_ENGLISH2_4_ep12-55c.jpg} &
    \includegraphics[width=.23\textwidth,clip,trim=1152 384 0    768 ]{supp/Pororo_ENGLISH2_4-Pororo_ENGLISH2_4_ep12-55c.jpg} \\[1mm]

    {\scriptsize \raisebox{5ex}{\shortstack{ReCap\\(Ours)}}} &
    \includegraphics[width=.23\textwidth,clip,trim=0    768 1152 384 ]{supp/Pororo_ENGLISH2_4-Pororo_ENGLISH2_4_ep12-55c.jpg} &
    \includegraphics[width=.23\textwidth,clip,trim=384  768 768  384 ]{supp/Pororo_ENGLISH2_4-Pororo_ENGLISH2_4_ep12-55c.jpg} &
    \includegraphics[width=.23\textwidth,clip,trim=768  768 384  384 ]{supp/Pororo_ENGLISH2_4-Pororo_ENGLISH2_4_ep12-55c.jpg} &
    \includegraphics[width=.23\textwidth,clip,trim=1152 768 0    384 ]{supp/Pororo_ENGLISH2_4-Pororo_ENGLISH2_4_ep12-55c.jpg} \\[1mm]

    {\scriptsize \raisebox{5ex}{\shortstack{Ground\\Truth}}} &
    \includegraphics[width=.23\textwidth,clip,trim=0    1152 1152 0]{supp/Pororo_ENGLISH2_4-Pororo_ENGLISH2_4_ep12-55c.jpg} &
    \includegraphics[width=.23\textwidth,clip,trim=384  1152 768  0]{supp/Pororo_ENGLISH2_4-Pororo_ENGLISH2_4_ep12-55c.jpg} &
    \includegraphics[width=.23\textwidth,clip,trim=768  1152 384  0]{supp/Pororo_ENGLISH2_4-Pororo_ENGLISH2_4_ep12-55c.jpg} &
    \includegraphics[width=.23\textwidth,clip,trim=1152 1152 0    0]{supp/Pororo_ENGLISH2_4-Pororo_ENGLISH2_4_ep12-55c.jpg} \\[1mm]
& \colhead{Crong feels sleepless but close his eyes and lies on his left.}
& \colhead{With his surprise \textcolor{magenta}{He} finds the red toy car.}
& \colhead{Pororo wakes up and looks at the toy car Crong found.}
& \colhead{\textcolor{magenta}{He} asks Crong where it was found.} \\[1mm]

\end{tabular}

\caption{\textbf{Qualitative comparison on PororoSV~\citep{li2019storygan} dataset} with method order from top to bottom: SD3, StoryGPT-V~\citep{shen2025storygptv}, ReCap (ours), and Ground Truth.}
\label{fig:supp2b}
\end{figure*}

\begin{figure*}[!t]
\centering
\setlength{\tabcolsep}{1.2pt}
\renewcommand{\arraystretch}{0}
\begin{tabular}{c c c c c c c}
{\tiny \raisebox{4.5ex}{\shortstack{SD3\\(baseline)}}} &
\includegraphics[width=.145\textwidth]{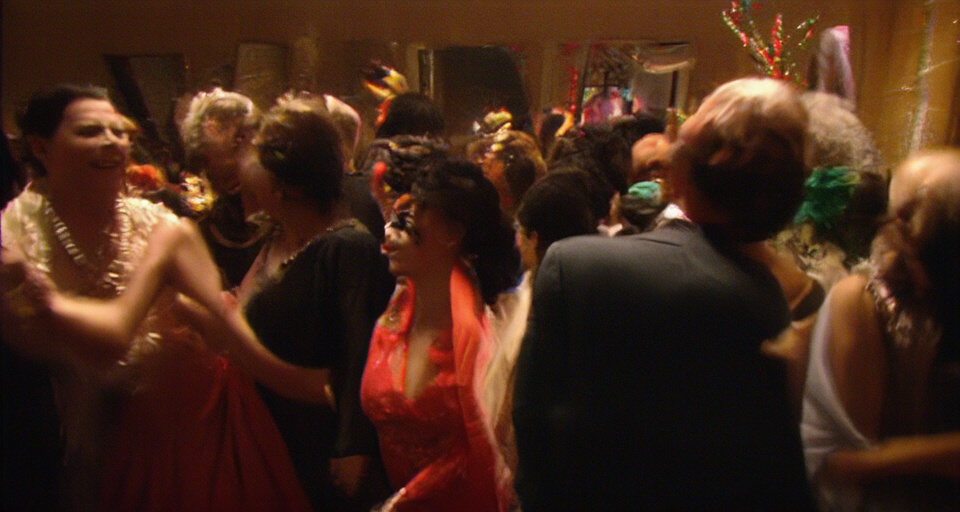} &
\includegraphics[width=.145\textwidth]{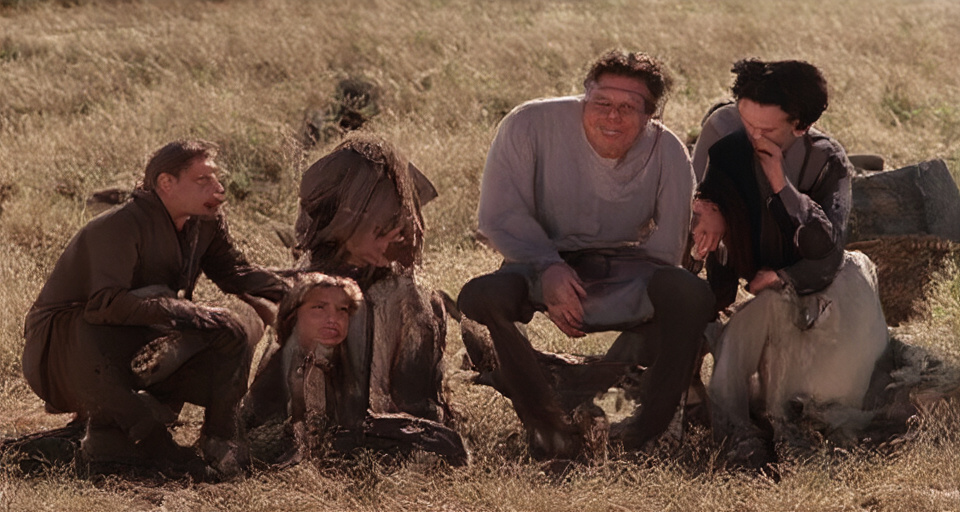} &
\includegraphics[width=.145\textwidth]{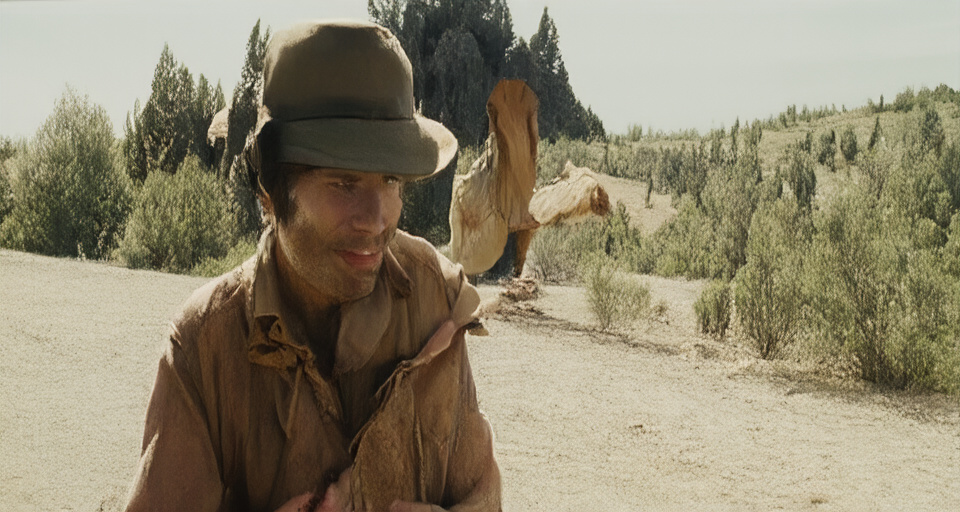} &
\includegraphics[width=.145\textwidth]{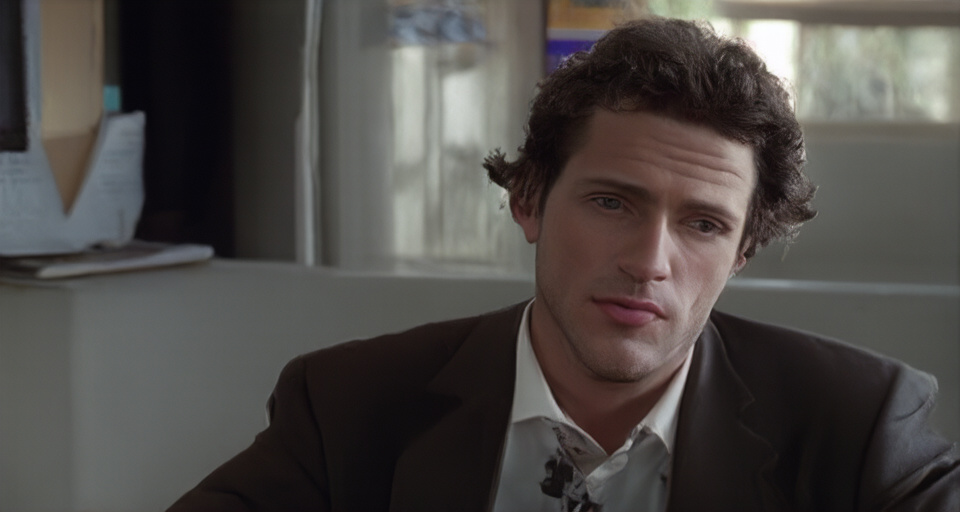} &
\includegraphics[width=.145\textwidth]{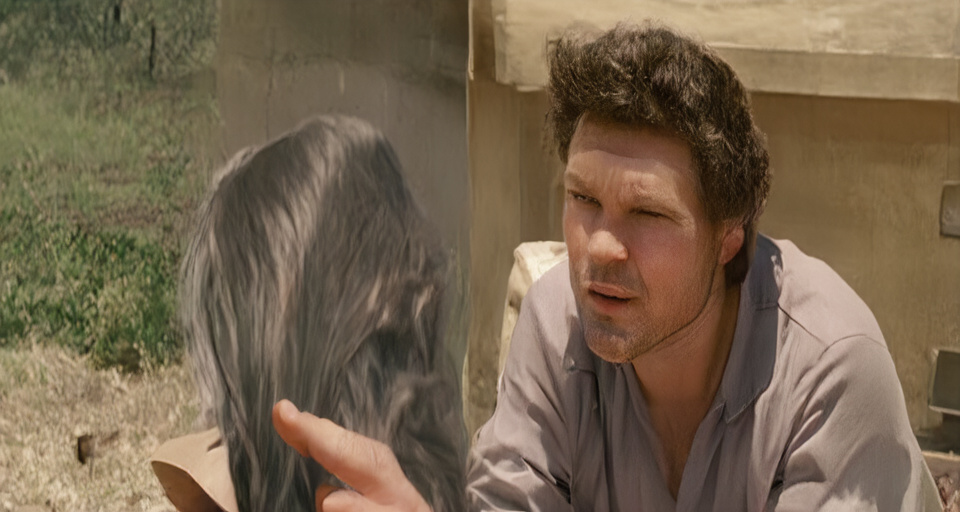} &
\includegraphics[width=.145\textwidth]{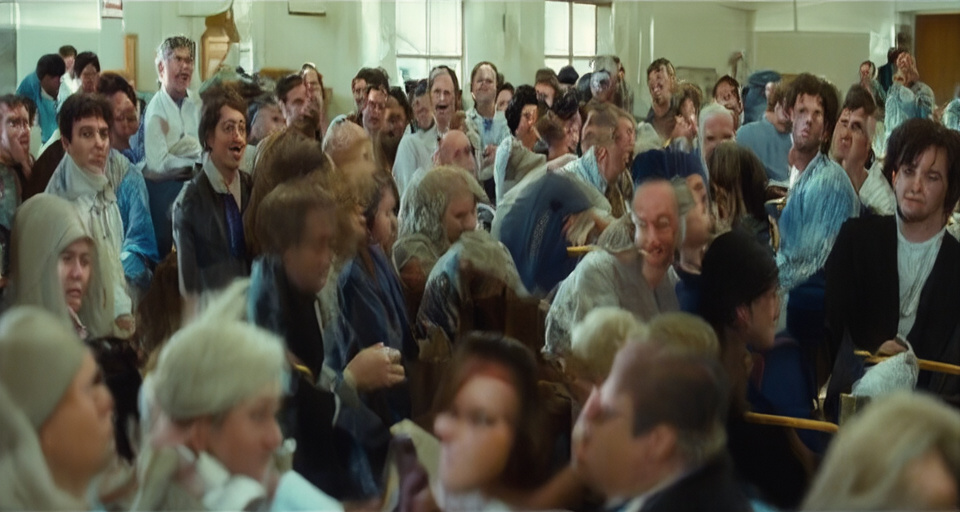} \\[1.5mm]

{\tiny \raisebox{4.5ex}{\shortstack{ReCap\\(Ours)}}} &
\includegraphics[width=.145\textwidth]{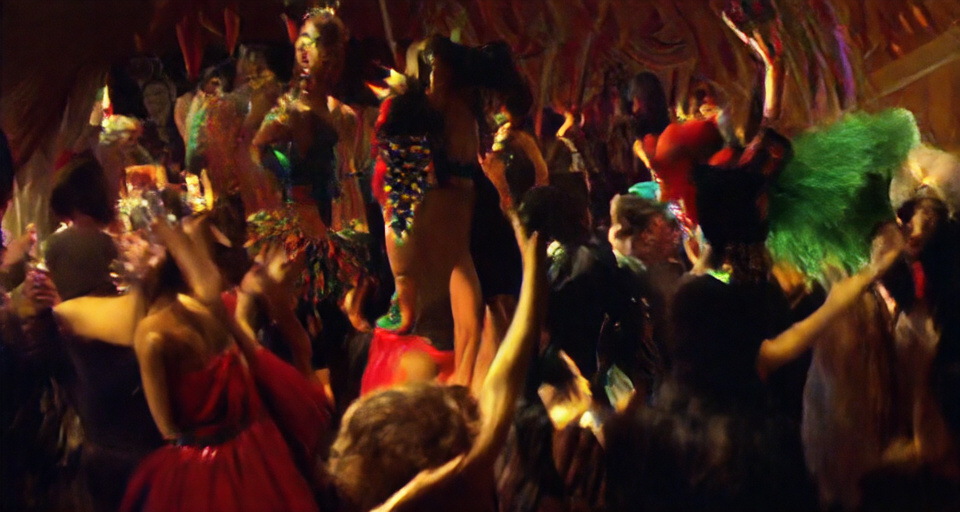} &
\includegraphics[width=.145\textwidth]{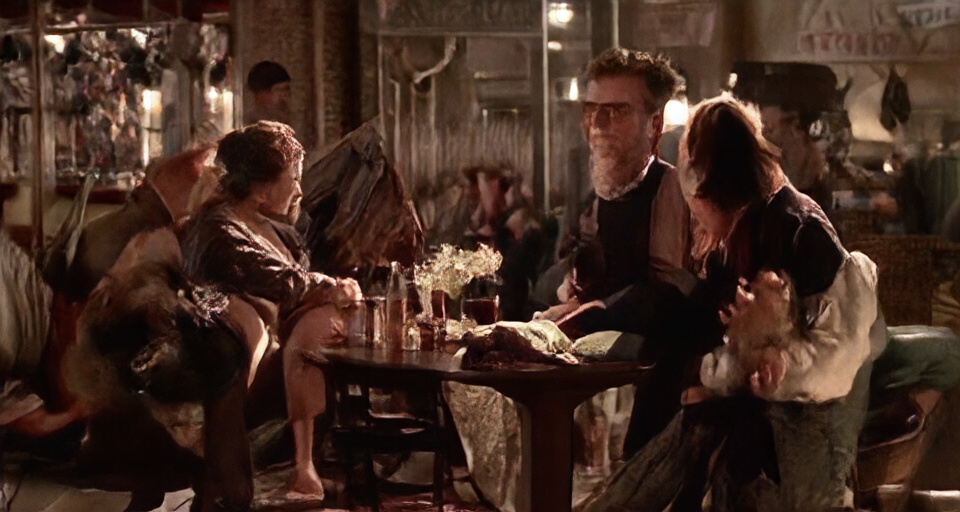} &
\includegraphics[width=.145\textwidth]{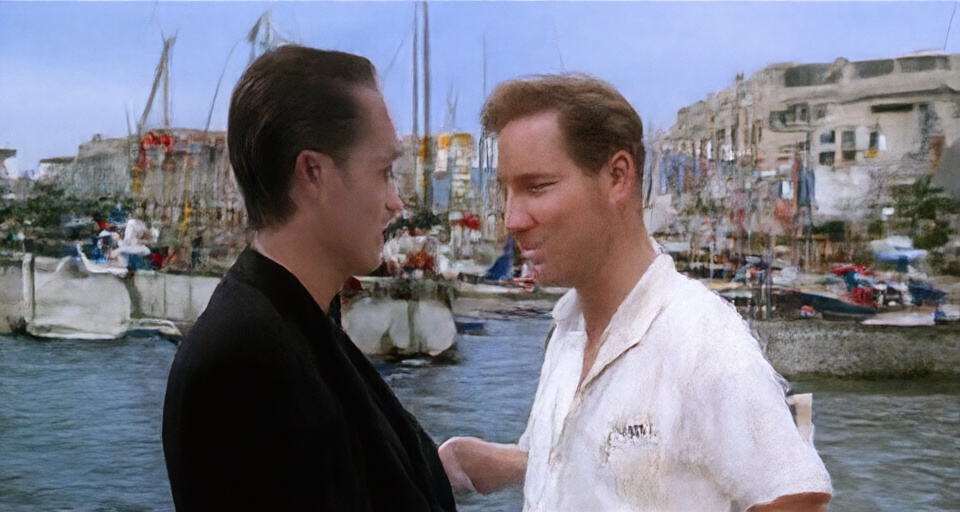} &
\includegraphics[width=.145\textwidth]{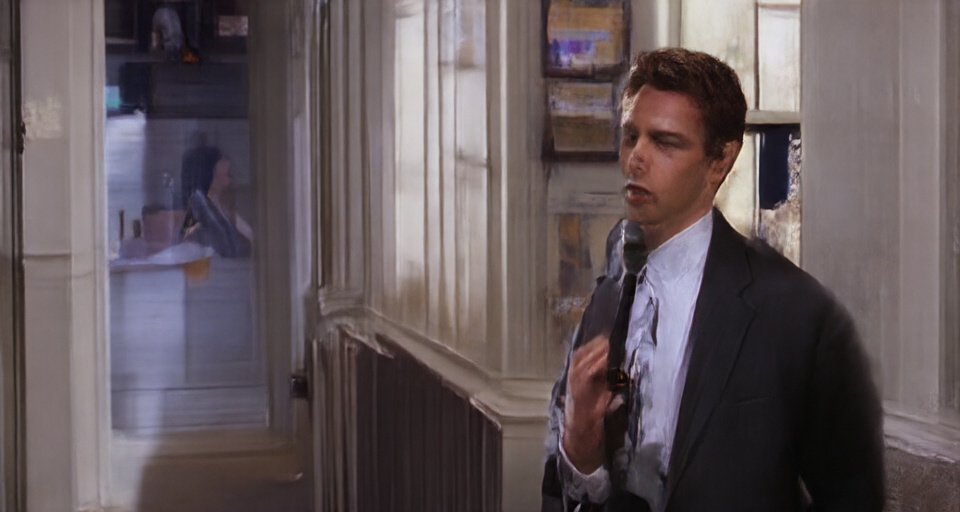} &
\includegraphics[width=.145\textwidth]{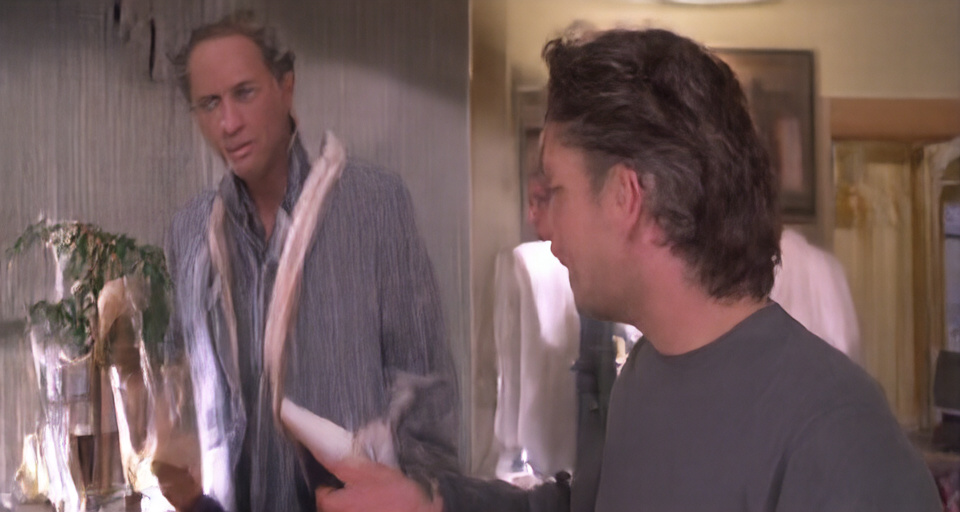} &
\includegraphics[width=.145\textwidth]{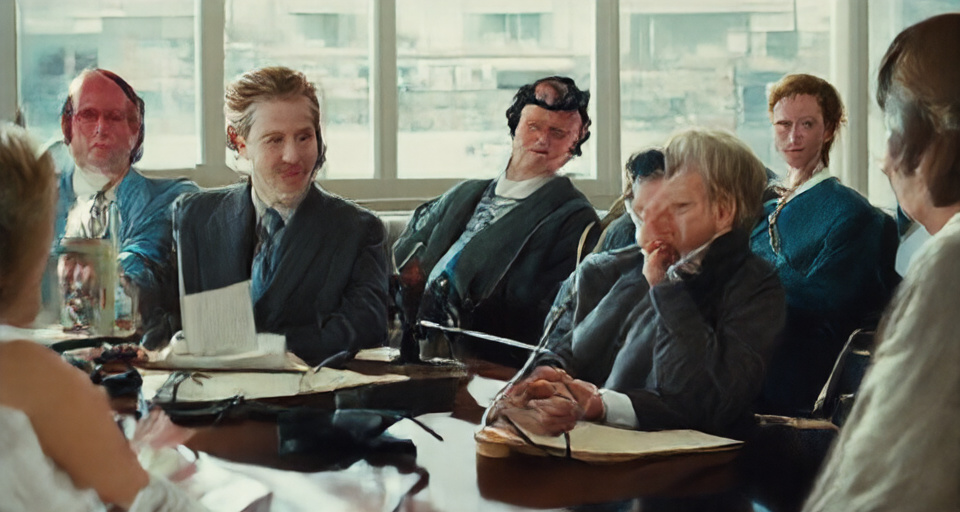} \\[1.5mm]

{\tiny \raisebox{4.5ex}{\shortstack{Ground\\Truth}}} &
\includegraphics[width=.145\textwidth]{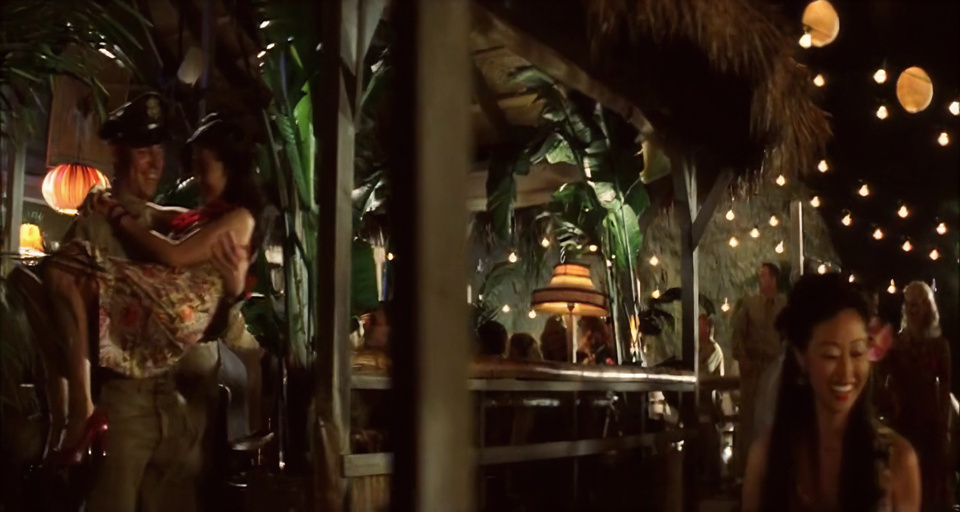} &
\includegraphics[width=.145\textwidth]{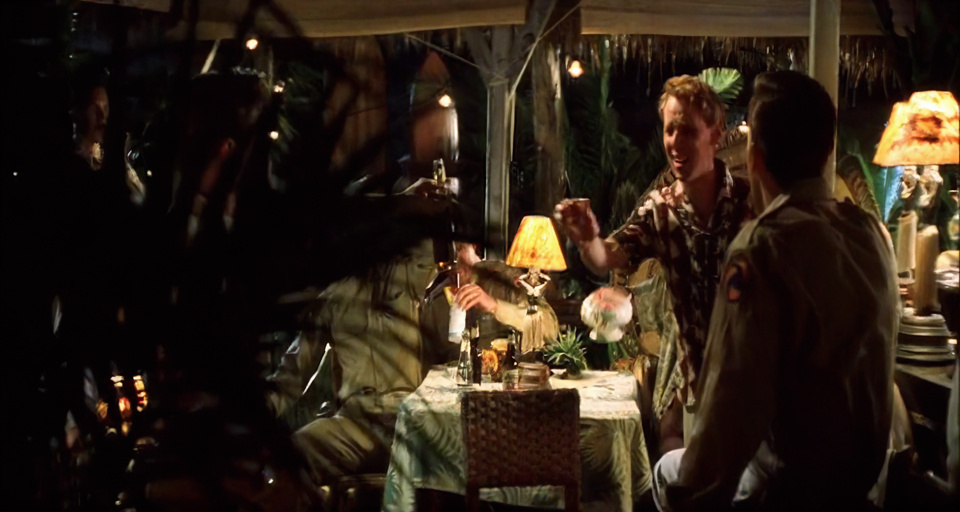} &
\includegraphics[width=.145\textwidth]{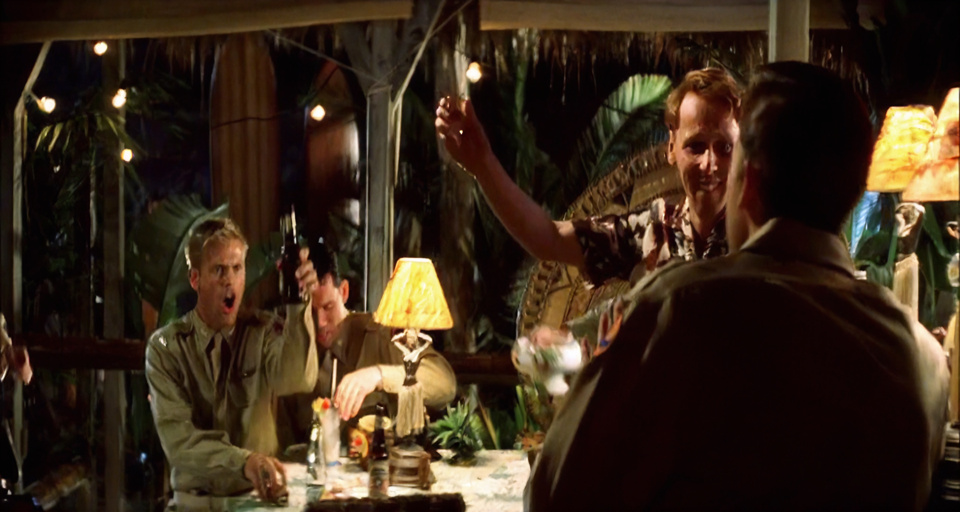} &
\includegraphics[width=.145\textwidth]{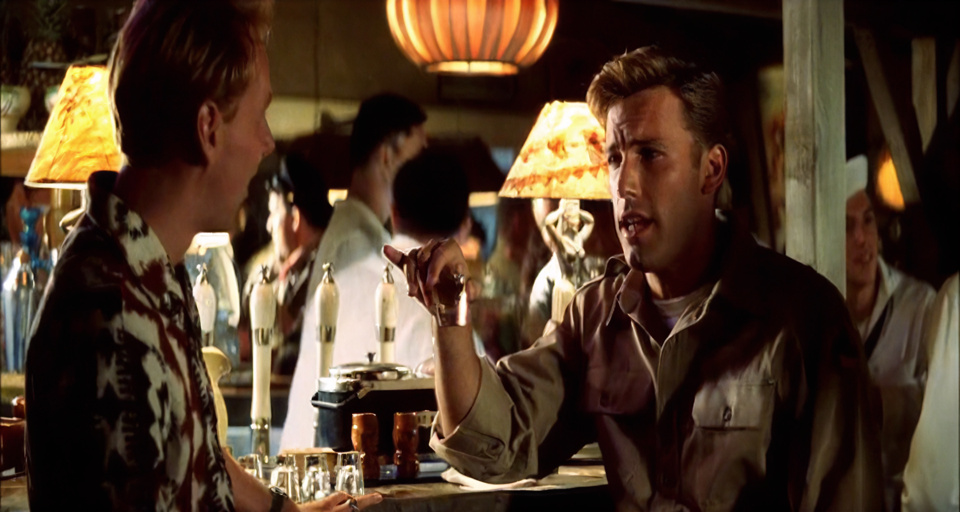} &
\includegraphics[width=.145\textwidth]{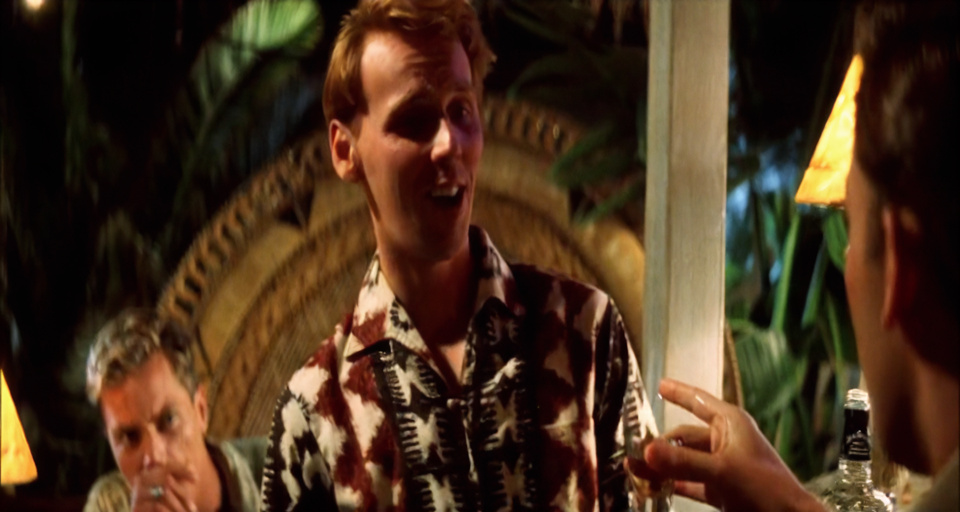} &
\includegraphics[width=.145\textwidth]{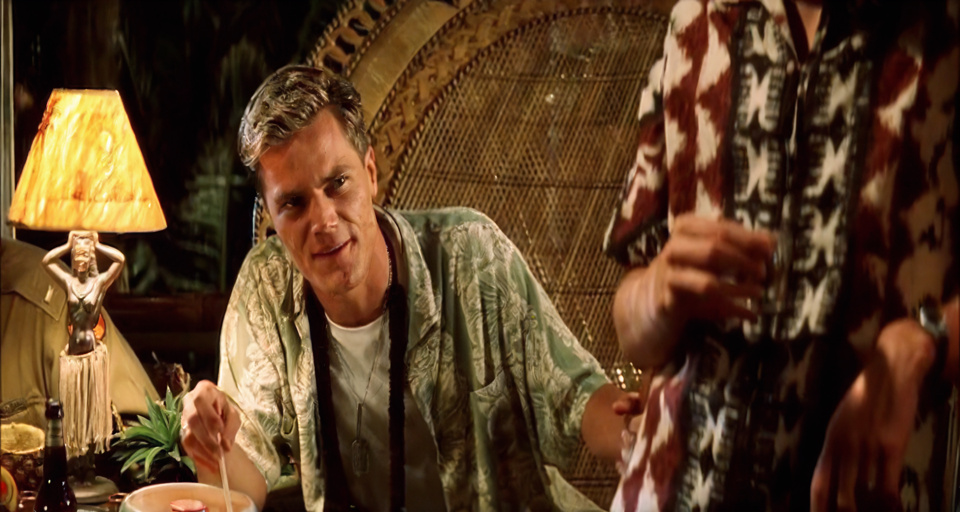} \\[0.8mm]

\end{tabular}
\vspace{0.25em}
\begin{tabular}{p{.08\textwidth}p{.145\textwidth}p{.145\textwidth}p{.145\textwidth}p{.145\textwidth}p{.145\textwidth}p{.145\textwidth}}
&
\raggedright\tiny there was a big themed party going on in the evening &
\raggedright\tiny a group of friends were there talking about their latest adventures &
\raggedright\tiny [male0] talked about his recent travels and how they went &
\raggedright\tiny [male1] was n't very impressed with his stories &
\raggedright\tiny [male0] challenged [male1] to find some better stories &
\raggedright\tiny everyone was interested to hear what [male1] would have to say.
\end{tabular}

\caption{Qualitative comparison on VWP~\citep{hong2023visual}. We employ our pronoun replacing strategy on top of the text taken directly from the dataset. The scene depicts a crowded party where the narrative focuses on a conversation between two individuals. SD3 generates inconsistent participants across frames and fails to ground the speakers within the crowd. Our method produces a more coherent sequence that better reflects the described conversation within the busy scene.}
\label{fig:supp3a}  \vspace{-1.5em}
\end{figure*}

\begin{figure*}[!t]
\centering
\setlength{\tabcolsep}{1.2pt}
\renewcommand{\arraystretch}{0}
\begin{tabular}{c c c c c c}
{\tiny \raisebox{4.5ex}{\shortstack{SD3\\(baseline)}}} &
\includegraphics[width=.175\textwidth]{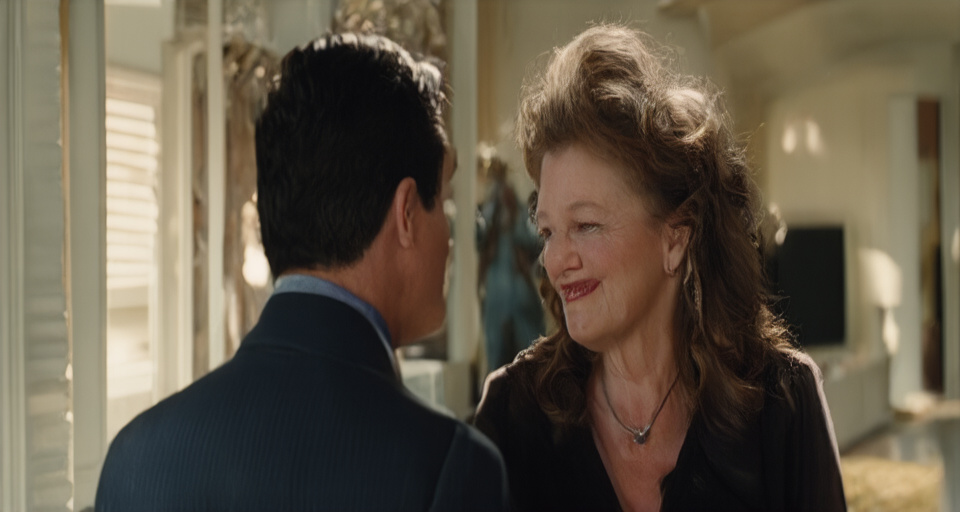} &
\includegraphics[width=.175\textwidth]{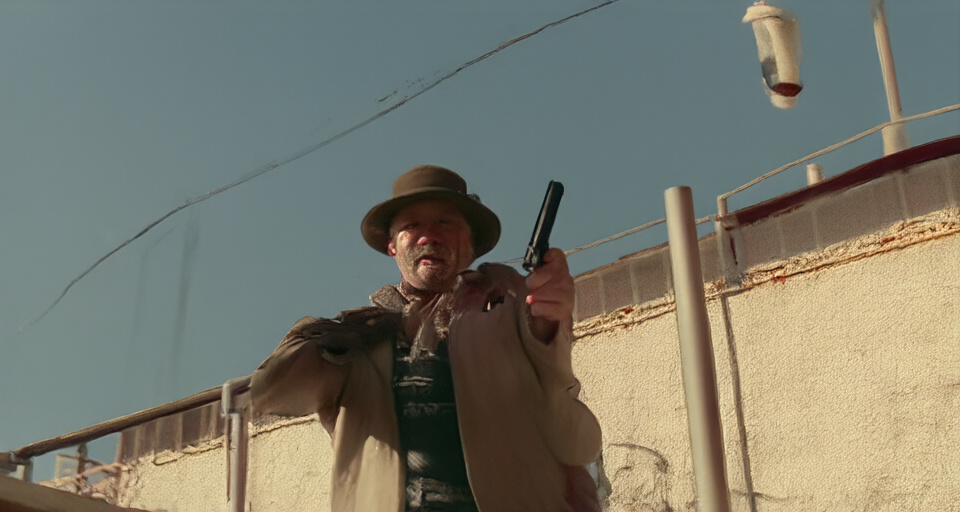} &
\includegraphics[width=.175\textwidth]{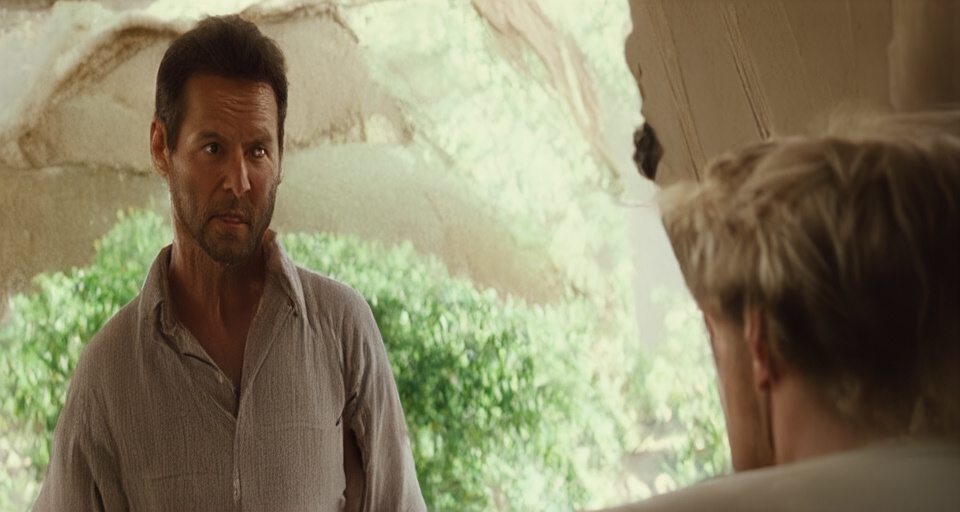} &
\includegraphics[width=.175\textwidth]{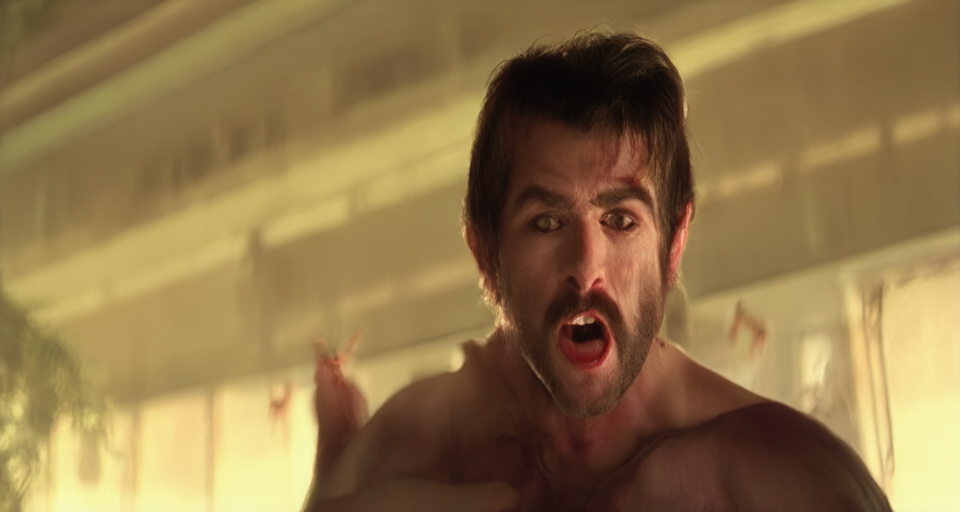} &
\includegraphics[width=.175\textwidth]{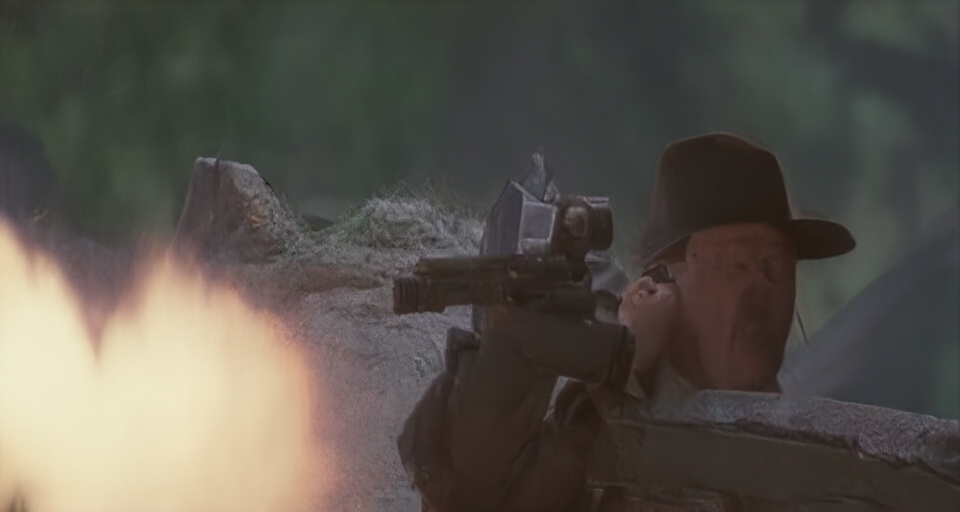} \\[1.5mm]

{\tiny \raisebox{4.5ex}{\shortstack{ReCap\\(Ours)}}} &
\includegraphics[width=.175\textwidth]{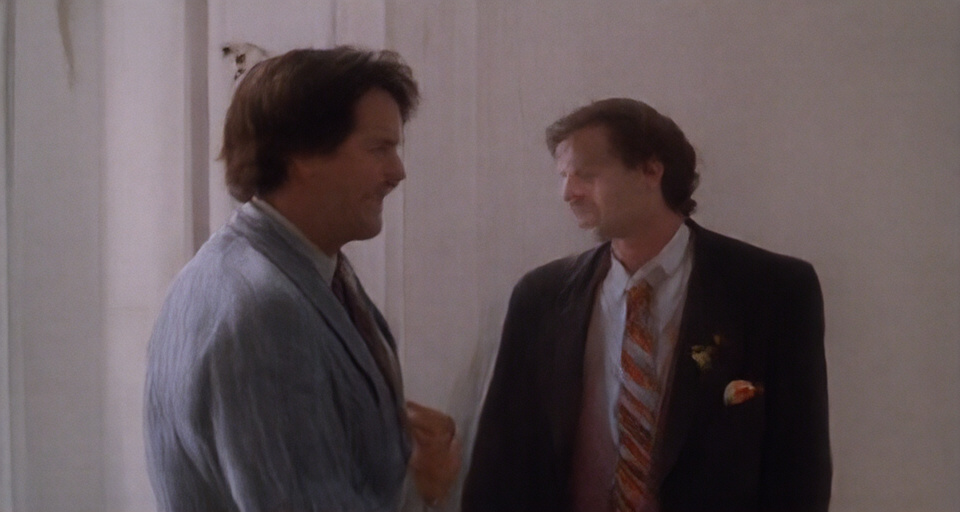} &
\includegraphics[width=.175\textwidth]{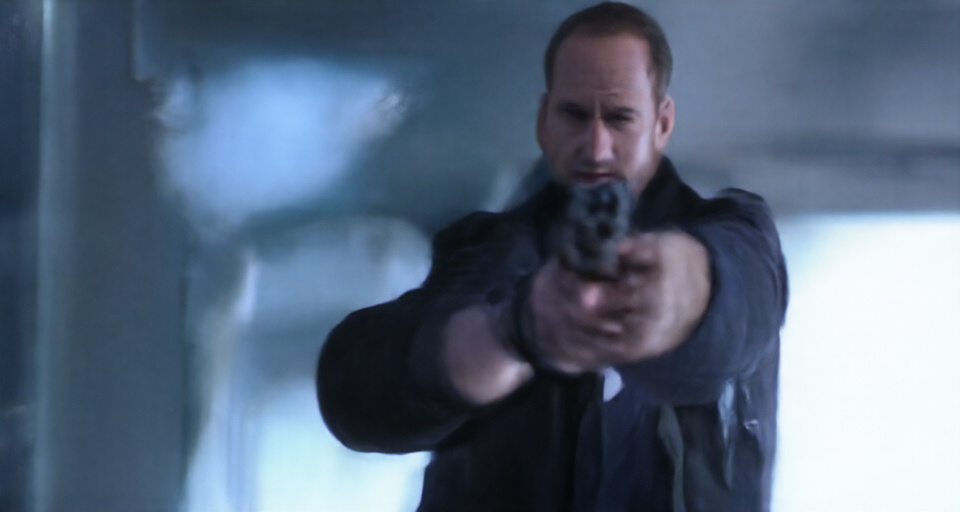} &
\includegraphics[width=.175\textwidth]{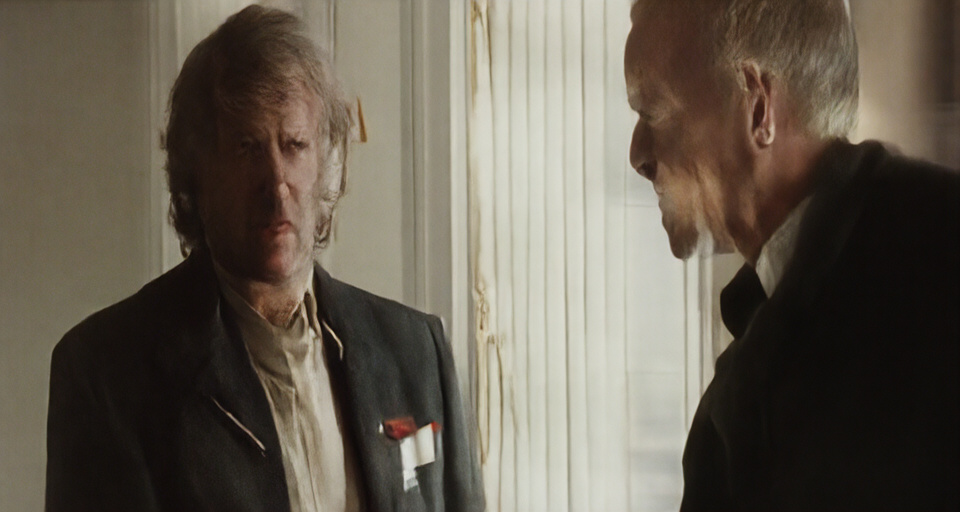} &
\includegraphics[width=.175\textwidth]{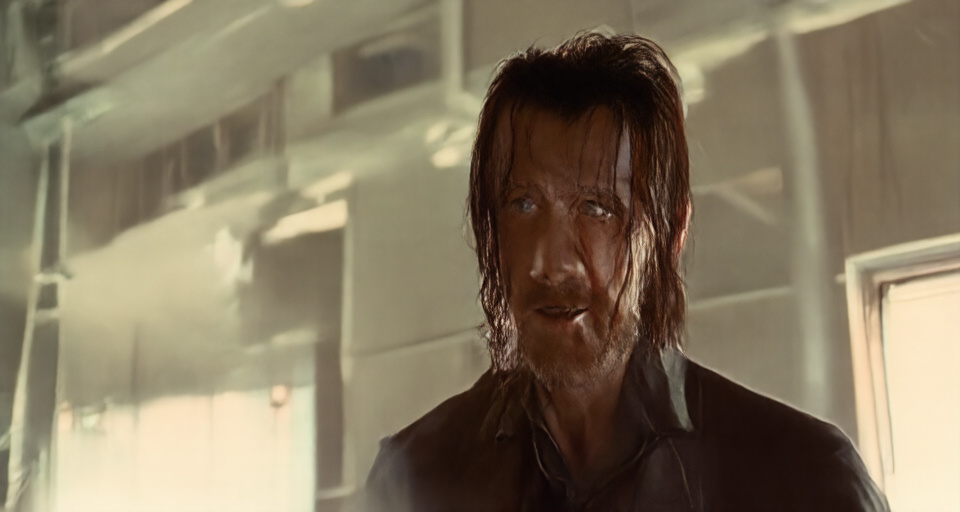} &
\includegraphics[width=.175\textwidth]{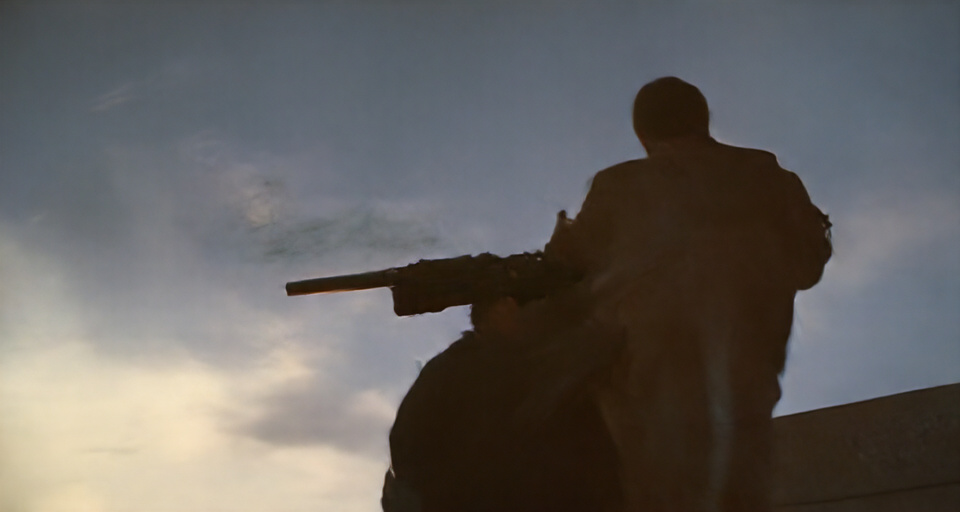} \\[1.5mm]

{\tiny \raisebox{4.5ex}{\shortstack{Ground\\Truth}}} &
\includegraphics[width=.175\textwidth]{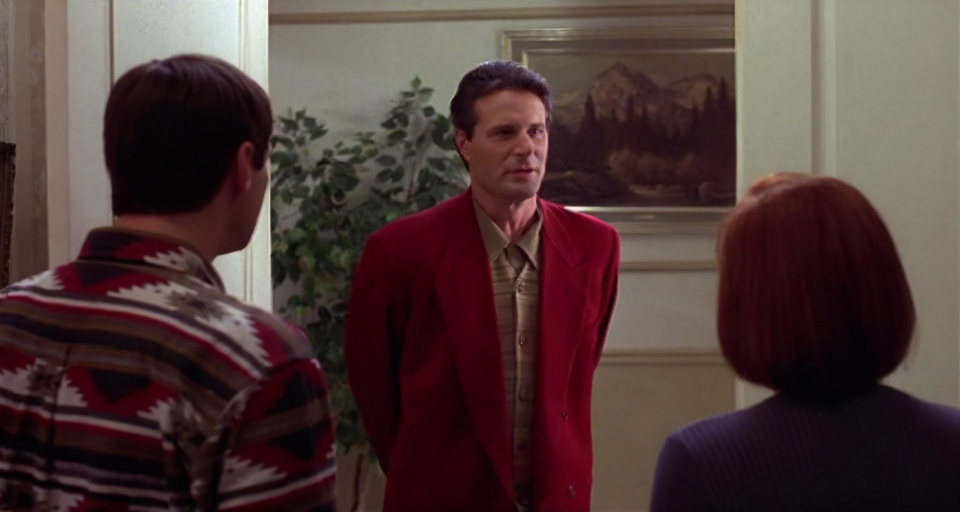} &
\includegraphics[width=.175\textwidth]{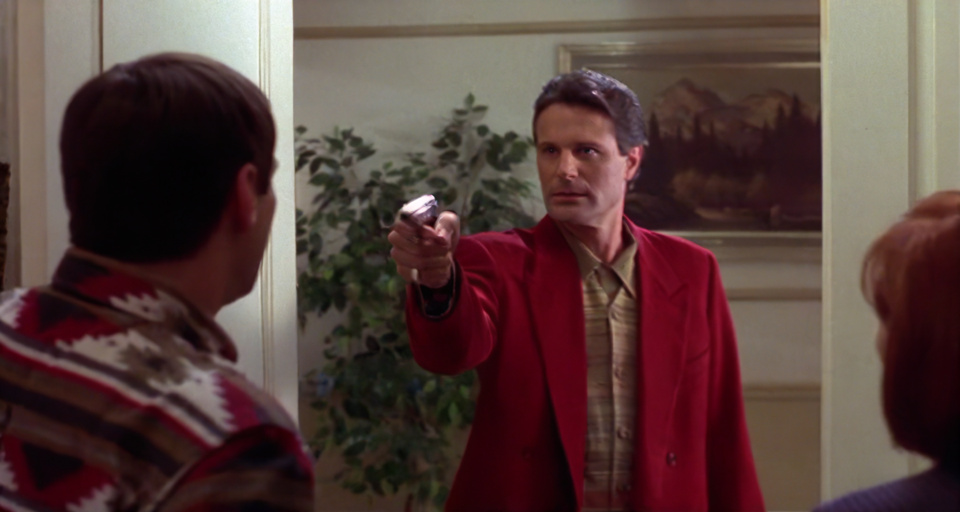} &
\includegraphics[width=.175\textwidth]{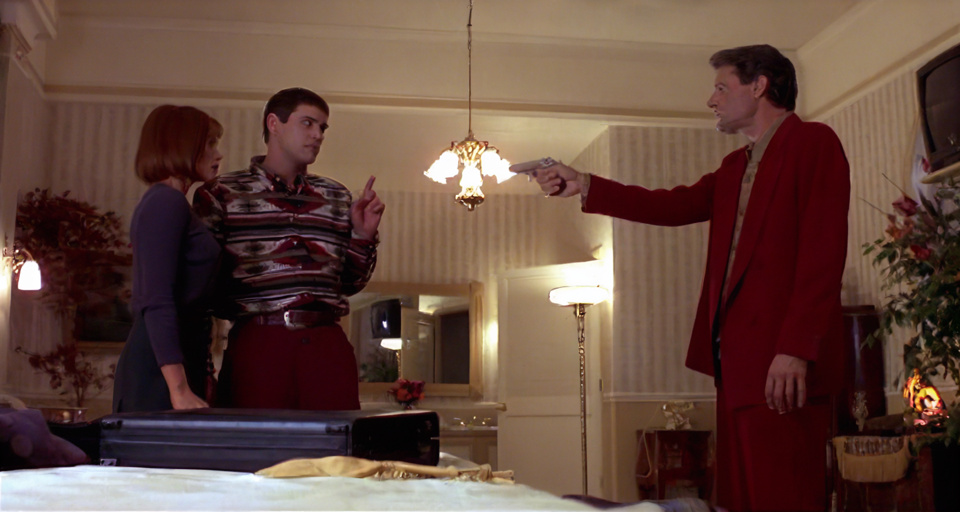} &
\includegraphics[width=.175\textwidth]{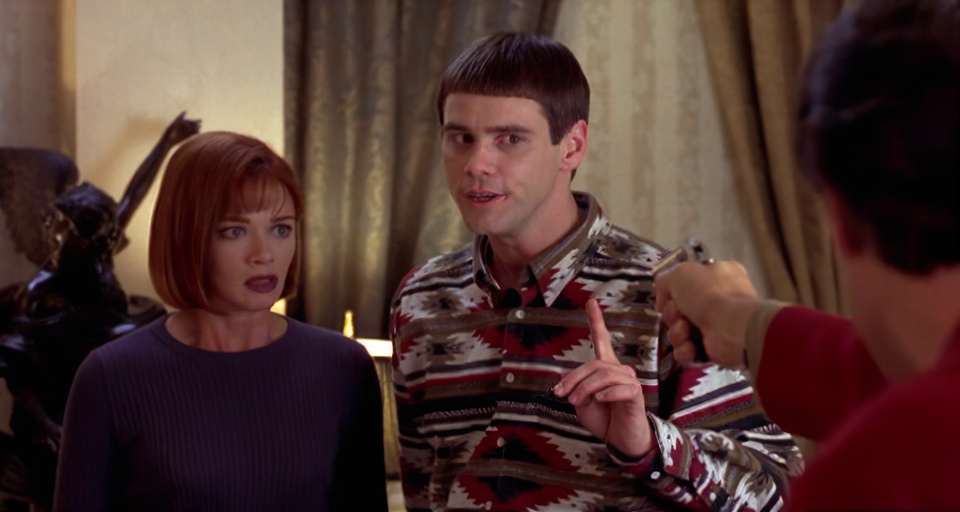} &
\includegraphics[width=.175\textwidth]{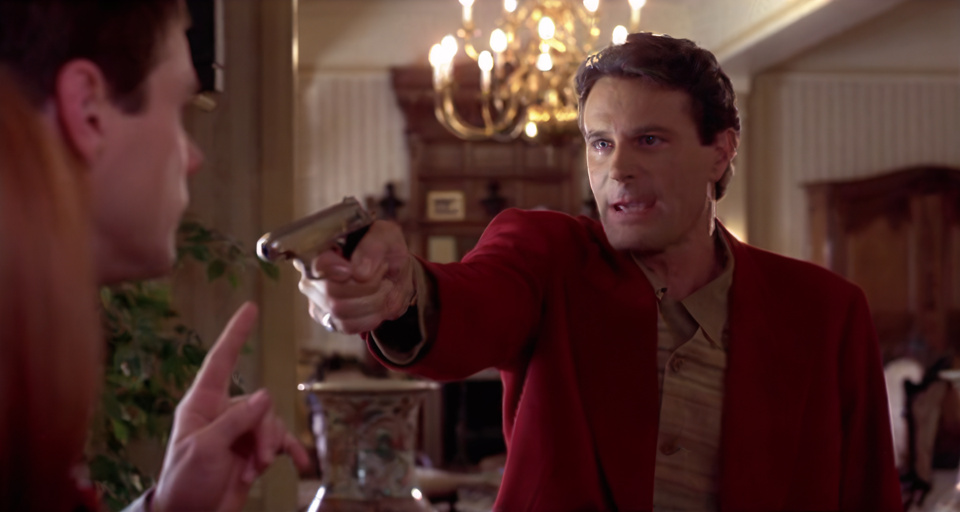} \\[0.8mm]
\end{tabular}
\vspace{0.25em}
\begin{tabular}{p{.08\textwidth}p{.175\textwidth}p{.175\textwidth}p{.175\textwidth}p{.175\textwidth}p{.175\textwidth}}
&
\raggedright\tiny [male0] finally met the man who married his ex wife &
\raggedright\tiny \textcolor{magenta}{he} pointed a gun at the man &
\raggedright\tiny \textcolor{magenta}{he} told the man that he wanted his wife back &
\raggedright\tiny the man explained that he was in love and he would happily take a bullet for his woman &
\raggedright\tiny [male0] decided that he was fine shooting someone and he did just that.
\end{tabular}

\caption{Qualitative results on VWP~\citep{hong2023visual}. We employ our pronoun replacing strategy on top of the text taken directly from the dataset. The narrative escalates from a meeting (frame 1) to a gun threat (frame 2). In frame 2 (``he pointed a gun at the man''), SD3 fails to depict the described interaction and produces unrelated visual content. With contextual guidance from the CORE module, our method better reflects the confrontation described in the story and maintains a more coherent interaction across frames.}
\label{fig:supp3c}  \vspace{-1.5em}
\end{figure*}

\begin{figure*}[!t]
\centering
\setlength{\tabcolsep}{1.2pt}
\renewcommand{\arraystretch}{0}
\begin{tabular}{c c c c c c}
{\tiny \raisebox{4.5ex}{\shortstack{SD3\\(baseline)}}} &
\includegraphics[width=.175\textwidth]{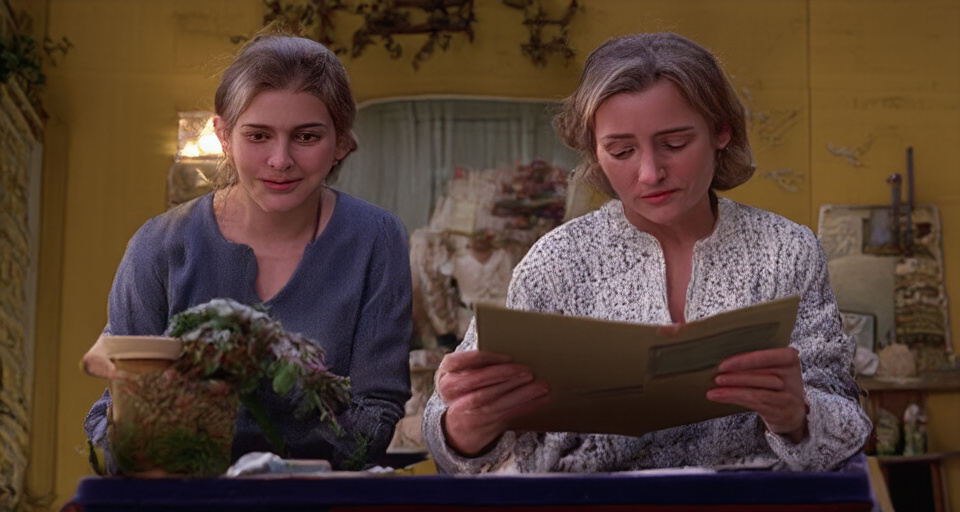} &
\includegraphics[width=.175\textwidth]{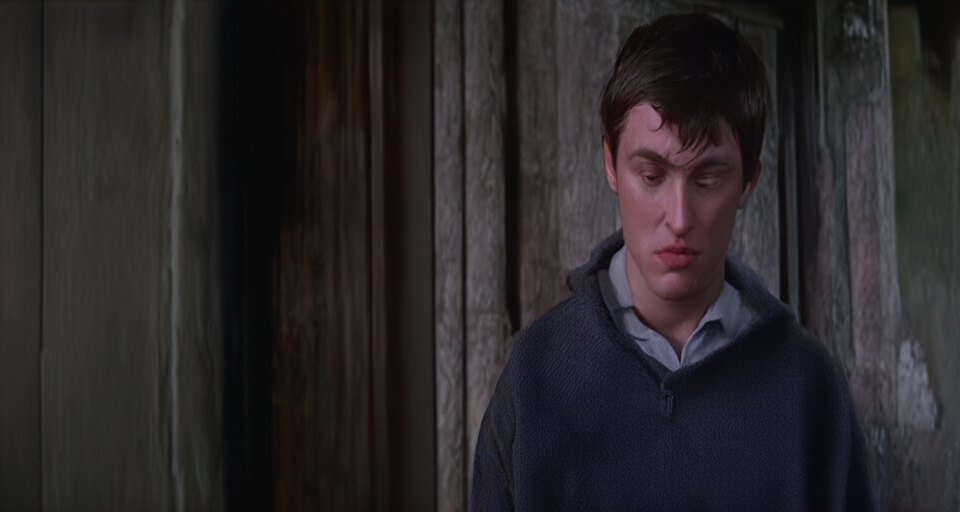} &
\includegraphics[width=.175\textwidth]{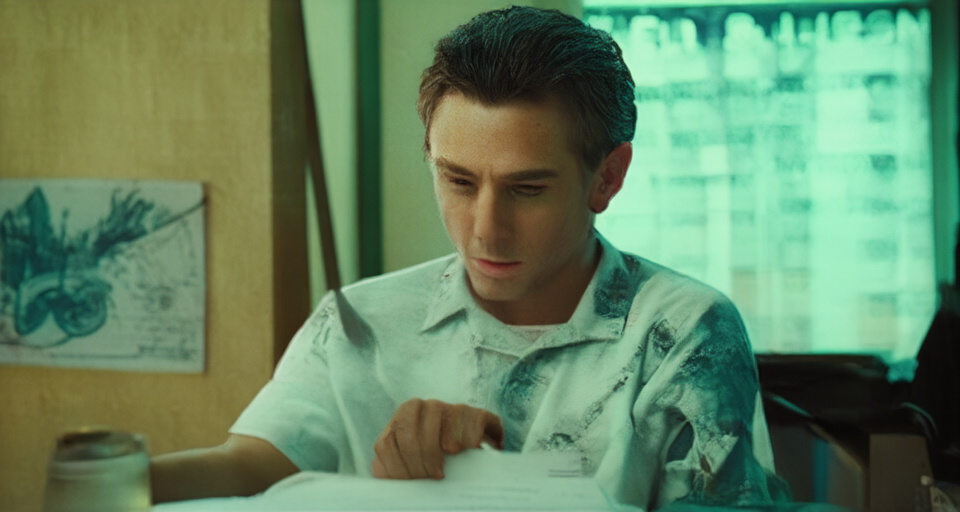} &
\includegraphics[width=.175\textwidth]{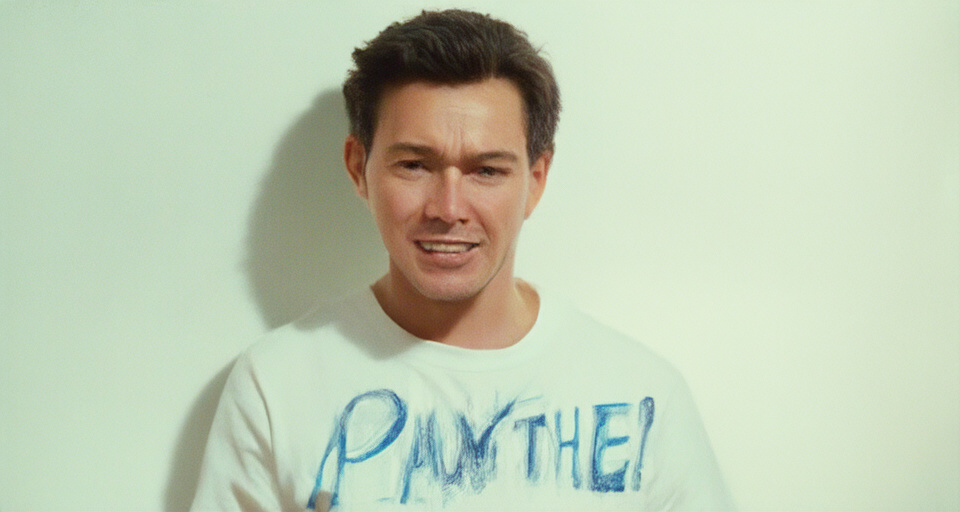} &
\includegraphics[width=.175\textwidth]{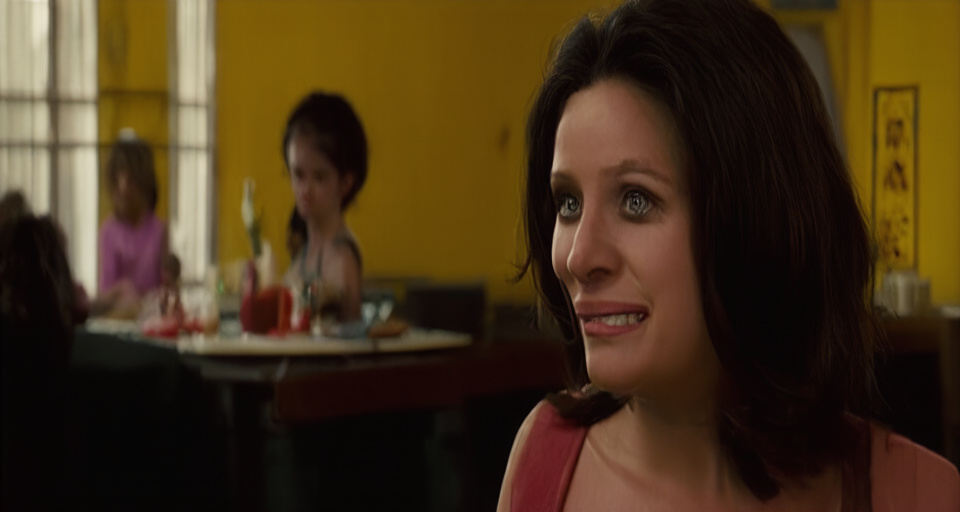} \\[1.5mm]

{\tiny \raisebox{4.5ex}{\shortstack{ReCap\\(Ours)}}} &
\includegraphics[width=.175\textwidth]{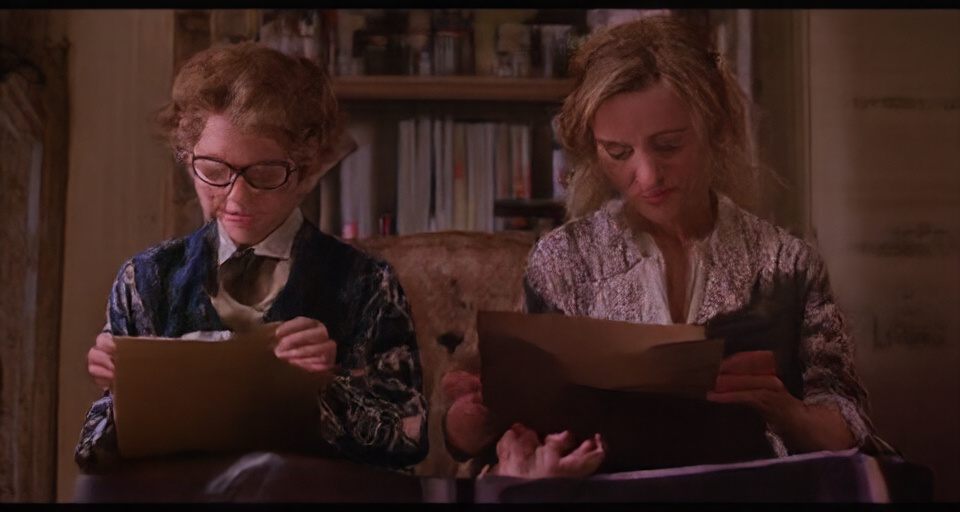} &
\includegraphics[width=.175\textwidth]{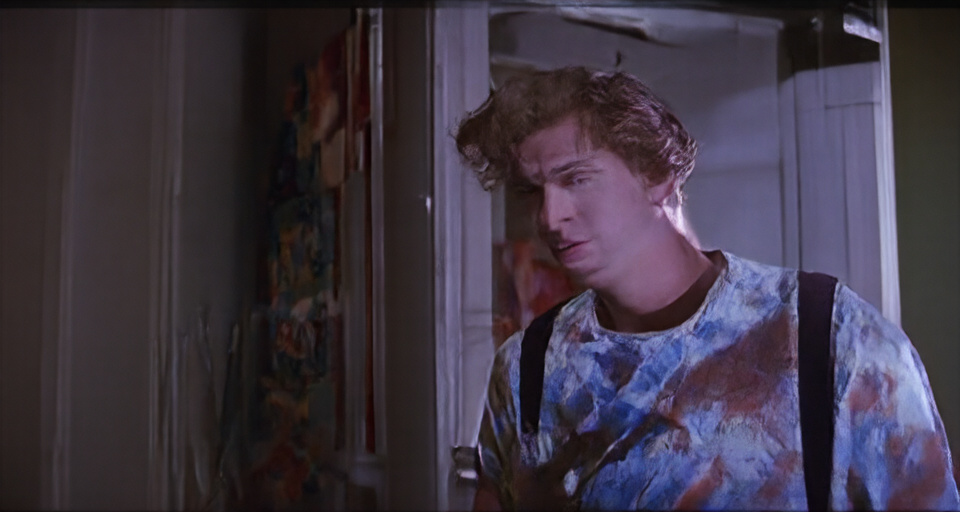} &
\includegraphics[width=.175\textwidth]{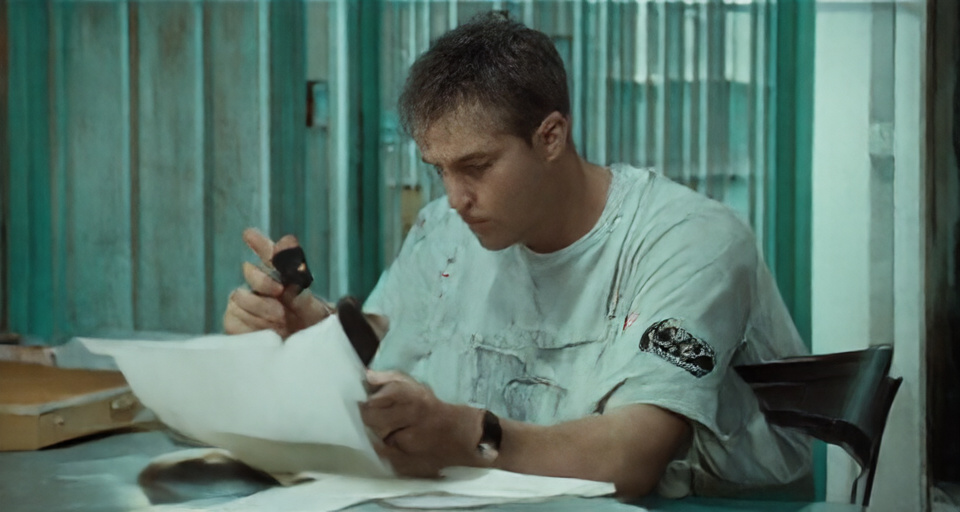} &
\includegraphics[width=.175\textwidth]{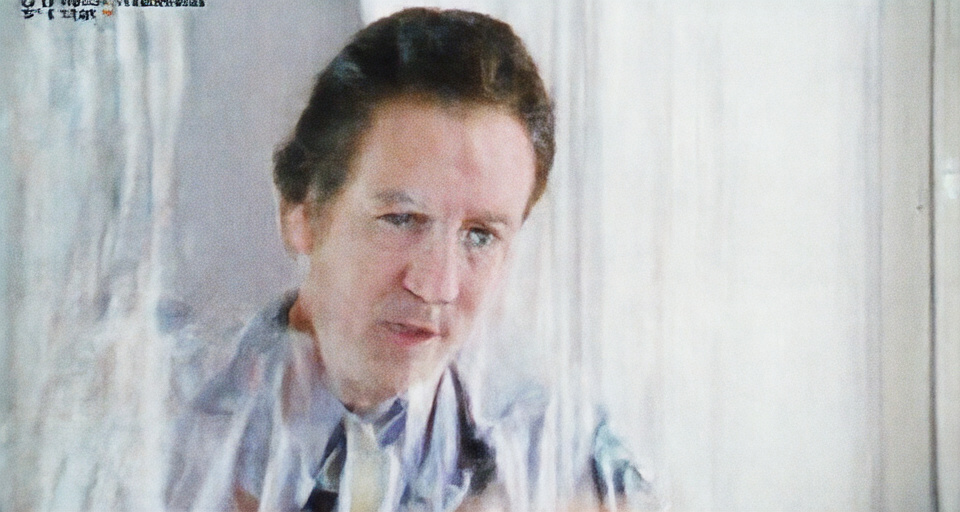} &
\includegraphics[width=.175\textwidth]{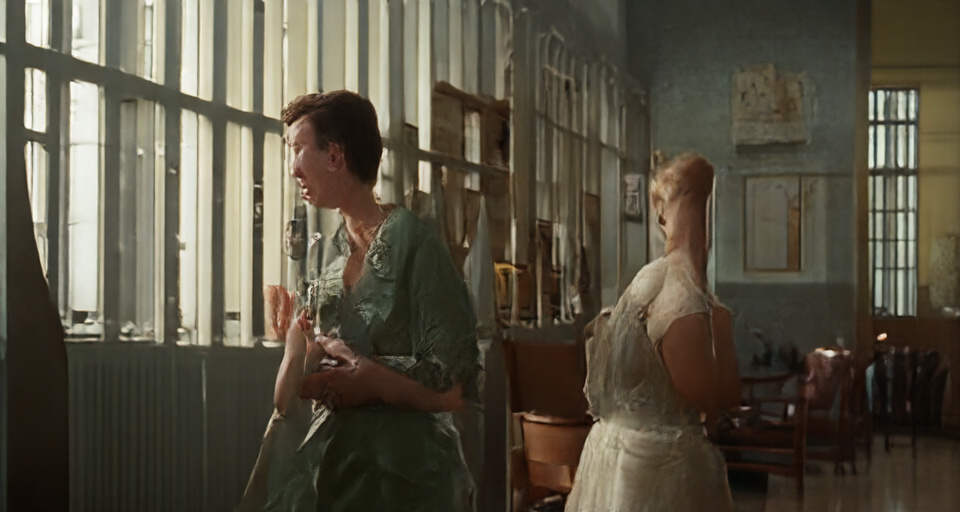} \\[1.5mm]

{\tiny \raisebox{4.5ex}{\shortstack{Ground\\Truth}}} &
\includegraphics[width=.175\textwidth]{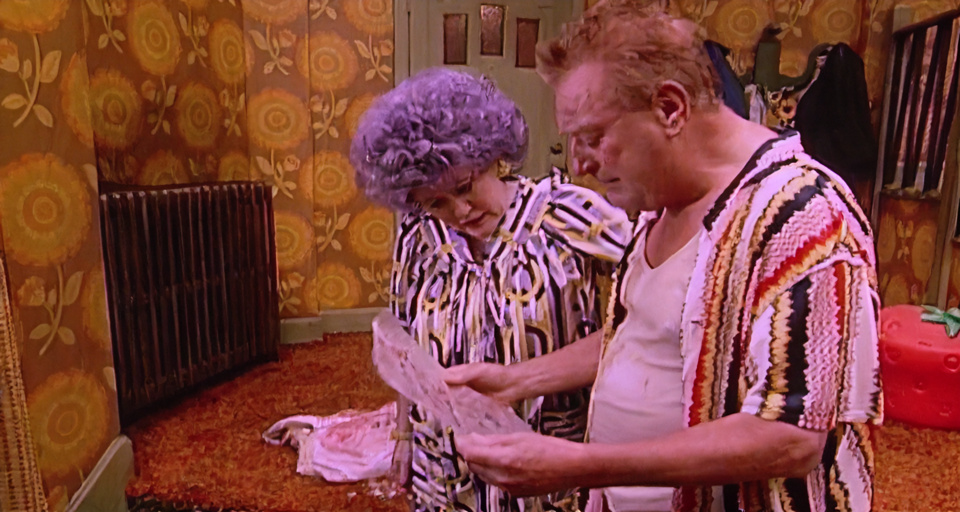} &
\includegraphics[width=.175\textwidth]{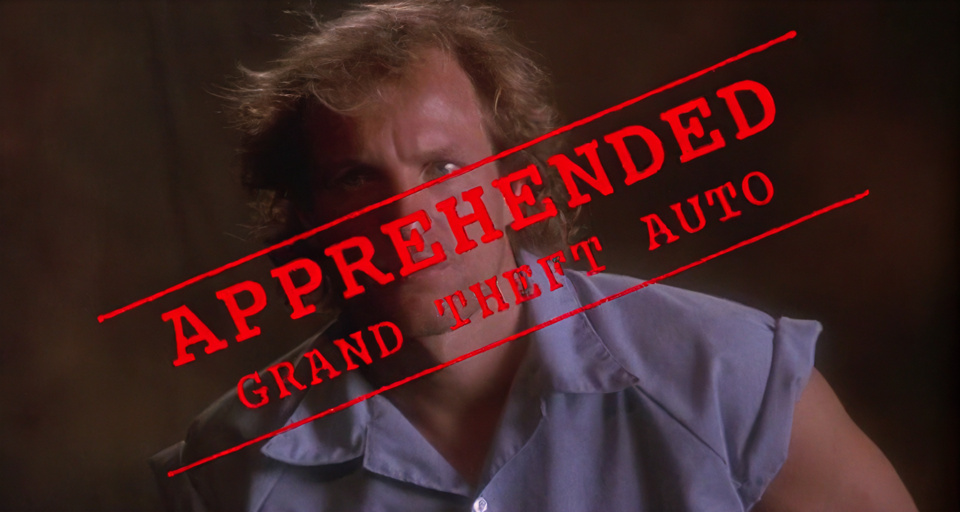} &
\includegraphics[width=.175\textwidth]{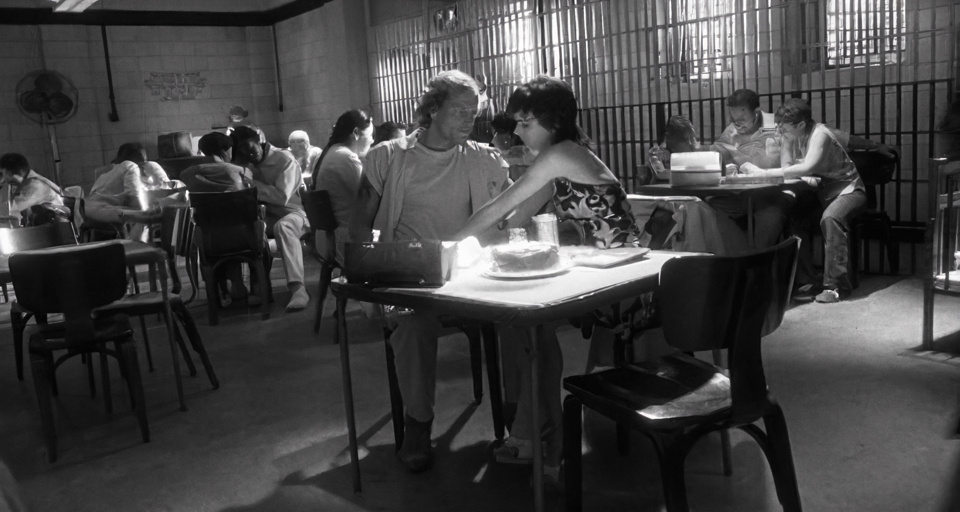} &
\includegraphics[width=.175\textwidth]{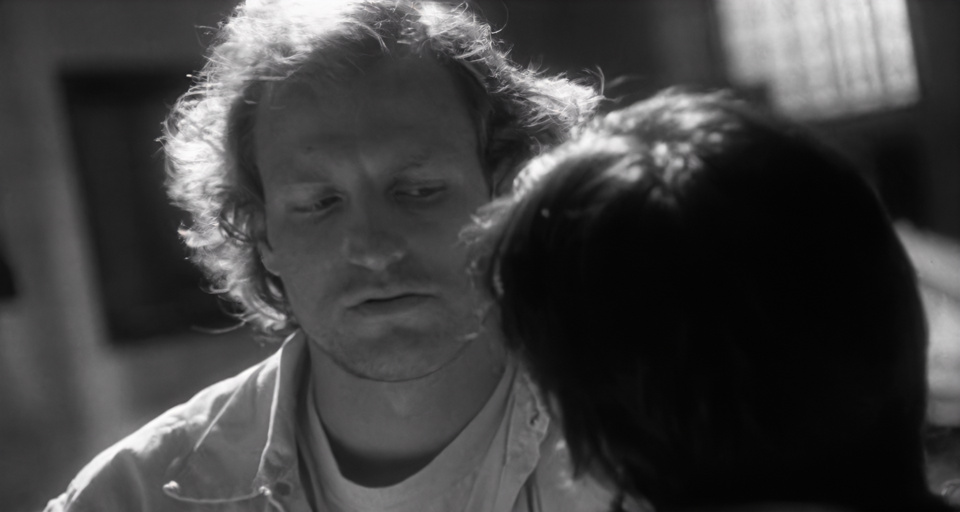} &
\includegraphics[width=.175\textwidth]{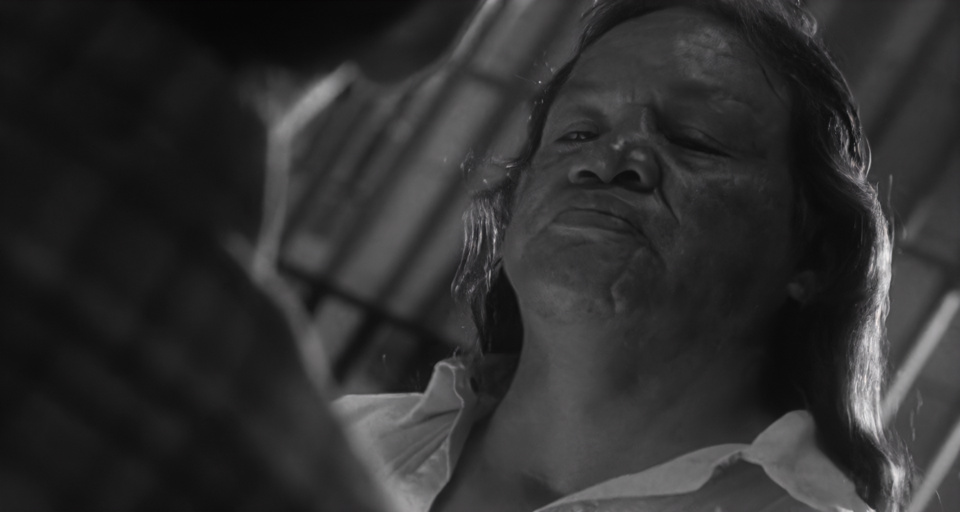} \\[0.8mm]
\end{tabular}
\vspace{0.25em}
\begin{tabular}{p{.08\textwidth}p{.175\textwidth}p{.175\textwidth}p{.175\textwidth}p{.175\textwidth}p{.175\textwidth}}
&
\raggedright\tiny [male0] and [female0] read a letter from their son &
\raggedright\tiny the son , [male1] , tells his parents the story of his arrest , and what he got into prison for he was arrested for grand theft auto , and is currently writing letters to his parents from prison &
\raggedright\tiny \textcolor{magenta}{he} tells a story of his prison days , and about a woman he met in prison &
\raggedright\tiny [male1] chats with this lady in the dining area of the prison , flirting with her &
\raggedright\tiny another prisoner warns [male1] that she is not what she seems , and he is hallucinating due to being locked away from any women for a very long time.
\end{tabular}

\caption{Qualitative results on VWP~\citep{hong2023visual}. We employ our pronoun replacing strategy on top of the text taken directly from the dataset. The story transitions between different narrative moments, beginning with parents reading a letter and continuing with the son’s prison experiences. SD3 produces inconsistent characters and fails to reflect the narrative events described in the text. In contrast, our method better preserves character identity and follows the narrative transition from the letter scene to the prison interactions.}
\label{fig:supp3d}  \vspace{-1.5em}
\end{figure*}

\begin{figure*}[t]
    \centering
    \setlength{\tabcolsep}{2pt}
    \renewcommand{\arraystretch}{0}

    \begin{tabular}{ccccc}
        \raisebox{0.6\height}{\rotatebox{90}{DINOV3}} \includegraphics[width=.19\textwidth]{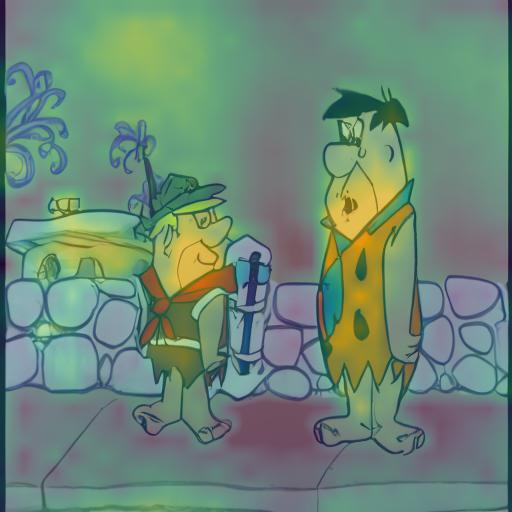} &
        \includegraphics[width=.19\textwidth]{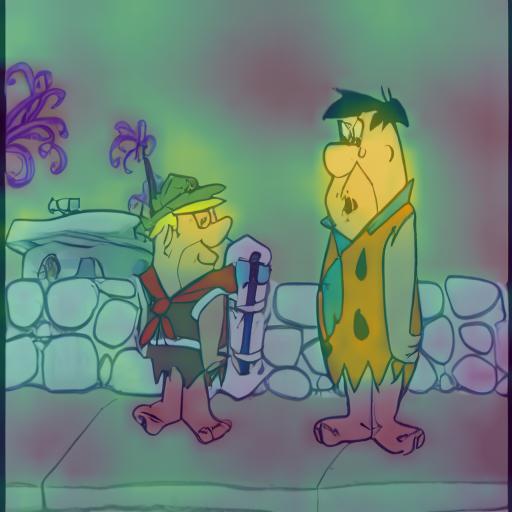} &
        \includegraphics[width=.19\textwidth]{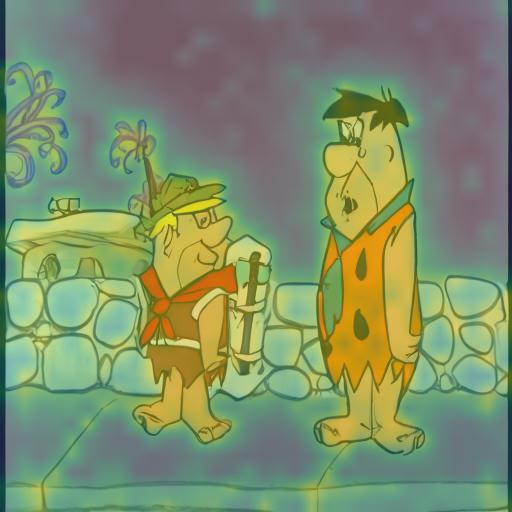} &
        \includegraphics[width=.19\textwidth]{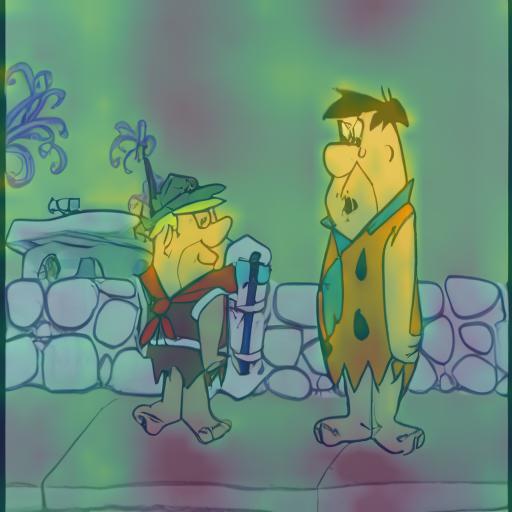} &
        \parbox[c][2.8cm][c]{.19\textwidth}{
        \centering
        \includegraphics[width=.19\textwidth]{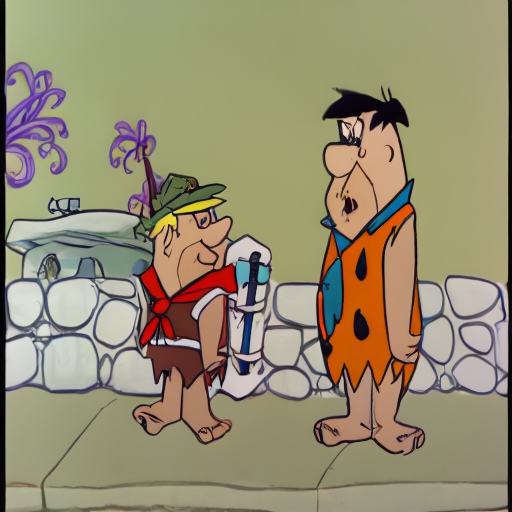} Ground-Truth
} \\[-30pt]
{\footnotesize \texttt{feat8}} & {\footnotesize \texttt{feat11}} & {\footnotesize \texttt{feat25}} & {\footnotesize \texttt{feat34}} & \\[10pt]
        \raisebox{1.6\height}{\rotatebox{90}{CLIP}}
        \includegraphics[width=.19\textwidth]{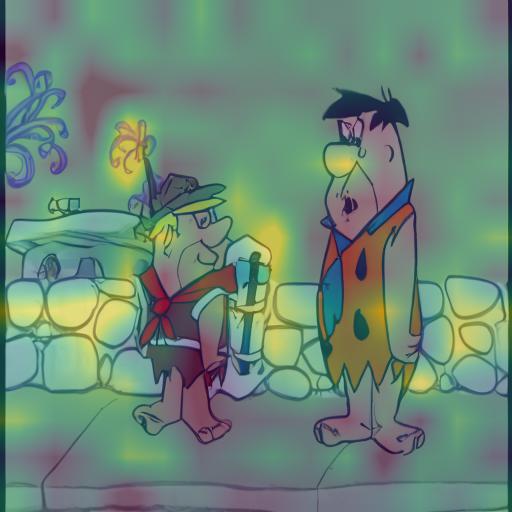} &
        \includegraphics[width=.19\textwidth]{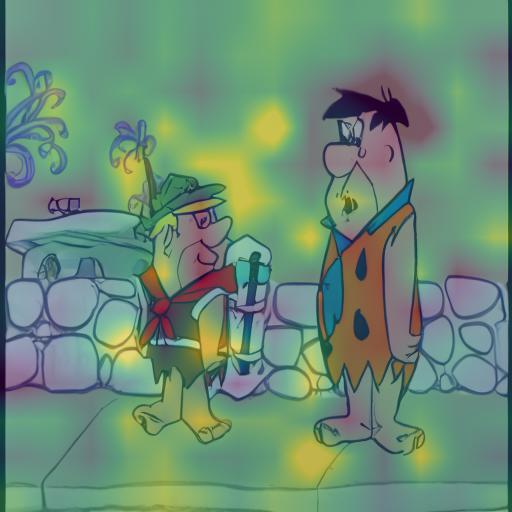} &
        \includegraphics[width=.19\textwidth]{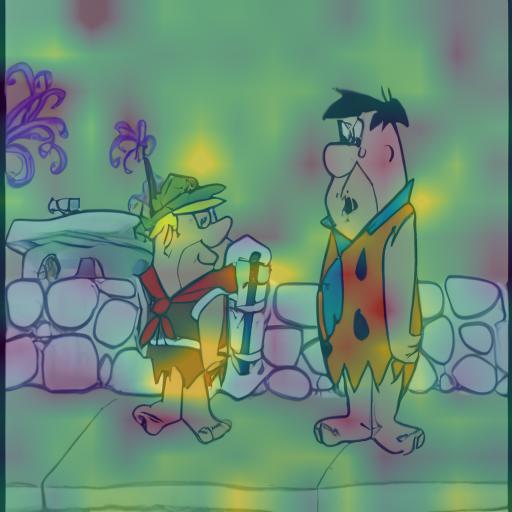} &
        \includegraphics[width=.19\textwidth]{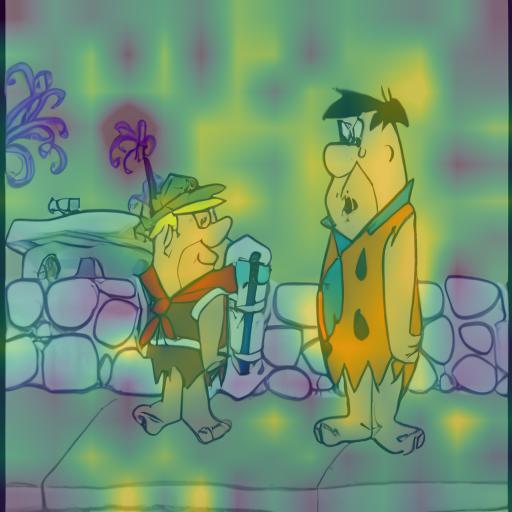} &
         \\[6pt]
{\footnotesize \texttt{feat9}} & {\footnotesize \texttt{feat14}} & {\footnotesize \texttt{feat17}} & {\footnotesize \texttt{feat40}} & \\[10pt]
        \raisebox{0.6\height}{\rotatebox{90}{DINOV3}}
        \includegraphics[width=.19\textwidth]{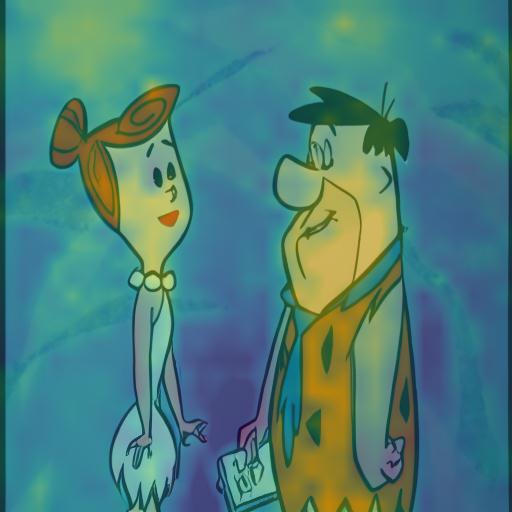} &
        \includegraphics[width=.19\textwidth]{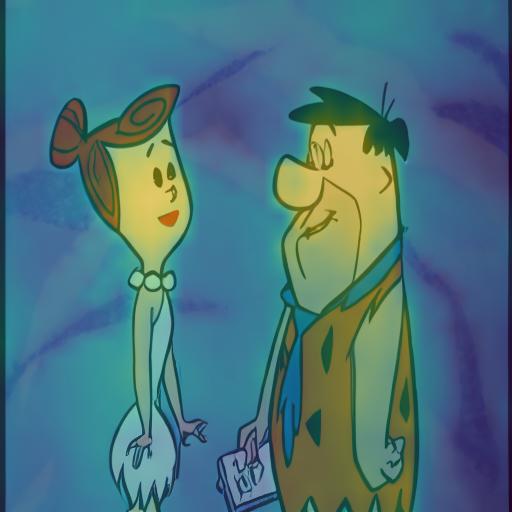} &
        \includegraphics[width=.19\textwidth]{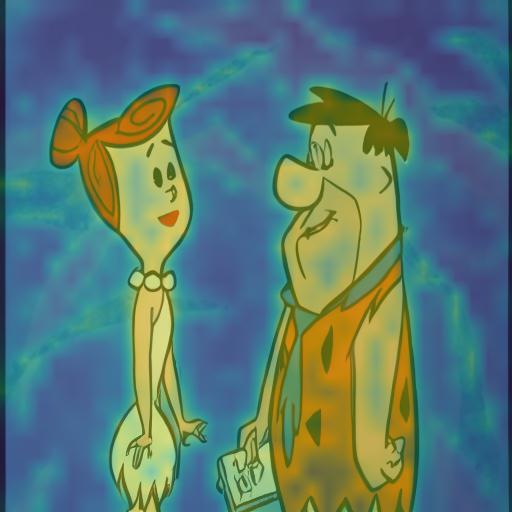} &
        \includegraphics[width=.19\textwidth]{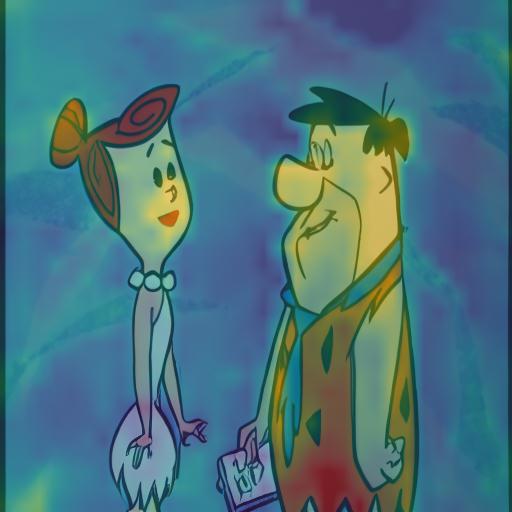} &
        \parbox[c][2.8cm][c]{.19\textwidth}{
        \centering
        \includegraphics[width=.19\textwidth]{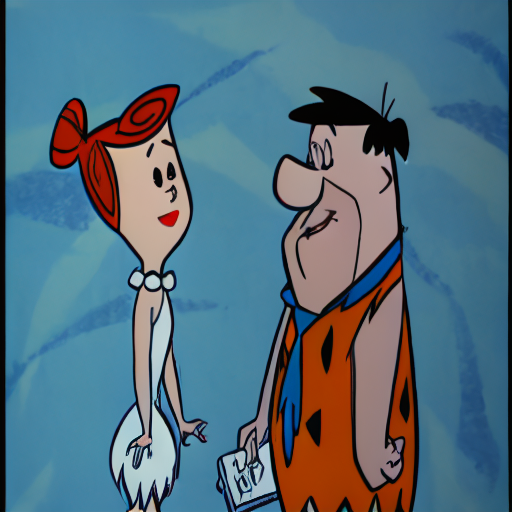} Ground-Truth
} \\[-30pt]
{\footnotesize \texttt{feat8}} & {\footnotesize \texttt{feat11}} & {\footnotesize \texttt{feat25}} & {\footnotesize \texttt{feat34}} & \\[10pt]
        \raisebox{1.6\height}{\rotatebox{90}{CLIP}}
        \includegraphics[width=.19\textwidth]{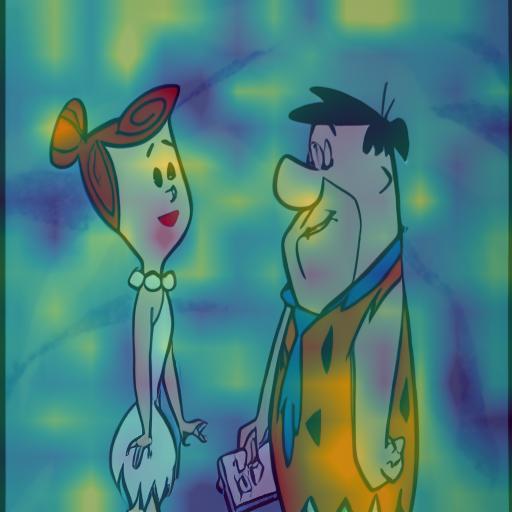} &
        \includegraphics[width=.19\textwidth]{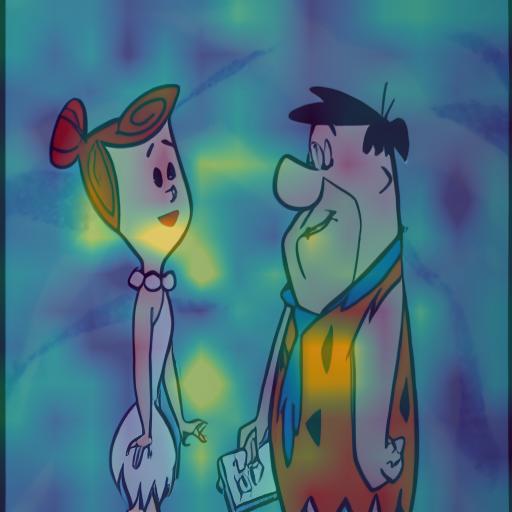} &
        \includegraphics[width=.19\textwidth]{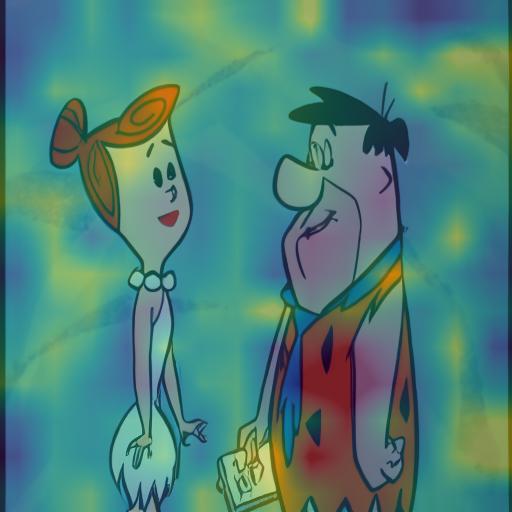} &
        \includegraphics[width=.19\textwidth]{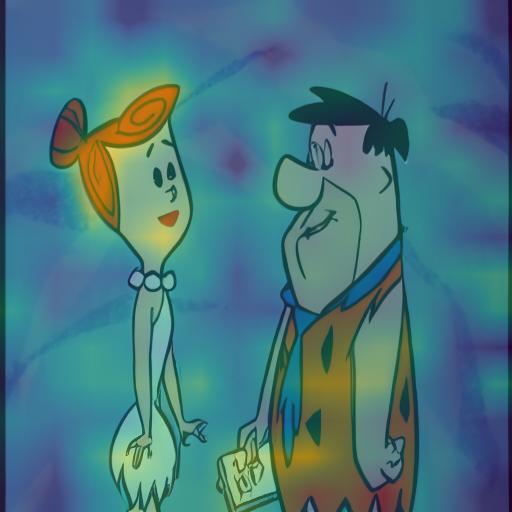} &
         \\[6pt]
{\footnotesize \texttt{feat9}} & {\footnotesize \texttt{feat14}} & {\footnotesize \texttt{feat17}} & {\footnotesize \texttt{feat40}} & \\[10pt]
    \end{tabular}

    \caption{\textbf{Comparison of feature activation maps from DINOv3 and CLIP encoders.} For each ground-truth frame (right), we visualize feature maps from different layers of both DINOv3 (top) and CLIP (bottom) backbones. DINOv3 features exhibit stronger, more localized activations on identity-defining regions such as character faces, clothing, and body contours (e.g., feat8, feat11, feat25, feat34), while CLIP features show more diffuse, global activations across the entire image (e.g., feat9, feat14, feat17, feat40). The heatmaps demonstrate why DINOv3's dense, spatially coherent features are better suited for maintaining fine-grained character consistency in story visualization.}
    \label{fig:dino_vis}
\end{figure*}

In this section, we present additional qualitative and analysis results that further support our main claims. Fig.~\ref{fig:fgbg_plot} examines robustness to longer temporal context and shows that increasing the number of conditioning frames does not degrade performance; our method remains stable in both foreground and background accuracy. On FlintstonesSV, Figs.~\ref{fig:supp1} and~\ref{fig:supp1b} illustrate stronger story visualization and story-text following, where our method better preserves character identity, scene continuity, and expressions across frames, while competing methods often exhibit identity drift, weaker temporal consistency, or less faithful alignment with the descriptions. On PororoSV, Figs.~\ref{fig:supp2a} and~\ref{fig:supp2b} show similar trends, demonstrating that our approach generalizes beyond FlintstonesSV and produces more coherent multi-frame stories with improved character consistency and closer correspondence to the input narrative. Fig.~\ref{fig:dino_vis} provides a feature-level explanation for these improvements: compared with CLIP, DINOv3 yields stronger and more localized activations on identity-defining regions such as faces, clothing, and body contours, which helps preserve fine-grained character consistency. Finally, on VWP, Figs.~\ref{fig:supp3a}--\ref{fig:supp3d} show that our method generates visually more coherent story sequences with better text alignment over longer narratives, capturing evolving scenes, interactions, and character appearance more faithfully than the SD3 baseline.

\paragraph{Long-Range Dependencies} While our $t-1$ conditioning is Markovian, we follow the standard pronoun replacement protocol from \citep{rahman2023makeastory}, where character mentions are replaced by pronouns only in the immediately preceding description ($t-1$). The guidance branch provides contextual cues when no character is present in the current frames. As shown in Fig.~\ref{fig:fgbg_plot}, increasing the number of frames does not degrade performance; our method remains robust in both foreground and background accuracy.

\end{document}